\documentclass[10pt, conference]{IEEEtran}
\IEEEoverridecommandlockouts
\usepackage[noadjust]{cite}
\usepackage{amsmath, amssymb, amsfonts, url}
\usepackage[ruled,vlined]{algorithm2e}
\usepackage{graphicx, tabularx, booktabs, subcaption}
\usepackage{textcomp}
\usepackage{xcolor}
\usepackage{adjustbox}
\usepackage[acronym]{glossaries}

\newcommand{\rev}[1]{\textcolor{black}{#1}}

\def\BibTeX{{\rm B\kern-.05em{\sc i\kern-.025em b}\kern-.08em
    T\kern-.1667em\lower.7ex\hbox{E}\kern-.125emX}}

\begin{document}

\title{Efficient Neural Networks with Discrete Cosine Transform Activations
\thanks{This work is part of the project SOFIA PID2023-147305OB-C32 funded by MICIU/AEI/10.13039/501100011033 and FEDER/UE.}
}

\author{\IEEEauthorblockN{Marc Martinez-Gost\IEEEauthorrefmark{1}\IEEEauthorrefmark{2}, Sara Pepe\IEEEauthorrefmark{1}, Ana Pérez-Neira\IEEEauthorrefmark{1}\IEEEauthorrefmark{2}\IEEEauthorrefmark{3}, 
Miguel Ángel Lagunas\IEEEauthorrefmark{2}}
\IEEEauthorblockA{
\IEEEauthorrefmark{1}Centre Tecnològic de Telecomunicacions de Catalunya, Spain\\
\IEEEauthorrefmark{2}Dept. of Signal Theory and Communications, Universitat Politècnica de Catalunya, Spain\\
\IEEEauthorrefmark{3}ICREA Acadèmia, Spain\\
\{mmartinez, spepe, aperez, m.a.lagunas\}@cttc.es
}}

\newacronym{AAF}{AAF}{adaptive activation functions}
\newacronym{AF}{AF}{activation functions}
\newacronym{AI}{AI}{Artificial Intelligence}
\newacronym{DCT}{DCT}{discrete cosine transform}
\newacronym{INR}{INR}{implicit neural representation}
\newacronym{IoT}{IoT}{Internet of Things}
\newacronym{LMS}{LMS}{least mean squares}
\newacronym{MSE}{MSE}{mean squared error}
\newacronym{MLP}{MLP}{multilayer perceptron}
\newacronym{ENN}{ENN}{expressive neural network}

\maketitle
\begin{abstract}
In this paper, we extend our previous work on the Expressive Neural Network (ENN), a multilayer perceptron with adaptive activation functions parametrized using the Discrete Cosine Transform (DCT). Building upon previous work that demonstrated the strong expressiveness of ENNs with compact architectures, we now emphasize their efficiency, interpretability and pruning capabilities. The DCT-based parameterization provides a structured and decorrelated representation that reveals the functional role of each neuron and allows direct identification of redundant components. Leveraging this property, we propose an efficient pruning strategy that removes unnecessary DCT coefficients with negligible or no loss in performance. Experimental results across classification and implicit neural representation tasks confirm that ENNs achieve state-of-the-art accuracy while maintaining a low number of parameters. Furthermore, up to 40\% of the activation coefficients can be safely pruned, thanks to the orthogonality and bounded nature of the DCT basis. Overall, these findings demonstrate that the ENN framework offers a principled integration of signal processing concepts into neural network design, achieving a balanced trade-off between expressiveness, compactness, and interpretability.
\end{abstract}

\section{Introduction}
\rev{Neural networks have become the cornerstone of modern \gls{AI}, achieving state-of-the-art performance across domains such as computer vision, natural language processing and scientific modeling. Their success stems from the ability to approximate highly nonlinear functions and extract complex structures from data. However, this expressiveness often relies on increasingly large architectures, which are harder to optimize and require substantial computational and energy resources \cite{scale_up}. Such demands become prohibitive in resource-constrained environments like \gls{IoT} or non-terrestrial networks, where processing power and battery capacity are limited. 
Furthermore, in domains such as wireless communications and other safety-critical applications, the lack of explainability and transparency in neural models poses an additional barrier to their trustworthy and adoption. This motivates the search for architectures that remain both efficient and interpretable.}

While various techniques exist to address individual challenges, such as residual connections to mitigate vanishing gradients \rev{\cite{residual}} or dropout to reduce overfitting \rev{\cite{dropout}}, they fall short of solving the underlying problem, which is the dependence on excessively large architectures. Notably, there is growing interest across fields in building smaller, more efficient models, as seen in the rise of small language models \cite{small_LM}.

\rev{A more principled path, instead, lies in redesigning the fundamental building block of neural networks, this is, the neuron itself. The standard neuron computes a weighted sum of its inputs followed by a fixed nonlinear activation function, which largely determines the network’s expressive power. From an implementation perspective, fixed activations are typically designed by hand, guided by theoretical insights, biological inspiration or empirical heuristics. While effective in specific contexts, they are not universally optimal: for instance, output activations vary across tasks (e.g., linear for regression or sigmoid for classification) and no single function can perform well across all datasets, tasks or layers. The Kolmogorov–Arnold theorem underscores the central role of nonlinearities in function approximation, yet it provides no guidance on which activation is best suited for a given problem. This limitation motivates the exploration of adaptive or structured activation mechanisms that offer finer control over model behavior and efficiency. Building on this idea, \gls{AAF} have emerged as a promising direction to enhance expressiveness at the unit level, enabling more compact and powerful architectures \cite{400_activations}. However, many existing \gls{AAF} still rely on specific parametric forms that restrict their adaptability and hinder interpretability.}

In our previous work, we introduced a nonlinear processing unit based on the \gls{DCT} and adapted using the \gls{LMS} algorithm. This design offers several key advantages: the DCT provides a natural compression that reduces the number of parameters, the LMS adaptation allows precise control over convergence speed, and the structure makes it possible to characterize the functions that the unit can represent. Building on this concept, we integrated the DCT-based processing unit directly into the artificial neuron as an \gls{AAF}. The resulting architecture, which we refer to as the \gls{ENN} \cite{ENN}, combines high learning capacity with a compact parameterization.

\rev{This paper builds upon the ENN framework to further demonstrate its effectiveness and generality across diverse learning tasks. We evaluate the model on binary classification and \gls{INR}, empirically showing that the ENN consistently outperforms state-of-the-art architectures. Beyond improved performance, the ENN exhibits a remarkably compact and structured parameterization, which simplifies model configuration and training. In the context of INR, we further reveal a distinctive advantage of the ENN: its parameters, corresponding to the DCT coefficients, can be easily pruned due to their natural decorrelation, leading to efficient compression without significant loss of performance.
}

The main contributions of this paper are described in the following:
\begin{enumerate}
    \item A comprehensive analysis of model complexity is presented, considering both the number of trainable parameters and the computational cost associated with the forward and backward passes.
    \item An extensive experimental evaluation is conducted, comparing the proposed ENN with state-of-the-art benchmarks on binary classification and \gls{INR} tasks.
    \item The intrinsic redundancies within the ENN are investigated, and a pruning algorithm is proposed to effectively eliminate them without compromising the performance.
\end{enumerate}

The remaining part of the paper proceeds as follows:
Section II introduces the \gls{ENN} model. Section III evaluates and compares its performance on binary classification tasks, while Section IV focuses on image regression. Section V analyzes the redundancies identified in the ENN, presents a proposed pruning algorithm and assesses its effectiveness.
Finally, Section VI concludes the paper.

\section{Expressive Neural Network}

Consider a \gls{MLP} with $L$ layers, where $M_{\ell}$ denotes the number of neurons in layer $\ell\in\{1,\dots,L\}$. 
The input to layer $\ell$, denoted by $\mathbf{s}_{\ell-1}$, corresponds to the output of the previous layer and is processed as
\begin{align}
    \mathbf{z}_{\ell} &= \mathbf{W}_{\ell}^T\mathbf{s}_{{\ell}-1} + \mathbf{b}_{\ell},
    \label{eq:linear_phase}\\
    \overline{\mathbf{z}}_{\ell} &= \frac{N}{2}(\mathbf{z}_{\ell}+1),
    \label{eq:scaling_phase}\\
    \mathbf{s}_{\ell} &= \sigma_{{\ell}} (\overline{\mathbf{z}}_{\ell}),
    \label{eq:non_linear_phase}
\end{align}
where $\mathbf{s}_0=\mathbf{x}$ and ${s}_L=\hat{y}$ corresponds to the input and output data of the model. We consider a supervised setting with a dataset $\mathcal{D}=\{\mathbf{x}^{(i)},y^{(i)}\}_{i=1}^{D}$, where $\mathbf{x}^{(i)}\in\mathbb{R}^{M_0}$ represents the $i$th sample and $M_0$ is the dimensionality of the input space; the corresponding reference or ground truth is $y^{(i)}\in\mathbb{R}$ and $D$ is the total number of samples in the dataset.

\rev{Equation \eqref{eq:linear_phase} corresponds to the linear processing, characterized by the matrix of linear weights $\mathbf{W}_{\ell}\in\mathbb{R}^{M_{\ell-1}\times M_{\ell}}$ and the bias vector $\mathbf{b}_{\ell}\in \mathbb{R}^{\ell}$.
Equation \eqref{eq:non_linear_phase} corresponds to the nonlinear processing at the neuron, computed element-wise. In the ENN, the nonlinear function at layer $\ell$ and neuron $m$ corresponds to
\begin{equation}
    \sigma_{\ell}(\bar{{z}}_\ell[m])=\sum_{q=1}^{Q} F_{\ell mq} \cos\left( \frac{\pi(2q - 1)(2\bar{{z}}_\ell[m] - 1)}{2N} \right),
    \label{eq:AAF}
\end{equation}
where $\bar{{z}}_\ell[m]$ is the $m$th entry of vector $\bar{\mathbf{z}}_\ell$ and $F_{\ell m q}$ is the $q$th coefficient. Notice that \eqref{eq:AAF} corresponds to a DCT approximation of a function with $Q$ coefficients. Within this parametrization, $N$ is the resolution of the DCT and the term $(2q-1)$ indicates that we are only using odd-indexed coefficients. As shown in \cite{ENN}, odd coefficients allow to represent a wide variety of functions and halves the number of parameters in the neuron.
Lastly, the scaling in \eqref{eq:scaling_phase} is necessary because input data are typically normalized to $\mathbf{x}\in[-1,1]^{M_0}$, whereas the DCT operates on inputs within $[0,N-1]$. Although this transformation could, in principle, be learned implicitly by the network, applying it explicitly facilitates training and ensures numerical consistency across layers.}

\rev{The parameters $F_{\ell m q}$ in \eqref{eq:AAF} correspond to the DCT coefficients. When these coefficients are made trainable, the mapping in \eqref{eq:AAF} becomes an \gls{AAF}, allowing each neuron to dynamically adjust its nonlinearity during training. We refer the reader to \cite{ENN} for a detailed description of the ENN architecture and the analytical gradients used to update the DCT coefficients.}

\subsection{Model Complexity}
The ENN introduces only a marginal increase in the number of trainable parameters relative to a standard MLP. In a conventional setting, each neuron has $M_\ell$ weights and a bias. In our formulation, $Q$ additional parameters are required to represent the learnable activation function. Summing across all layers and neurons, the total number of trainable parameters becomes
\begin{equation}
    N_\text{p}=\sum_{\ell=1}^L\sum_{m_\ell=1}^{M_\ell} (m_{\ell}+Q+1) \approx \sum_{\ell=1}^L M(M+Q+1),
    \label{eq:complexity}
\end{equation}
assuming a uniform width $M_\ell = M$ across layers. The leading-order complexity is therefore $\mathcal{O}(LM^2 + LMQ)$. For large networks, where $M \gg Q$, the dominant term is $\mathcal{O}(LM^2)$, indicating that the introduction of the DCT-based AAF does not significantly inflate the parameter count. This contrasts with traditional polynomial models, which significantly require more parameters to achieve comparable approximation accuracy.

Regarding computational complexity, evaluating a DCT involves computing a finite sum of cosine terms, which is computationally efficient and compatible with modern hardware. Furthermore, the backpropagation step admits an analytical closed-form expression for the gradient of the activation function with respect to its inputs and coefficients. This facilitates efficient training by eliminating the need for numerical differentiation or symbolic computation. 

In summary, the DCT-based activation design achieves a favorable balance between expressiveness and efficiency. It enhances the functional richness of neural networks through periodic, learnable nonlinearities, while incurring only a modest increase in parameters and computational cost. 

\subsection{Network Design Principles}
The Kolmogorov-Arnold representation theorem provides valuable theoretical insight into the structure of neural networks. The theorem states that any continuous function can be decomposed as 
\begin{equation} f(x_1,\dots,x_{M_0}) = \sum_{i=1}^{2{M_0}+1} \Phi_i\left( \sum_{j=1}^{M_0} \phi_{ij}(x_j) \right), \label{eq:kolmo} \end{equation}
where $\Phi_i$ and $\phi_{ij}$ are univariate functions, known as the outer and inner functions, respectively. This result suggests a fundamental role for univariate nonlinearities in constructing two-layer MLPs \cite{kolmo_mlp}. In particular, the inner functions resemble activation functions applied in the first (i.e., hidden) layer, while the outer functions correspond to the output layer transformations.

The Kolmogorov–Arnold theorem carries two key implications for neural network design.
First, it motivates that a single hidden layer is theoretically sufficient to approximate any continuous multivariate function, highlighting the expressive power of shallow architectures.
Second, it establishes a lower bound on the number of univariate functions required for such a representation: for an input dimension $M_0$, at most $2M_0 + 1$ neurons in the hidden layer are needed.

\section{Binary Classification}

\Gls{INR} offers a novel framework for signal modeling by parameterizing signals as continuous functions rather than relying on traditional discrete representations. Conventional signal formats store values explicitly at fixed resolution points. For instance, images are composed of pixel grids and audio signals are defined by discrete amplitude samples. In contrast, INR define a signal as a continuous mapping from coordinate space to signal values. Since such functions are generally intractable in closed form, they can approximated using neural networks, which implicitly encode the entire signal through learned parameters.

The task involves learning a mapping from input pixel coordinates $\mathbf{x}_n \in [-1,1]^2$ to gray-scaled pixel amplitudes $y_n\in[-1,1]$. The model is trained to minimize the \gls{MSE} between the predicted and true pixel values,
\begin{equation}
    \varepsilon_n^2=(y_n-\hat{y}_n(\mathbf{x}_n))^2,
    \label{eq:loss}
\end{equation}
where we explicitly write the output as a function of the input data.

The most common application of INR lies in image learning and representation, where neural networks are trained to reconstruct images by learning the mapping from spatial coordinates to pixel intensities. While this technique claims to provide significant advantages in various applications, we argue that INR may not be the most appropriate approach for image representation. In fact, the 2-dimensional DCT achieves excellent performance with much lower complexity, requiring no training and offering a compact representation that leverages well-understood statistics of natural images. Therefore, in this work we only consider INR as an additional benchmark to evaluate and challenge our proposed models in a more general and difficult setting for function representation.

The binary classification problem can be naturally interpreted as a particular case of an INR, where the objective is to learn a continuous mapping from input coordinates to class labels. For simplicity, we consider a two-dimensional input $\mathbf{x}_n \subset\mathbb{R}^2$ and a binary ground truth $y_n \in \{-1,1\}$. Under this formulation, we treat this task as a regression problem.
The MSE can be directly employed as the loss function for training and enable direct optimization with continuous-valued losses. This choice simplifies the optimization process and makes the learning dynamics more interpretable, since the model is trained to minimize a continuous error metric rather than a probabilistic loss such as cross-entropy.

Compared to conventional INR with discrete images, this problem is inherently more challenging, as it involves learning a decision boundary defined over a continuous input domain. In other words, the model must approximate a continuous target function defined over a domain of theoretically infinite resolution.


\subsection{Benchmarks and Training Setup}
We consider a dataset $\mathcal{D} = \{(\mathbf{x}_n, y_n)\}$, where $\mathbf{x}_n$ is normalized to $[-1,1]^2$ and size $D = 400.000$ samples drawn from a uniform distribution over the input domain.

According to \eqref{eq:kolmo}, with $M_0 = 2$ input dimensions, an MLP requires $M_1=5$ neurons and a single hidden layer.
In practice, we configure the ENN with $M_1 = 6$ to provide a small performance margin while maintaining the theoretical guidance offered by the theorem. We set $Q=6$ and $N=512$ for all the experiments.

We compare the ENN against selected benchmark architectures at the same total number of trainable parameters to ensure a fair comparison:
\begin{itemize}
    \item \textbf{ReLU:} A classical MLP with fixed ReLU. This model is considered due to its wide use in contemporary neural network architectures. Since the network has no AAF, the width of the hidden layer can be increased up to $M_1=17$.
    \item \textbf{Fourier:} An MLP with the Fourier series AAF \cite{liao2020fourier}. To ensure fairness, we fix the width at $M_1=6$, which limits the number of coefficients to $Q=2$, since each frequency component includes both sine and cosine terms.
    \item \textbf{Kolmogorov–Arnold Network (KAN):} Proposed in \cite{liu2024kan}, KAN replaces the architecture of an MLP by the one proposed in the Kolmogorov–Arnold representation theorem. Therefore, each input is processed independently by an AAF and the outputs are linearly combined. The result is finally processed by another AAF. Each AAF is modeled as a quadratic spline ($k=2$) with $G=3$ intervals.
\end{itemize}

All models are trained using stochastic gradient descent (SGD) with a learning rate of $\alpha_2 = 10^{-3}$. 

\begin{table}[t]
  \begin{center}
    \begin{tabular}{cc|cccc}
      \toprule 
       & \small{Map} \hspace{1.pt} & \hspace{1.pt} \small{ReLU} & \hspace{1.pt} \small{KAN} & \hspace{1.pt} \small{Fourier} & \hspace{1.pt} \small{ENN}\\
      \midrule
      \normalsize{$\text{P}_1$} &
        \adjustbox{valign=c}{
          \makebox[0.3in][c]{
            \begin{tabular}{@{}c@{}}
              \scriptsize{}\\[-0.em]
              \includegraphics[width=0.5in, angle=90]{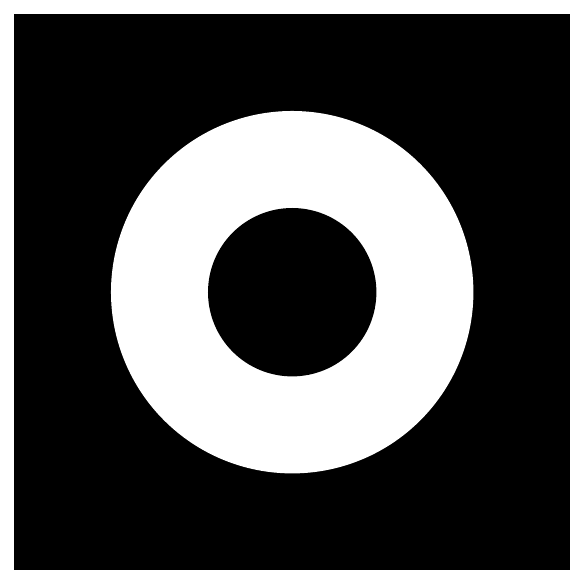}
            \end{tabular}
          }
        } \hspace{1.pt}
        & \hspace{1.pt}
        \adjustbox{valign=c}{
          \makebox[0.3in][c]{
            \begin{tabular}{@{}c@{}}
              \small{89.19\%}\\[-0.0em]
              \includegraphics[width=0.5in, angle=90]{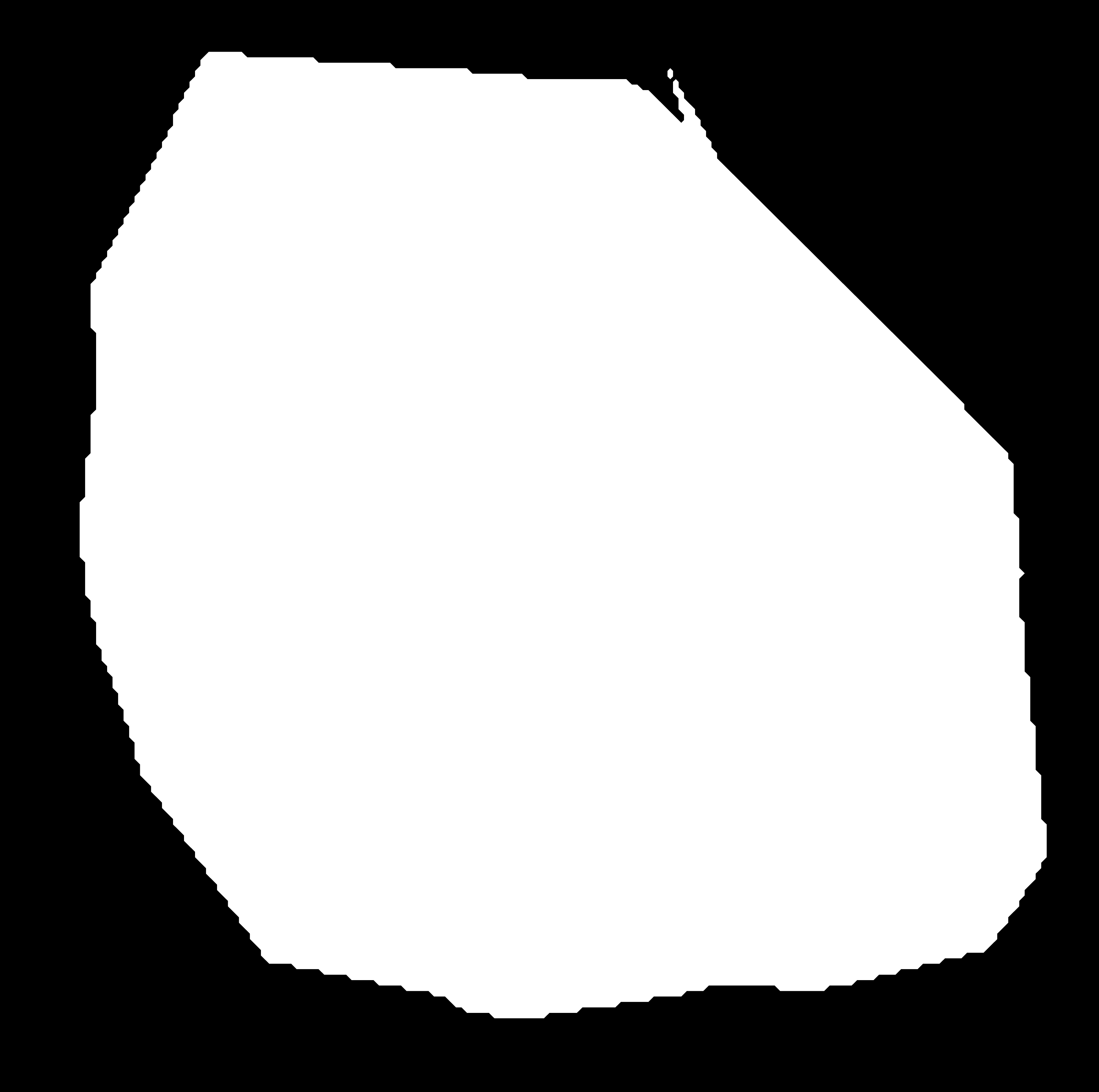}
            \end{tabular}
          }
        } 
        & \hspace{1.pt}
        \adjustbox{valign=c}{
          \makebox[0.3in][c]{
            \begin{tabular}{@{}c@{}}
              \small{98.69\%}\\[-0.em]
              \includegraphics[width=0.5in, angle=90]{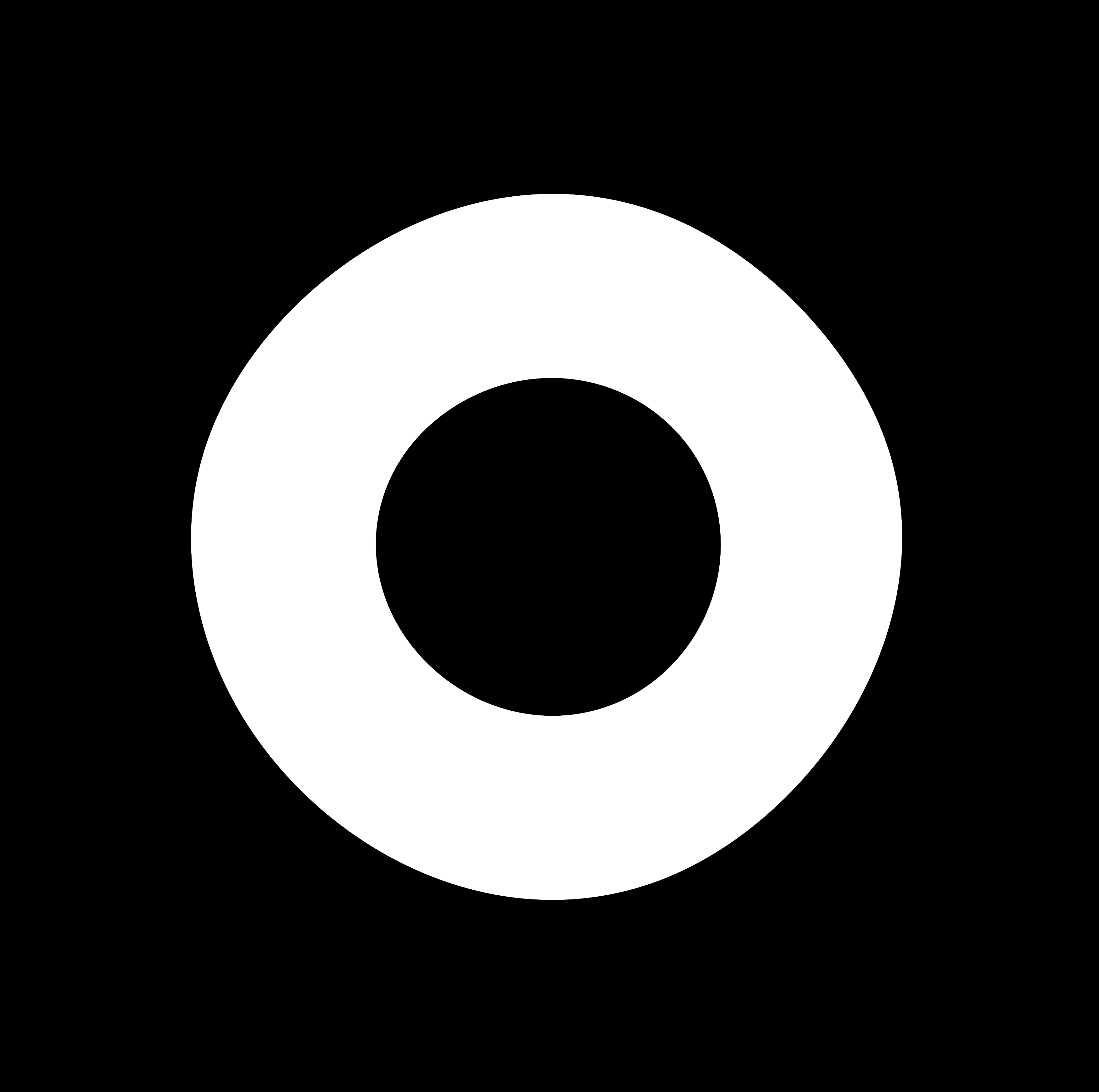}
            \end{tabular}
          }
        }
        & \hspace{1.pt}
        \adjustbox{valign=c}{
          \makebox[0.3in][c]{
            \begin{tabular}{@{}c@{}}
              \small{97.97\%}\\[-0.em]
              \includegraphics[width=0.5in, angle=90]{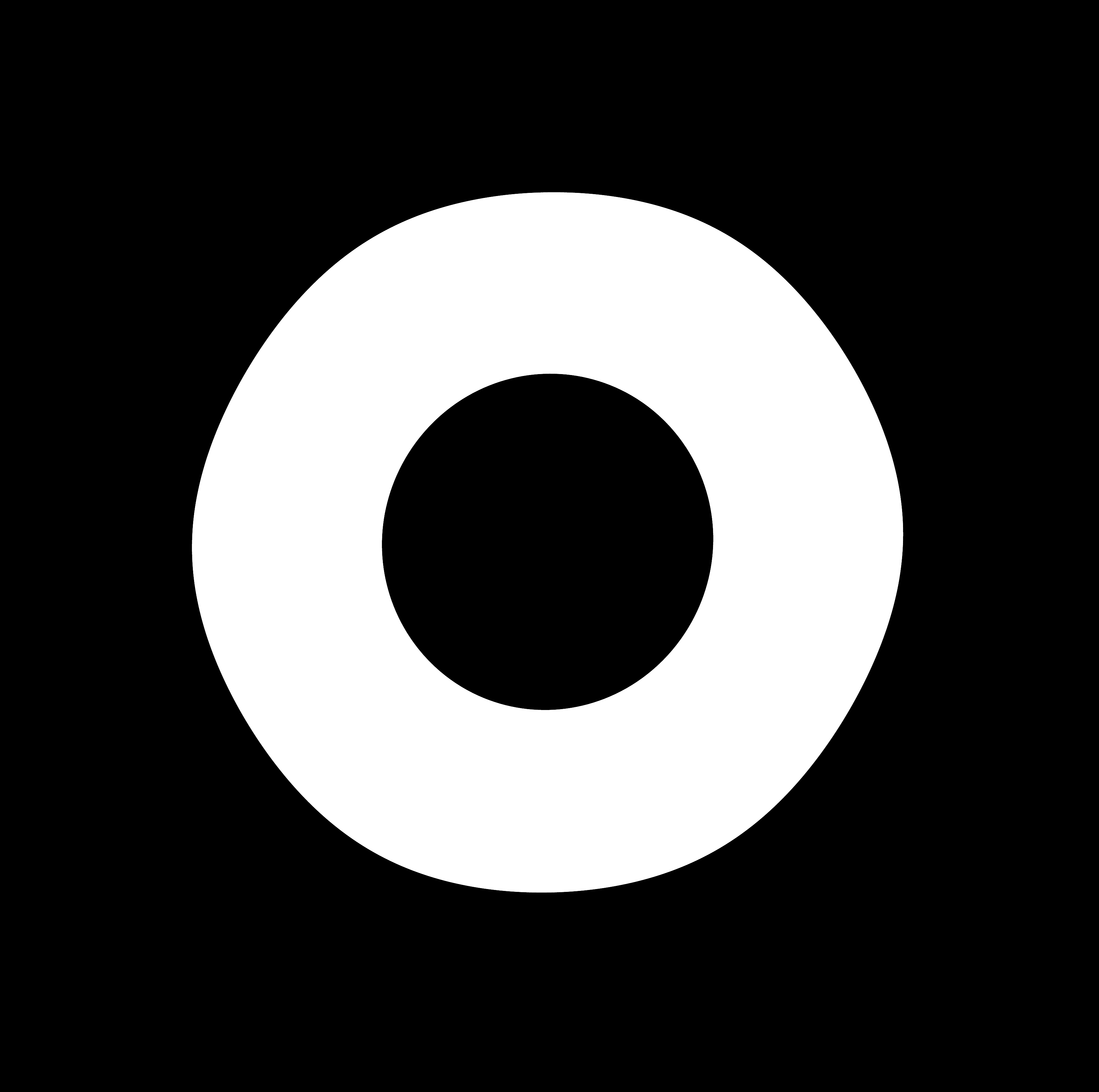}
            \end{tabular}
          }
        } 
        & \hspace{1.pt}
        \adjustbox{valign=c}{
          \makebox[0.3in][c]{
            \begin{tabular}{@{}c@{}}
              \small{\textbf{99.48\%}}\\[-0.em]
              \includegraphics[width=0.5in, angle=90]{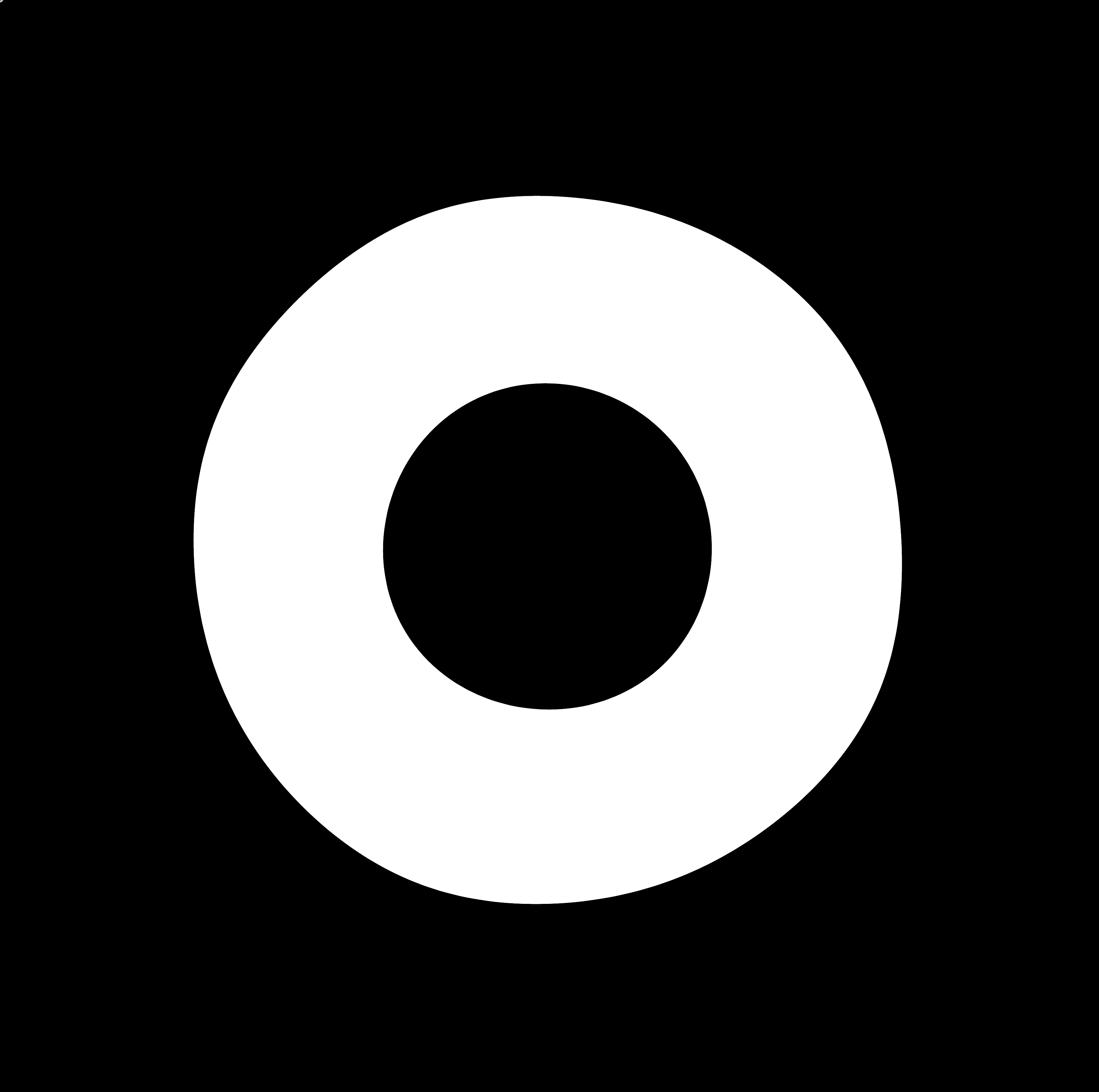}
            \end{tabular}
          }
        } \\[3.2em]
      
      \normalsize{$\text{P}_2$} &
        \adjustbox{valign=c}{
          \makebox[0.3in][c]{
            \begin{tabular}{@{}c@{}}
              \scriptsize{}\\[-0.em]
              \includegraphics[width=0.5in, angle=90]{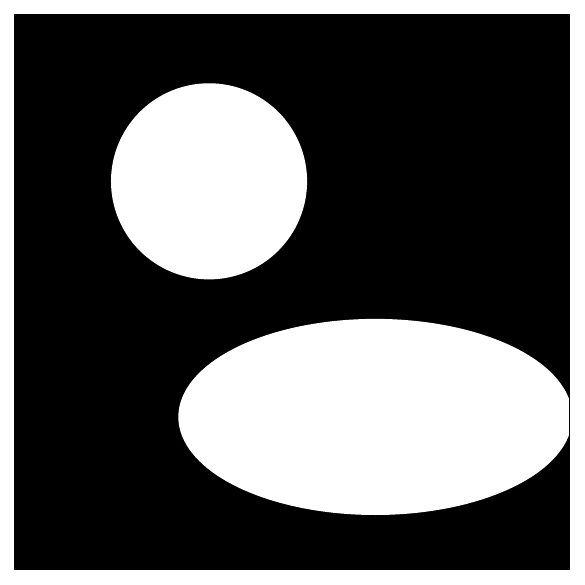}
            \end{tabular}
          }
        } \hspace{1.pt}
        & \hspace{1.pt}
        \adjustbox{valign=c}{
          \makebox[0.3in][c]{
            \begin{tabular}{@{}c@{}}
              \small{96.27\%}\\[-0.em]
              \includegraphics[width=0.5in, angle=0]{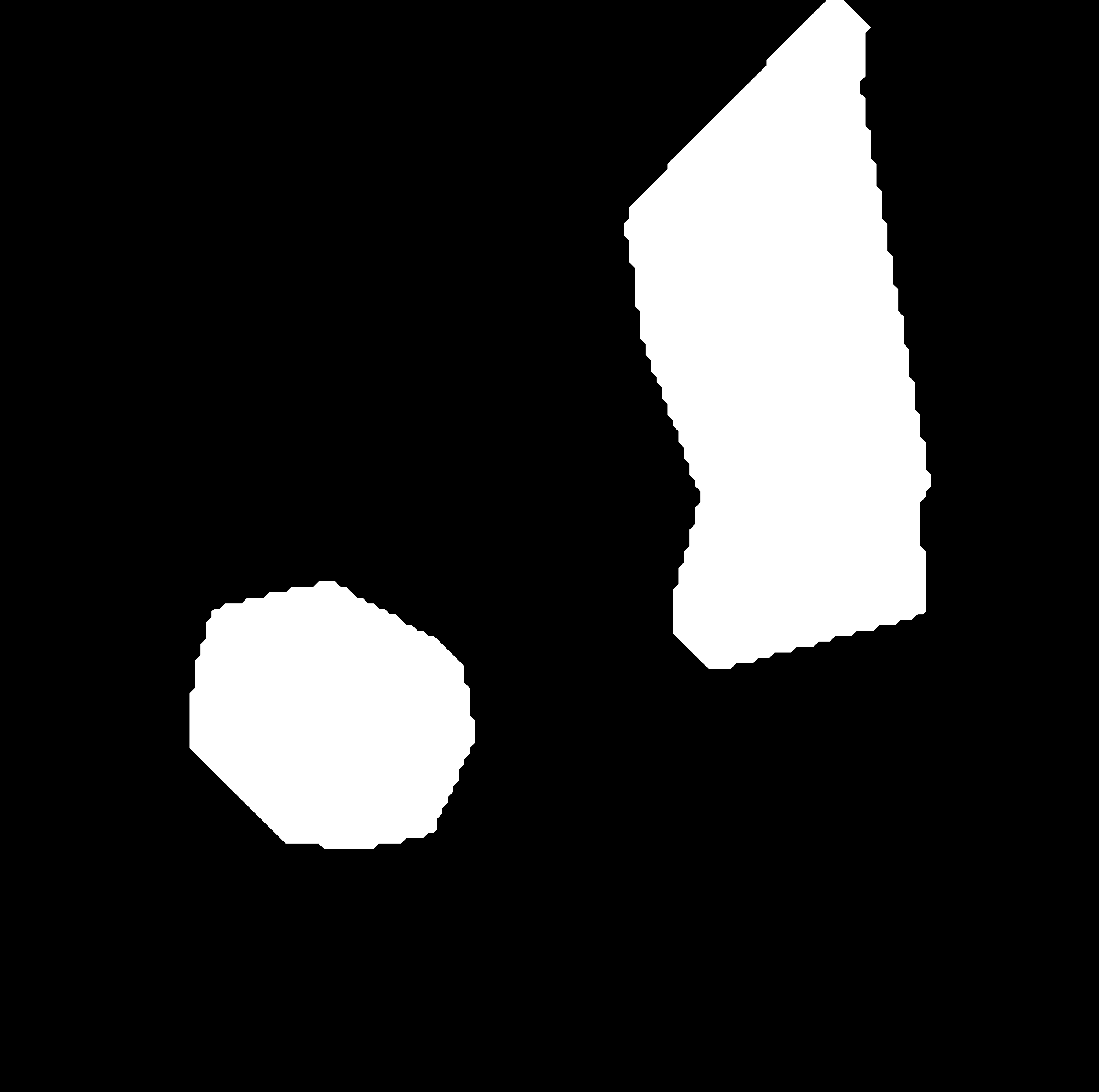}
            \end{tabular}
          }
        } 
        & \hspace{1.pt}
        \adjustbox{valign=c}{
          \makebox[0.3in][c]{
            \begin{tabular}{@{}c@{}}
              \small{94.70\%}\\[-0.em]
              \includegraphics[width=0.5in, angle=0]{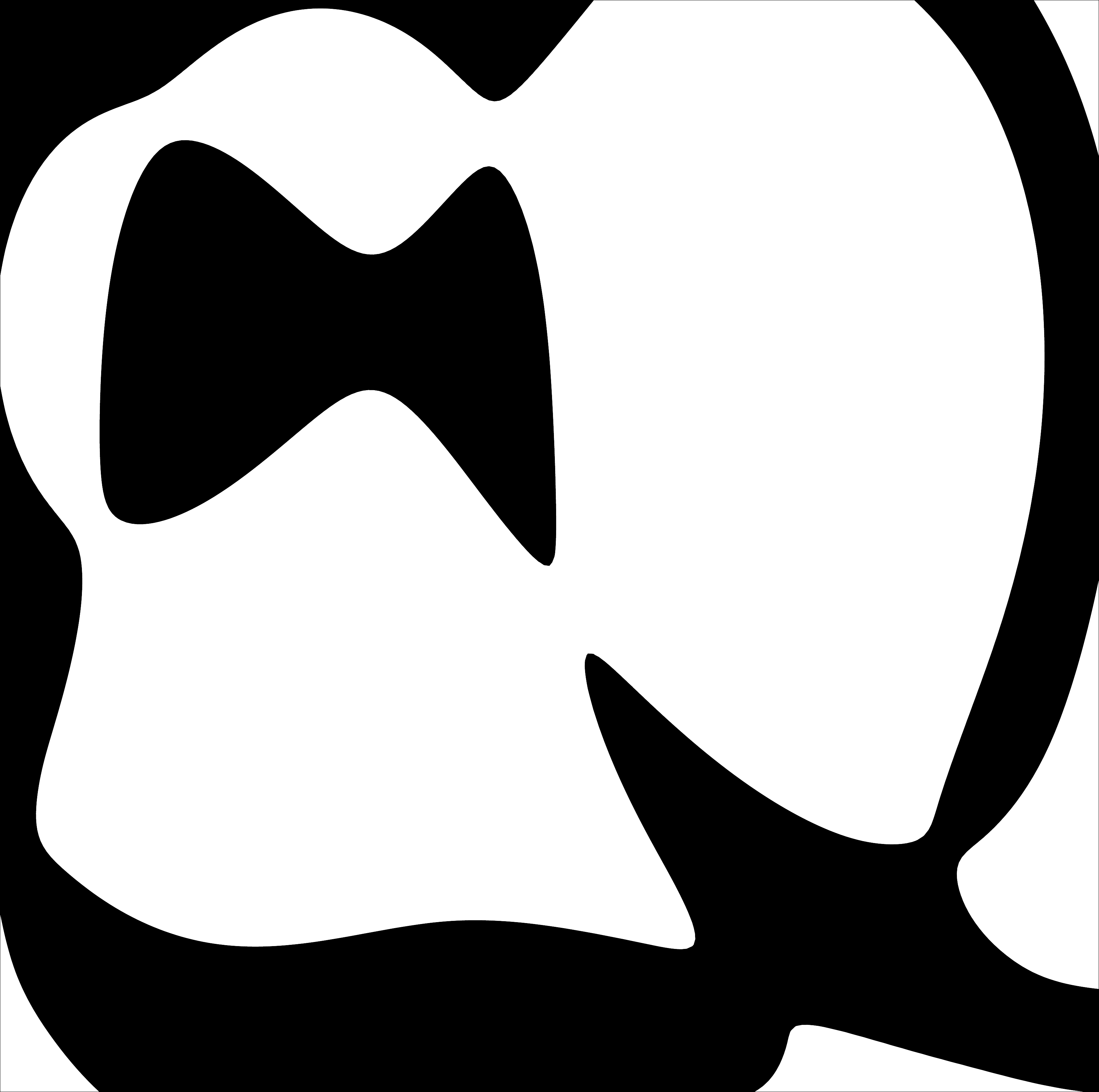}
            \end{tabular}
          }
        }
        & \hspace{1.pt}
        \adjustbox{valign=c}{
          \makebox[0.3in][c]{
            \begin{tabular}{@{}c@{}}
              \small{95.85\%}\\[-0.em]
              \includegraphics[width=0.5in, angle=0]{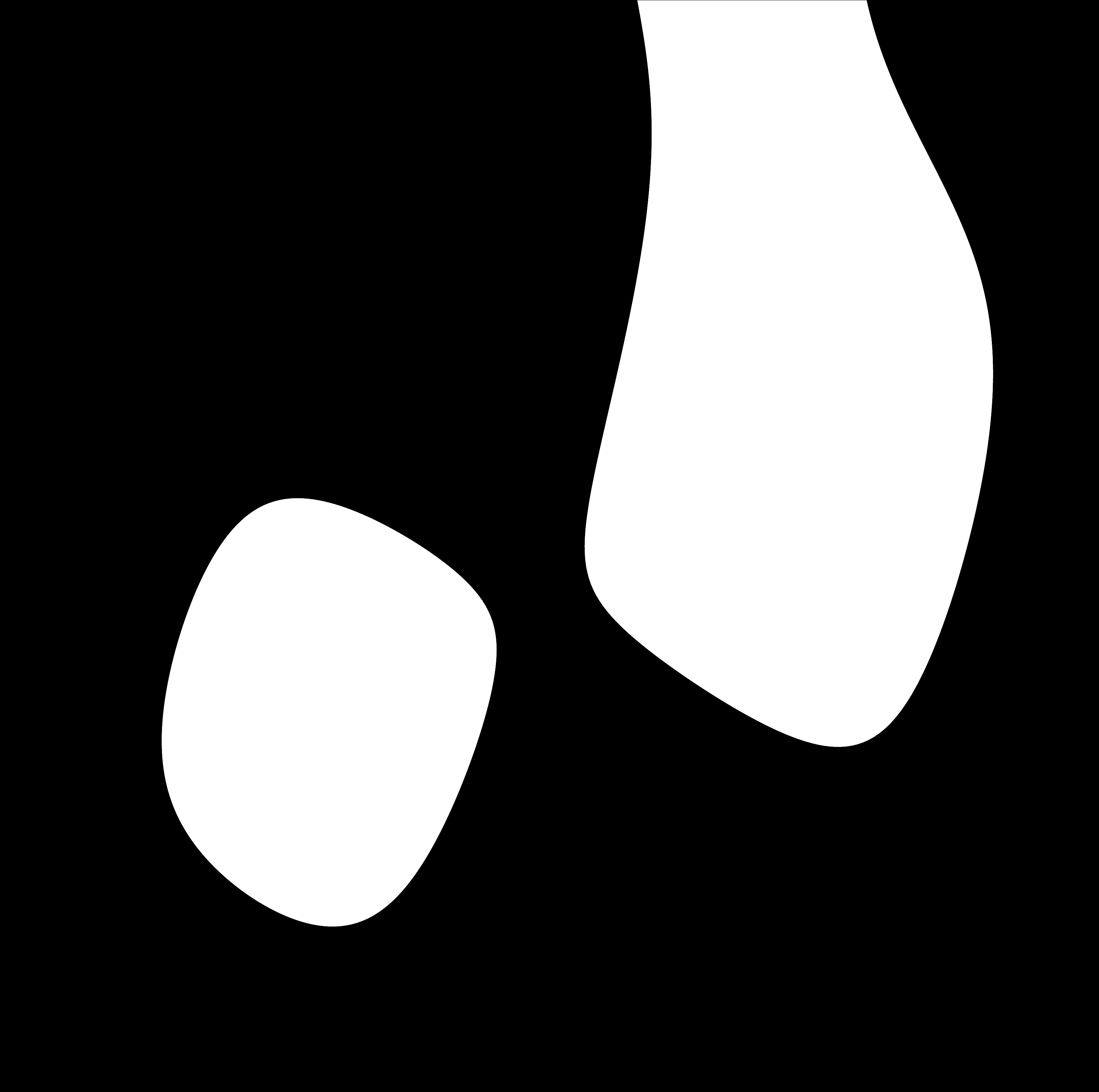}
            \end{tabular}
          }
        } 
        & \hspace{1.pt}
        \adjustbox{valign=c}{
          \makebox[0.3in][c]{
            \begin{tabular}{@{}c@{}}
              \small{\textbf{99.40\%}}\\[-0.em]
              \includegraphics[width=0.5in, angle=0]{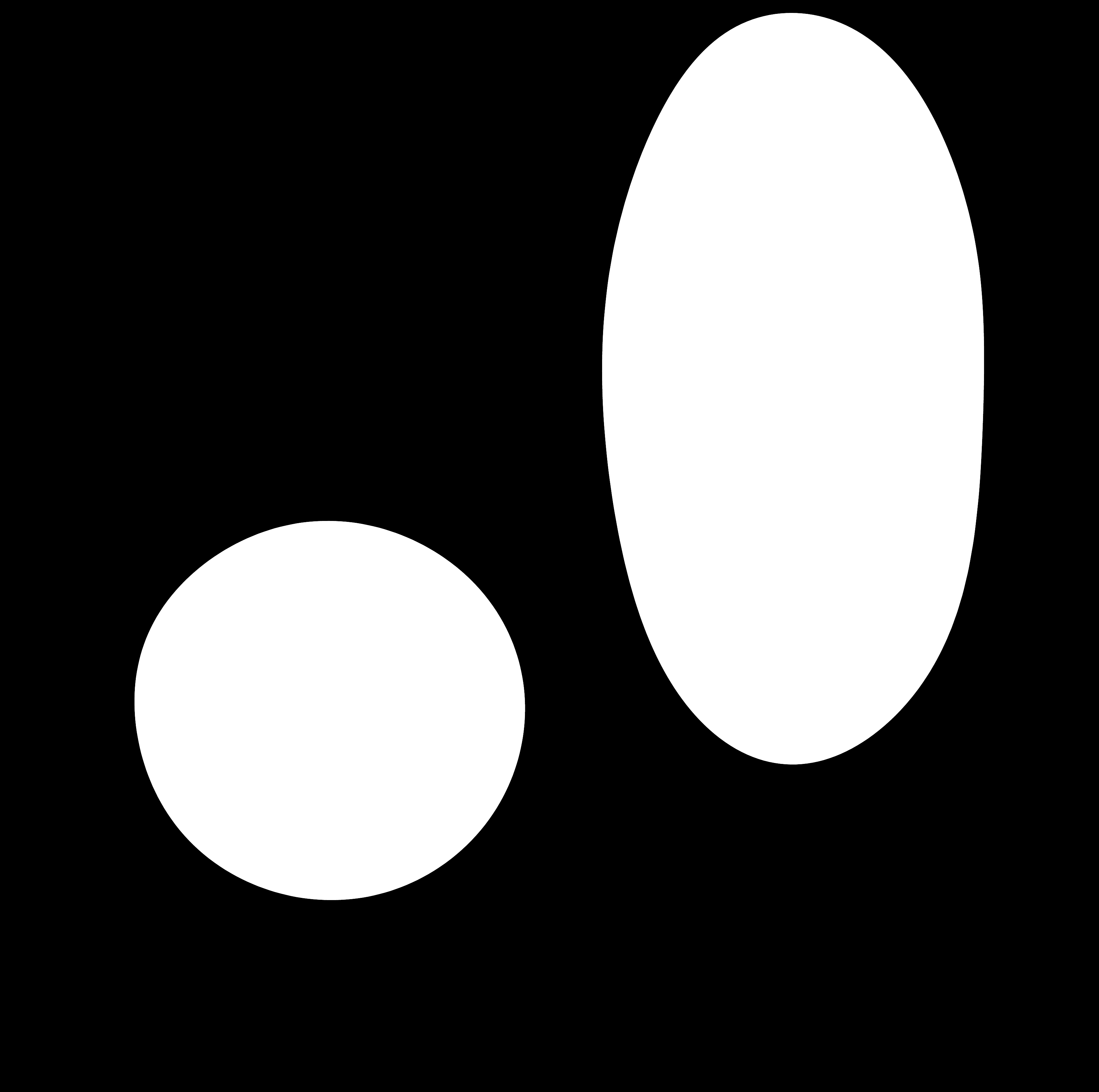}
            \end{tabular}
          }
        } \\[3.2em]
      
      \normalsize{$\text{P}_3$} &
        \adjustbox{valign=c}{
          \makebox[0.3in][c]{
            \begin{tabular}{@{}c@{}}
              \scriptsize{}\\[-0.em]
              \includegraphics[width=0.5in, angle=0]{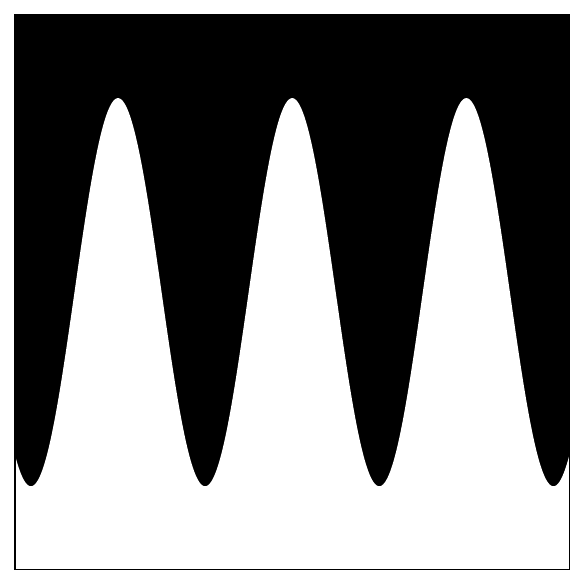}
            \end{tabular}
          }
        } \hspace{1.pt}
      &   \hspace{1.pt}
      \adjustbox{valign=c}{
          \makebox[0.3in][c]{
            \begin{tabular}{@{}c@{}}
              \small{86.37\%}\\[-0.em]
              \includegraphics[width=0.5in, angle=0]{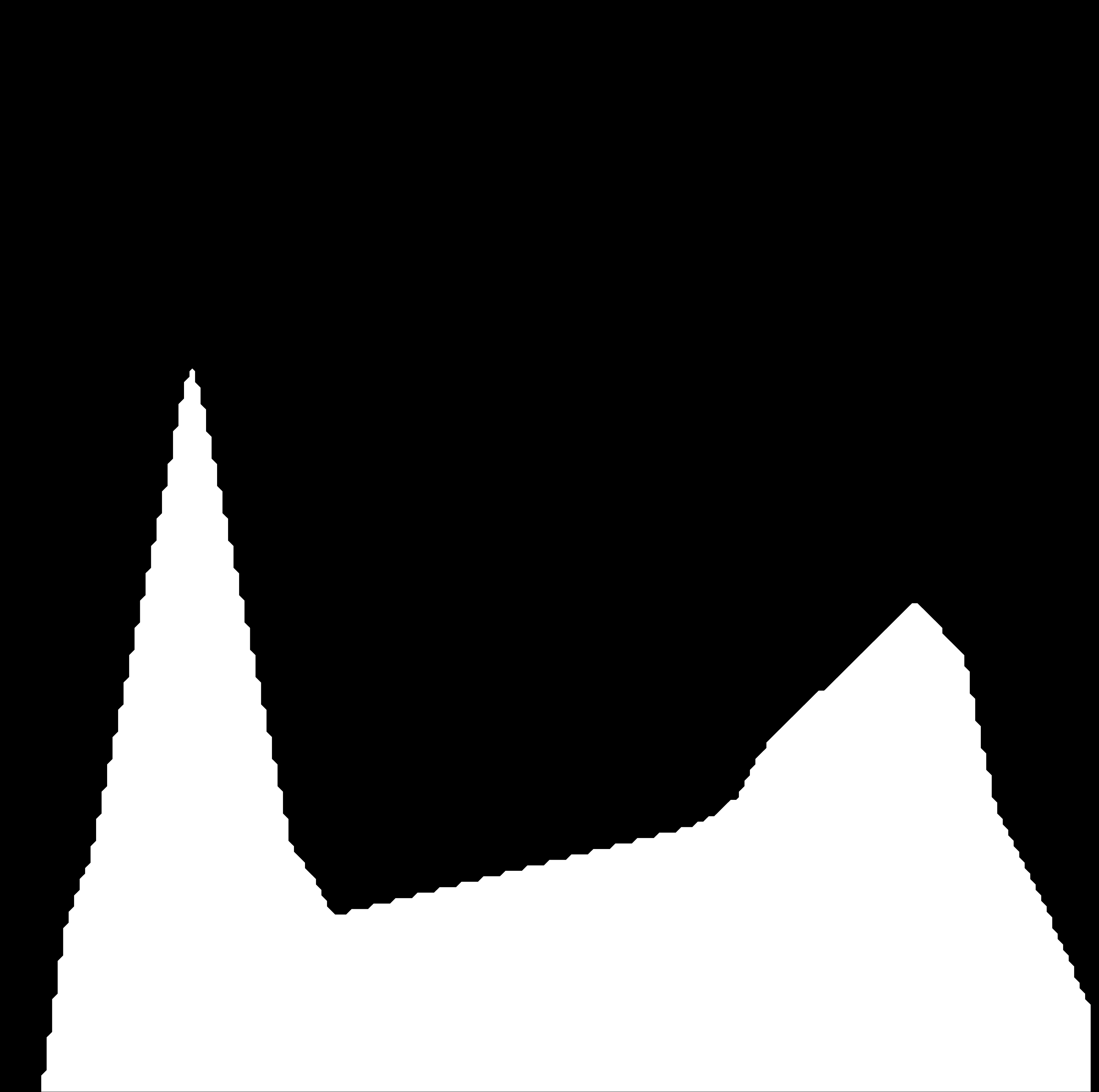}
            \end{tabular}
          }
        }
      & \hspace{1.pt}
      \adjustbox{valign=c}{
          \makebox[0.3in][c]{
            \begin{tabular}{@{}c@{}}
              \small{84.96\%}\\[-0.em]
              \includegraphics[width=0.5in, angle=0]{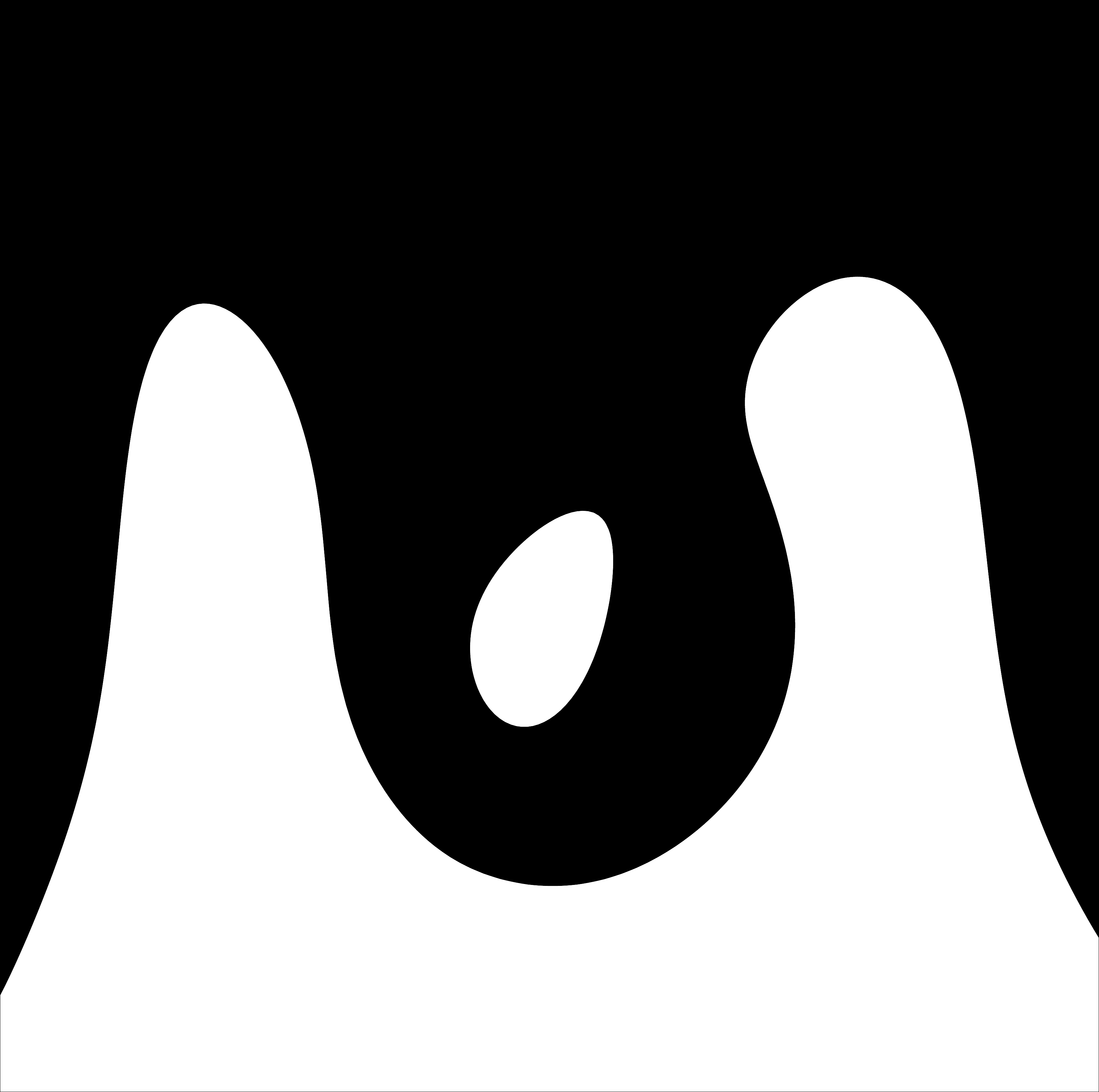}
            \end{tabular}
          }
        }
      & \hspace{1.pt}
      \adjustbox{valign=c}{
          \makebox[0.3in][c]{
            \begin{tabular}{@{}c@{}}
              \small{77.34\%}\\[-0.em]
              \includegraphics[width=0.5in, angle=0]{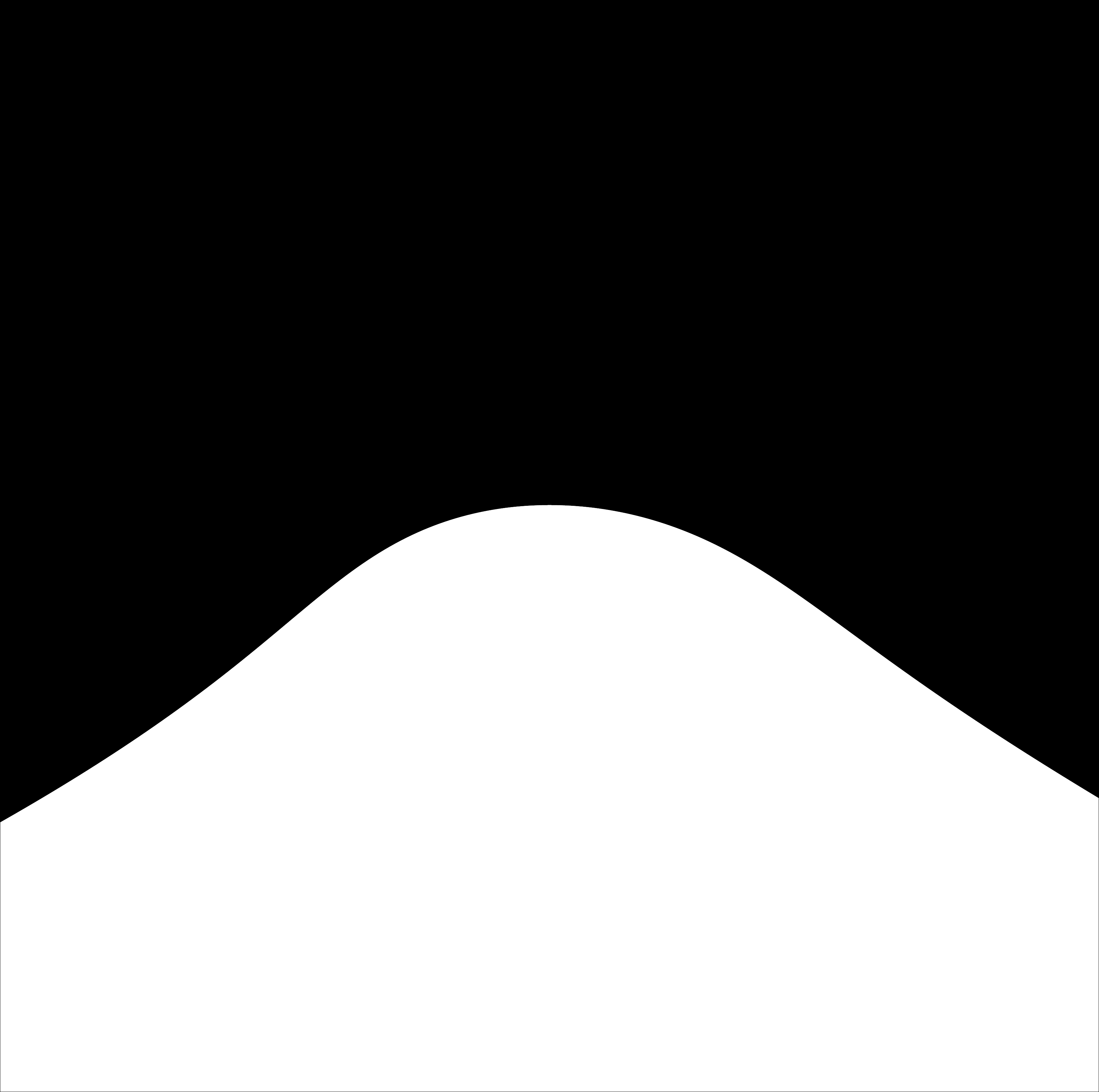}
            \end{tabular}
          }
        }
      & \hspace{1.pt}
      \adjustbox{valign=c}{
          \makebox[0.3in][c]{
            \begin{tabular}{@{}c@{}}
              \small{\textbf{98.60\%}}\\[-0em]
              \includegraphics[width=0.5in, angle=0]{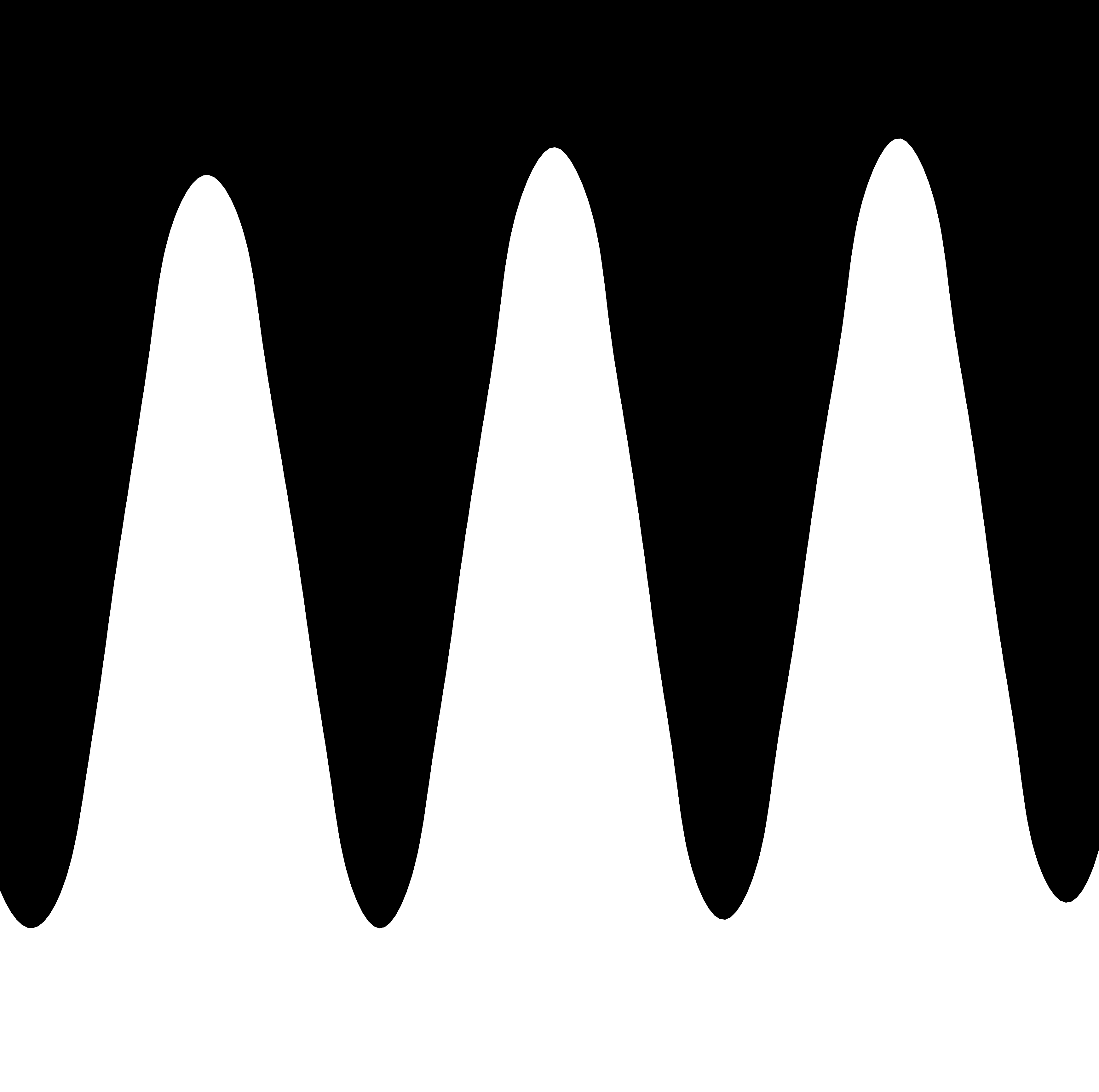}
            \end{tabular}
          }
        } \\[3.2em]
      
      \normalsize{$\text{P}_4$} &
        \adjustbox{valign=c}{
          \makebox[0.3in][c]{
            \begin{tabular}{@{}c@{}}
              \scriptsize{}\\[-0.em]
              \includegraphics[width=0.5in, angle=0]{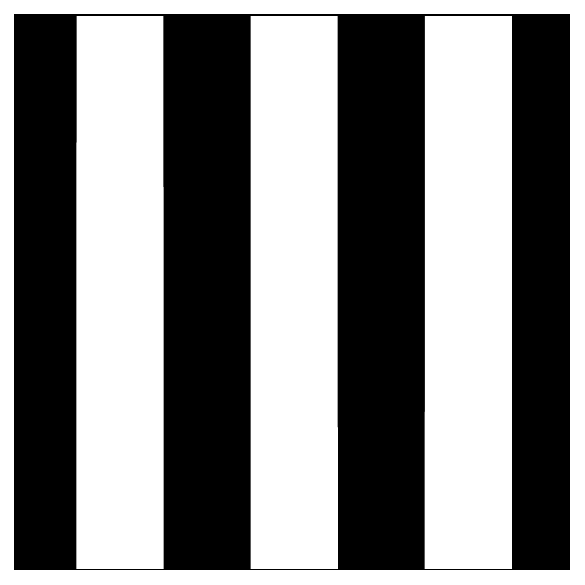}
            \end{tabular}
          }
        } \hspace{1.pt}
      &   \hspace{1.pt}
      \adjustbox{valign=c}{
          \makebox[0.3in][c]{
            \begin{tabular}{@{}c@{}}
              \small{96.49\%}\\[-0.em]
              \includegraphics[width=0.5in, angle=0]{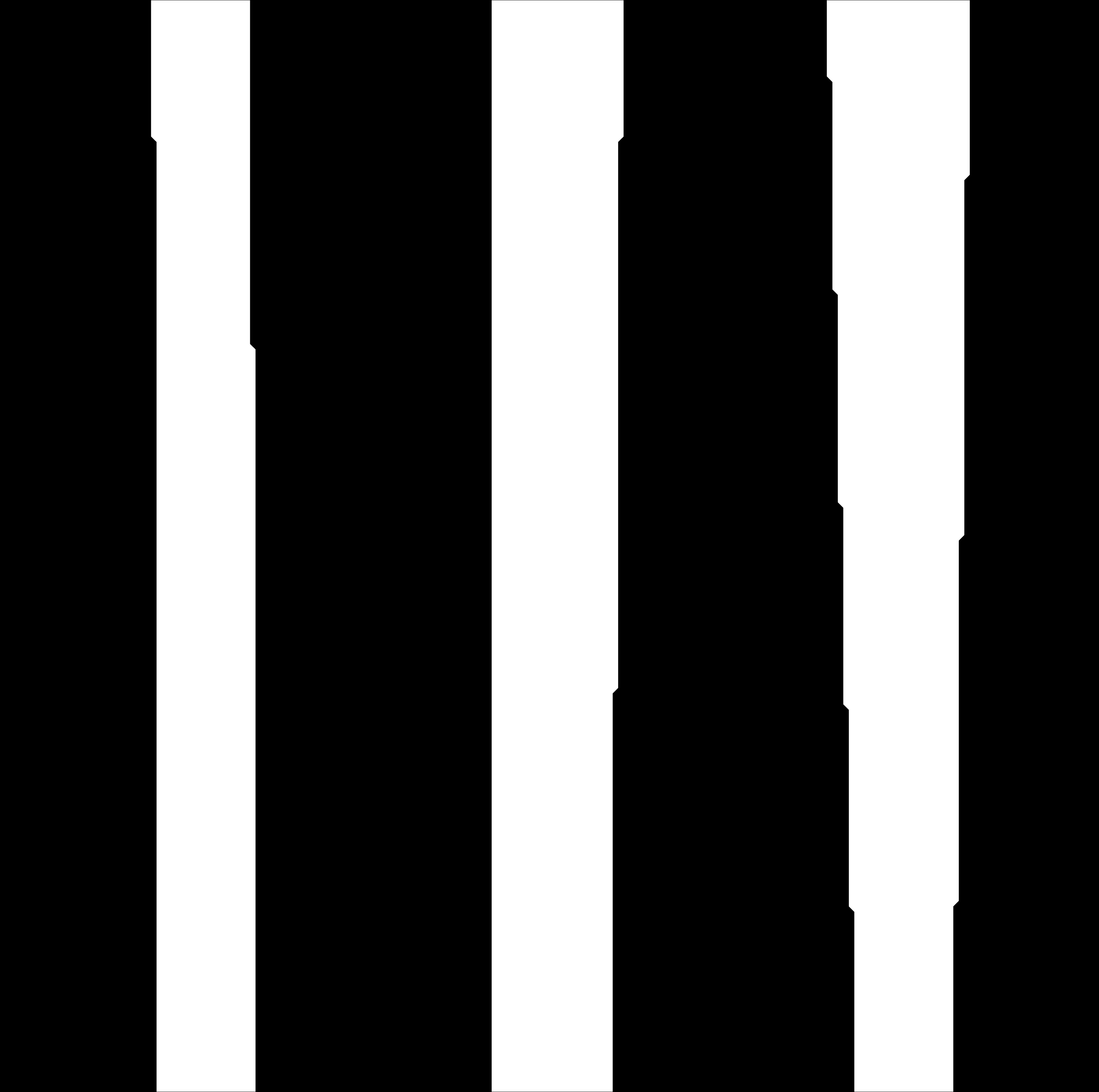}
            \end{tabular}
          }
        }
      & \hspace{1.pt}
      \adjustbox{valign=c}{
          \makebox[0.3in][c]{
            \begin{tabular}{@{}c@{}}
              \small{96.11\%}\\[-0.em]
              \includegraphics[width=0.5in, angle=0]{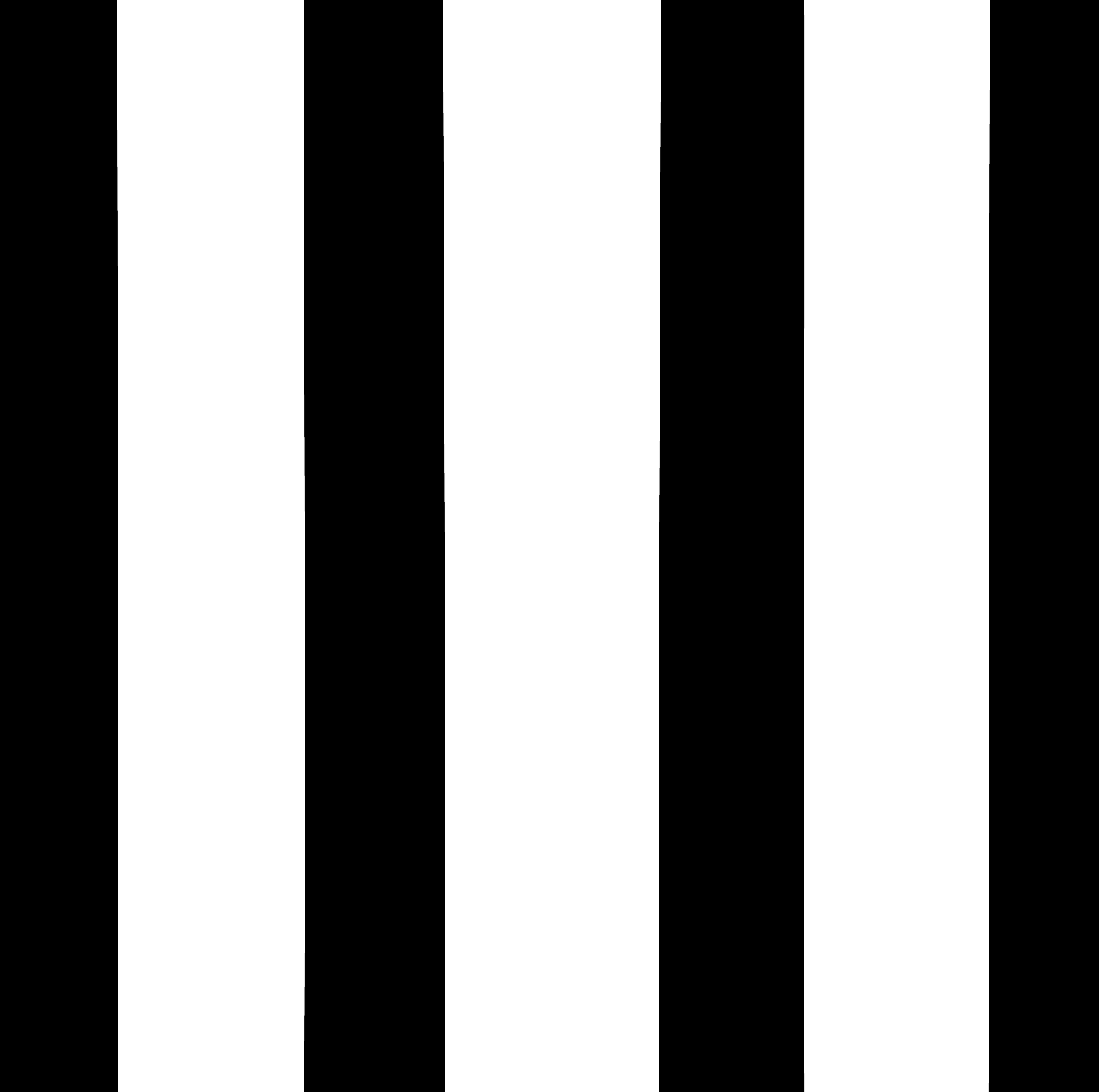}
            \end{tabular}
          }
        }
      & \hspace{1.pt}
      \adjustbox{valign=c}{
          \makebox[0.3in][c]{
            \begin{tabular}{@{}c@{}}
              \small{95.79\%}\\[-0.em]
              \includegraphics[width=0.5in, angle=0]{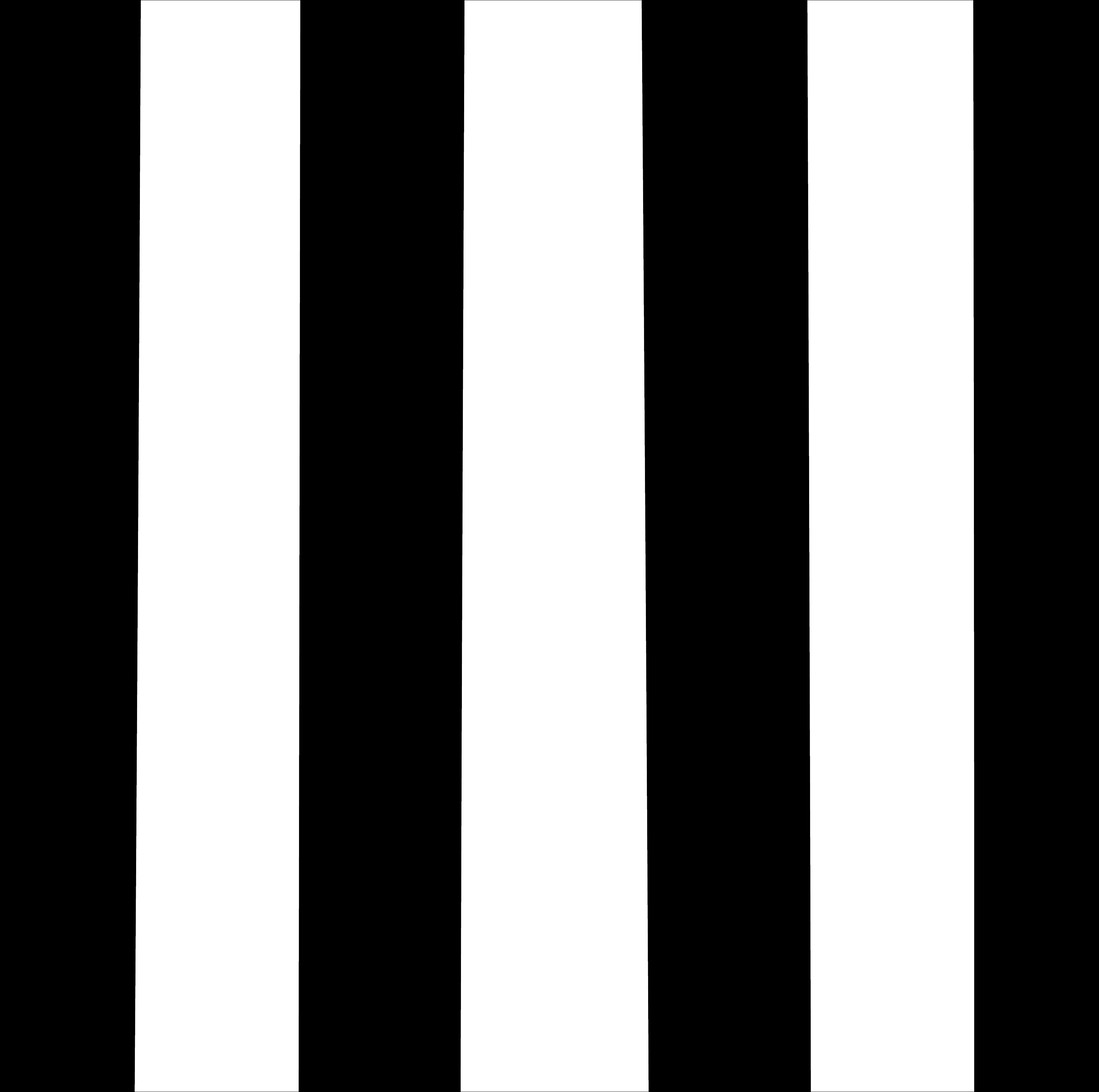}
            \end{tabular}
          }
        }
      & \hspace{1.pt}
      \adjustbox{valign=c}{
          \makebox[0.3in][c]{
            \begin{tabular}{@{}c@{}}
              \small{\textbf{99.62\%}}\\[-0.em]
              \includegraphics[width=0.5in, angle=0]{img/8_test_map_fourier.pdf}
            \end{tabular}
          }
        } \\[3.2em]
      
      \normalsize{$\text{P}_5$} &
        \adjustbox{valign=c}{
          \makebox[0.3in][c]{
            \begin{tabular}{@{}c@{}}
              \scriptsize{}\\[-0.em]
              \includegraphics[width=0.5in, angle=0]{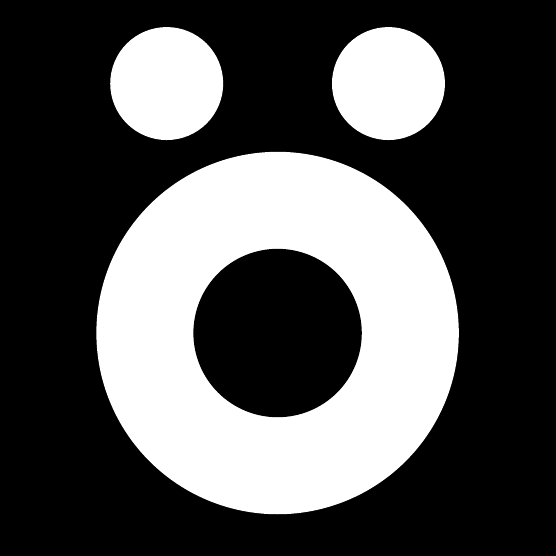}
            \end{tabular}
          }
        } \hspace{1.pt}
      &   \hspace{1.pt}
      \adjustbox{valign=c}{
          \makebox[0.3in][c]{
            \begin{tabular}{@{}c@{}}
              \small{{83.02\%}}\\[-0.em]
              \includegraphics[width=0.5in, angle=90]{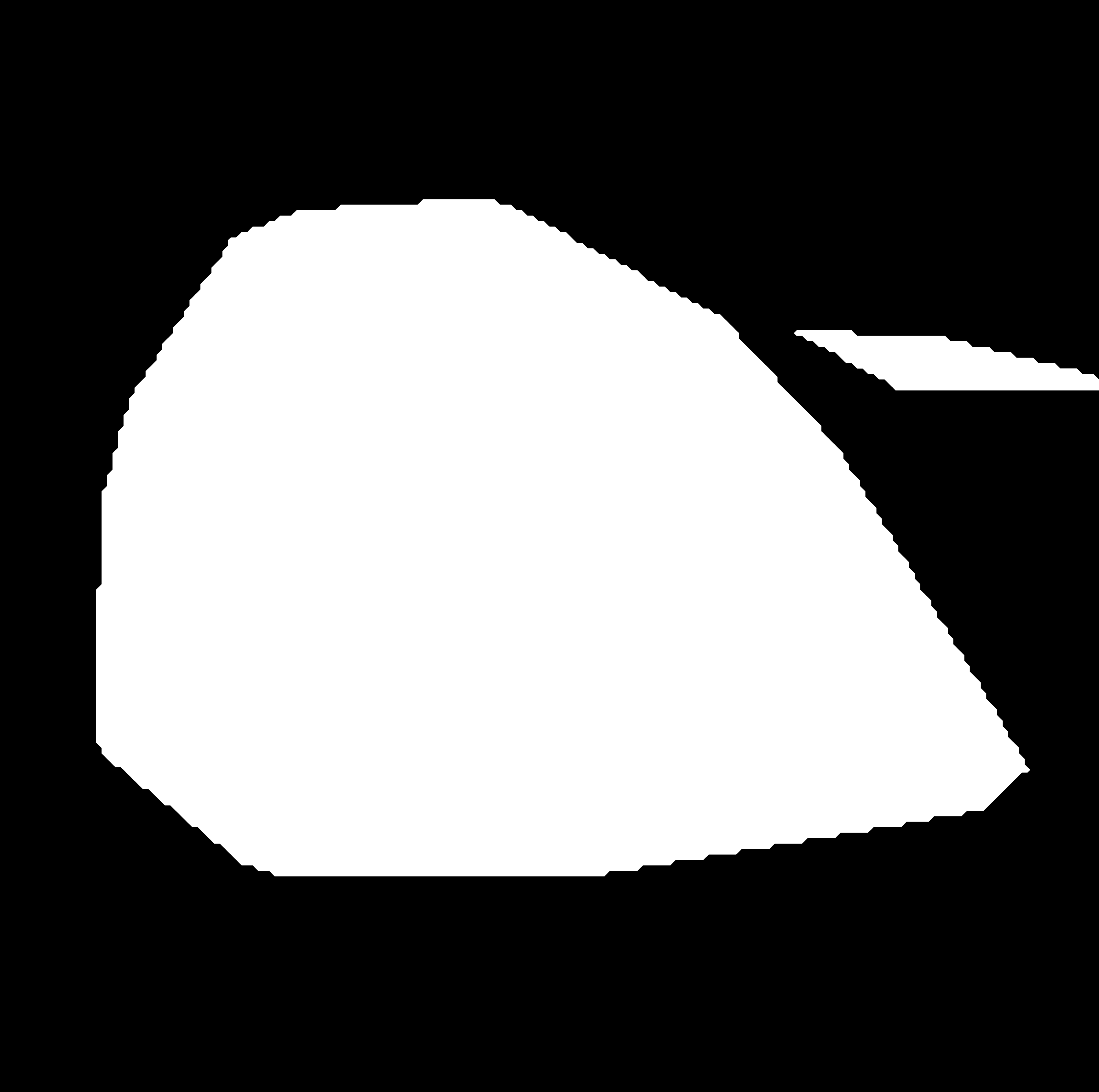}
            \end{tabular}
          }
        }
      & \hspace{1.pt}
      \adjustbox{valign=c}{
          \makebox[0.3in][c]{
            \begin{tabular}{@{}c@{}}
              \small{{92.24\%}}\\[-0.em]
              \includegraphics[width=0.5in, angle=90]{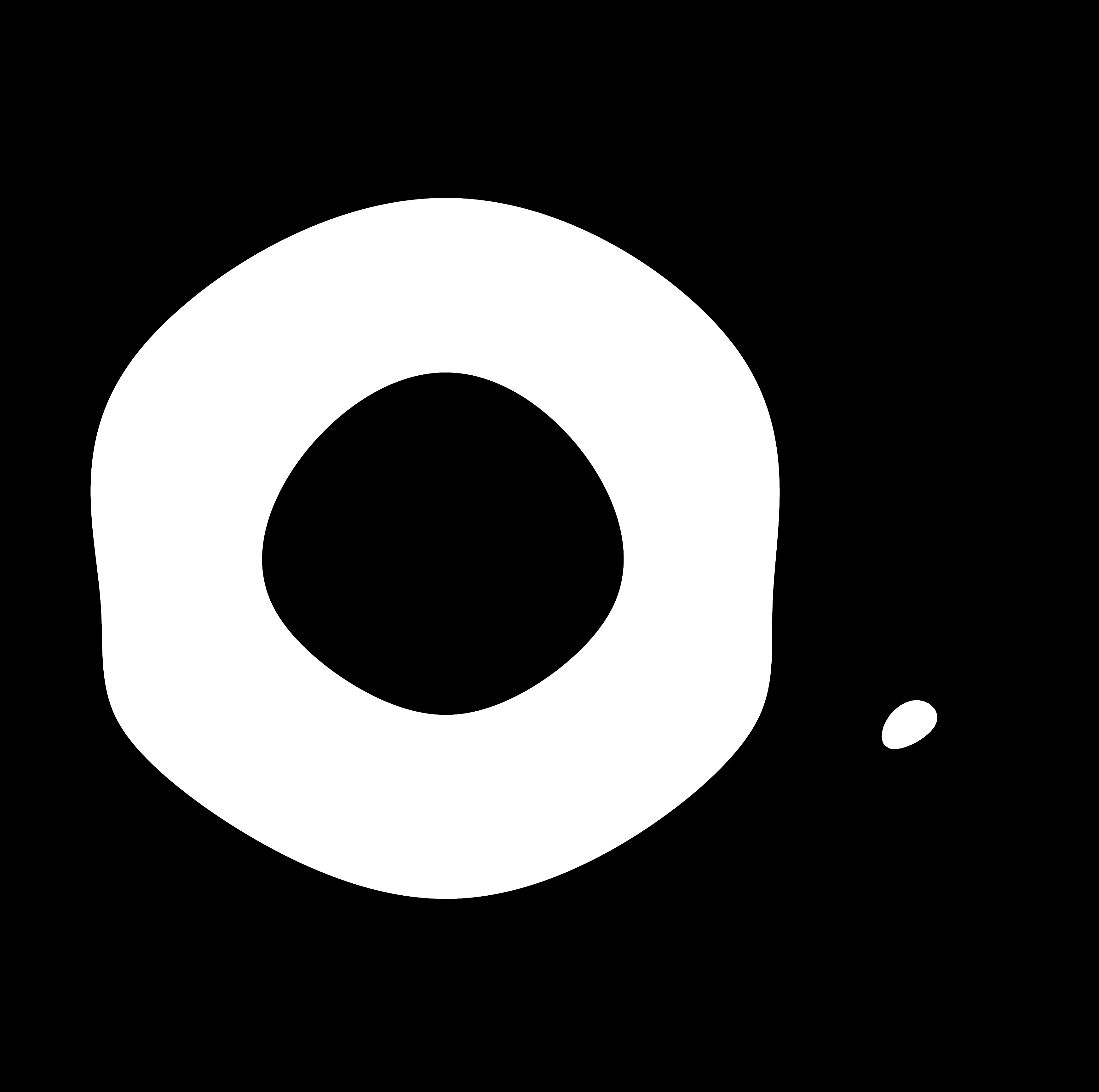}
            \end{tabular}
          }
        }
      & \hspace{1.pt}
      \adjustbox{valign=c}{
          \makebox[0.3in][c]{
            \begin{tabular}{@{}c@{}}
              \small{{89.42\%}}\\[-0.em]
              \includegraphics[width=0.5in, angle=90]{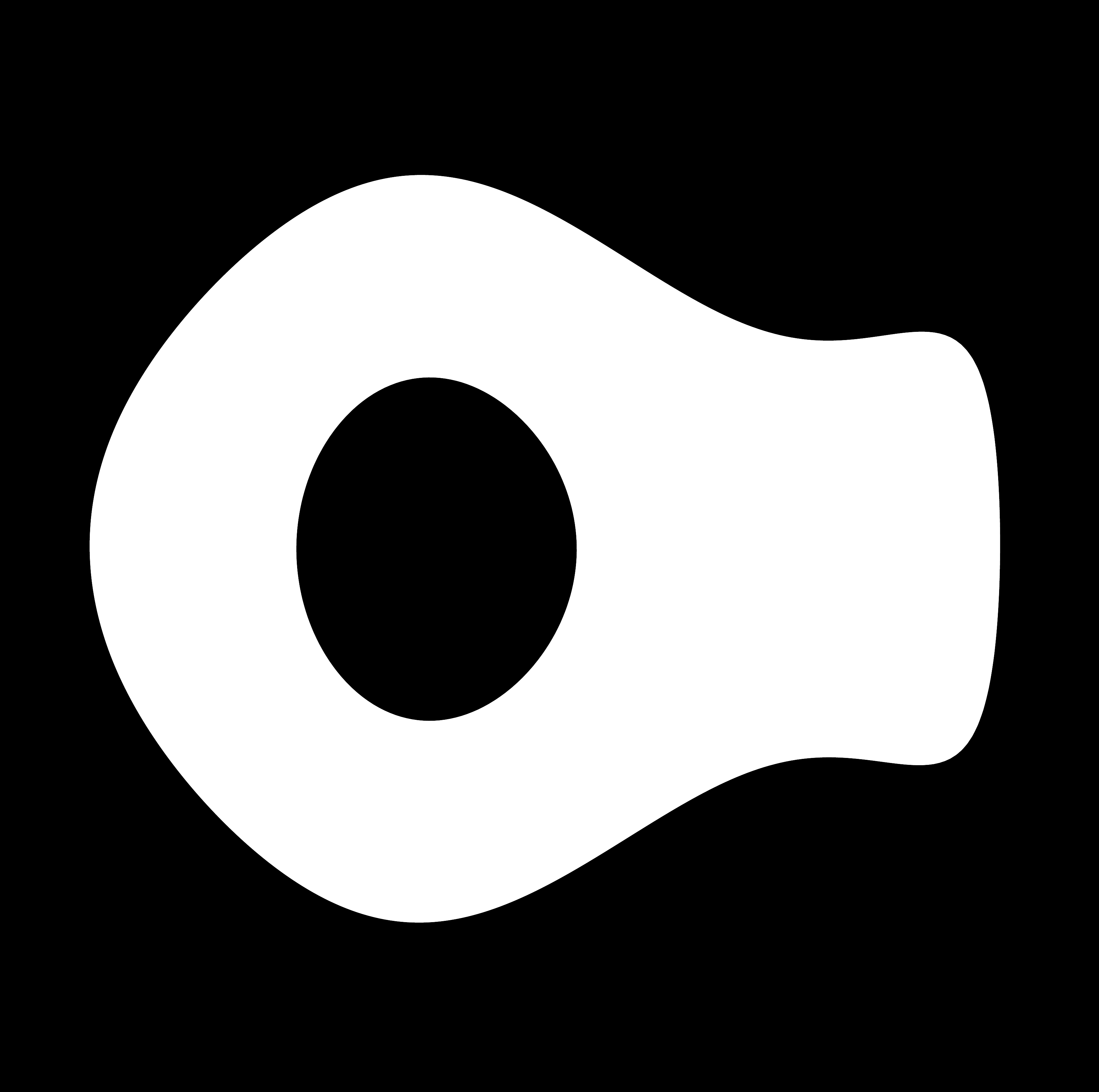}
            \end{tabular}
          }
        } 
      & \hspace{1.pt}
      \adjustbox{valign=c}{
          \makebox[0.3in][c]{
            \begin{tabular}{@{}c@{}}
              \small{\textbf{98.22\%}}\\[-0em]
              \includegraphics[width=0.5in, angle=90]{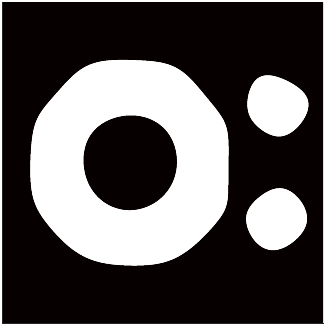}
            \end{tabular}
          }
        } \\
        \midrule
        & \small{Parameters} \hspace{2.pt} & \hspace{1.pt} \small{69} & \hspace{1.pt} \small{66} & \hspace{1.pt} \small{74} & \hspace{1.pt} \small{67}\\
      \bottomrule
    \end{tabular}
    \caption{Accuracy and decision boundary for different binary classification problems. The total number of trainable parameters for each model is also reported.}
    \label{tab:classification_results}
    \vspace{-0pt}
  \end{center}
\end{table}

\subsection{Results}
Table~\ref{tab:classification_results} summarizes the classification accuracies, decision boundaries and total number of trainable parameters for each model across five  binary classification problems. 
These results clearly demonstrate that the ENN achieves stable and high performance across all classification problems, whereas the benchmark models progressively deteriorate as the complexity of the decision regions increases. This consistent performance highlights the ENN’s ability to represent a wide variety of nonlinear boundaries efficiently.

One of the key reasons behind this robustness lies in the mathematical properties of the DCT-based AAF. The DCT gradient is well behaved (i.e., neither vanishing nor exploding) because cosine functions are bounded, and their derivatives are also bounded cosine functions. This ensures stable gradient propagation during training, allowing the ENN to maintain convergence even in highly nonlinear regimes. In contrast, the activation functions used in other models, such as the spline-based KAN, do not possess such well-characterized gradients, which often leads to unstable optimization and degraded performance.

A second important factor is the scalability of the parameterization with respect to the input dimension $M_0$, as seen in \eqref{eq:complexity}. By contrast, the Fourier model doubles its parameters due to the inclusion of both sine and cosine components for each frequency, while the KAN suffers even more severely, as its parameter count increases exponentially with the input dimension. As a result, when the network size must remain small, the spline functions in KAN are forced to use low-order polynomials and coarse resolutions, which severely limits their representational capacity.

In addition to its expressiveness, the ENN stands out for requiring a small number of hyperparameters. This is because the DCT resolution $N$ can be set to a high value without increasing the number of trainable parameters, and the number of DCT coefficients can be fixed at $Q=6$. Empirically, this choice of $Q$ has been observed to faithfully represent a wide range of univariate functions, ensuring that each adaptive activation function remains flexible without inflating the parameter count.

Overall, these findings confirm that the ENN not only offers a more compact and interpretable parameterization but also maintains stable gradients and scalable complexity, making it particularly well suited for modeling complex decision boundaries under constrained model sizes.

\begin{figure}[t]
    \centering
    \begin{subfigure}[b]{0.11\textwidth}
        \centering
        \includegraphics[width=\textwidth, height=58pt]{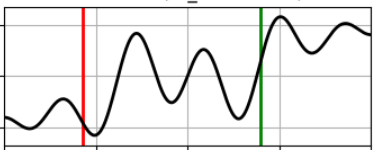}
        \caption{Neuron 1.}
        \label{fig:aaf1}
    \end{subfigure}
    \hfill
    \begin{subfigure}[b]{0.11\textwidth}
        \centering
        \includegraphics[width=\textwidth, height=58pt]{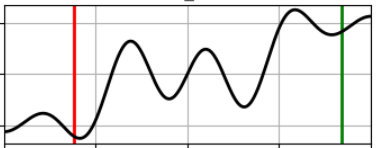}
        \caption{Neuron 2.}
        \label{fig:aaf2}
    \end{subfigure}
    \hfill
    \begin{subfigure}[b]{0.11\textwidth}
        \centering
        \includegraphics[width=\textwidth, height=58pt]{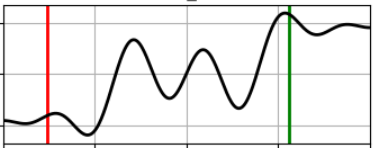}
        \caption{Neuron 3.}
        \label{fig:aaf3}
    \end{subfigure}
    \hfill
    \begin{subfigure}[b]{0.11\textwidth}
        \centering
        \includegraphics[width=\textwidth, height=58pt]{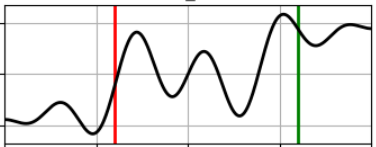}
        \caption{Neuron 4.}
        \label{fig:aaf4}
    \end{subfigure}

    \vspace{0.3cm}

    \begin{subfigure}[b]{0.11\textwidth}
        \centering
        \includegraphics[width=\textwidth, height=58pt]{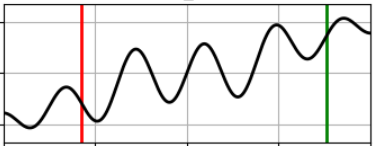}
        \caption{Neuron 5.}
        \label{fig:aaf5}
    \end{subfigure}
    \hfill
    \begin{subfigure}[b]{0.11\textwidth}
        \centering
        \includegraphics[width=\textwidth, height=58pt]{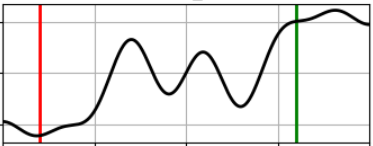}
        \caption{Neuron 6.}
        \label{fig:aaf6}
    \end{subfigure}
    \hfill
    \begin{subfigure}[b]{0.11\textwidth}
        \centering
        \includegraphics[width=\textwidth, height=58pt]{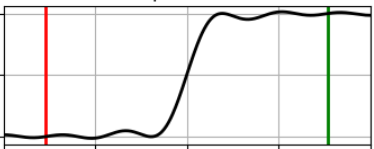}
        \caption{Output.}
        \label{fig:aaf_out}
    \end{subfigure}
    \hfill
    \begin{subfigure}[b]{0.11\textwidth}
        \centering
        \includegraphics[width=\textwidth]{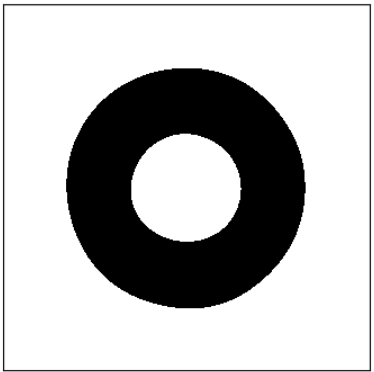}
        \caption{Map.}
        \label{fig:map}
    \end{subfigure}

    \caption{\gls{AAF} for the 6 hidden and output neurons in a binary classification task. Vertical lines indicate the input range where each neuron operates. The last figure shows the learned decision boundary.}
    \label{fig:aaf_ring}
\end{figure}

\subsection{Interpretable Representations}

Figure~\ref{fig:aaf_ring} illustrates the AAF learned by the ENN for problem $\text{P}_1$. The vertical lines indicate the range of input values effectively used by each neuron, which is particularly relevant since the DCT basis is inherently periodic and defined over an infinite domain. Interestingly, all AAF in the hidden layer have converged to the same representation, reflecting the radial symmetry of problem $\text{P}_1$.
Of particular interest is the AAF in the output layer, which has learned a sigmoidal shape. This emerges naturally from the training, as the output must map continuous inputs into the two target classes, $-1$ and $1$.
Notably, the ENN is the only model among the benchmarks that allows learning the output activation function. In all other models, the output nonlinearity is fixed a priori to a sigmoid. This ability to adaptively learn the output activation is an additional advantage of the proposed ENN, contributing to its expressiveness and flexibility.

\begin{figure}[t]
    \centering
    \begin{subfigure}[b]{0.15\textwidth}
        \centering
        \includegraphics[width=\textwidth]{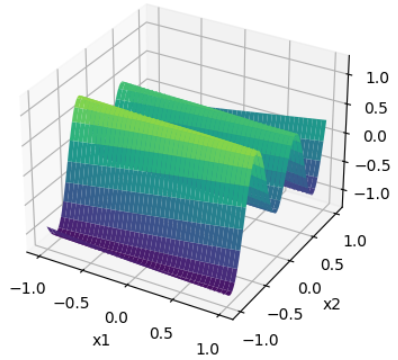}
        \caption{Neuron 1.}
        \label{fig:bump1}
    \end{subfigure}
    \hfill
    \begin{subfigure}[b]{0.15\textwidth}
        \centering
        \includegraphics[width=\textwidth]{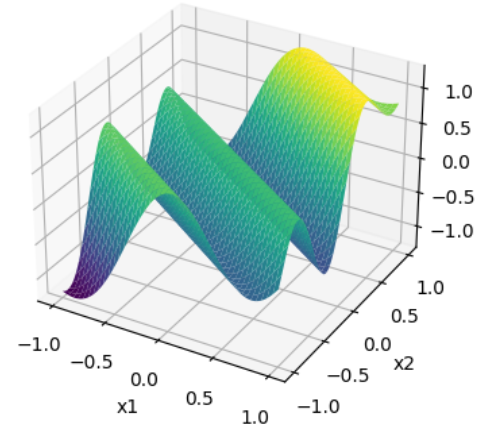}
        \caption{Neuron 2.}
        \label{fig:bump2}
    \end{subfigure}
    \hfill
    \begin{subfigure}[b]{0.15\textwidth}
        \centering
        \includegraphics[width=\textwidth]{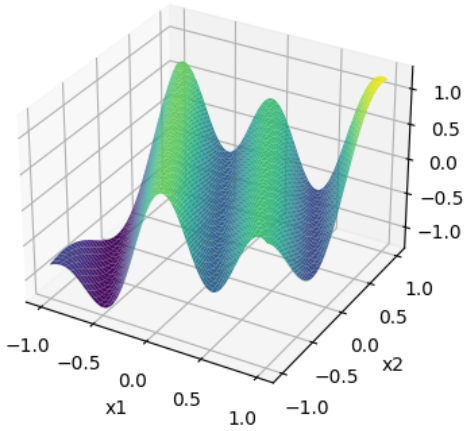}
        \caption{Neuron 3.}
        \label{fig:bump3}
    \end{subfigure}
    
    \vspace{0.3cm}

    \begin{subfigure}[b]{0.15\textwidth}
        \centering
        \includegraphics[width=\textwidth]{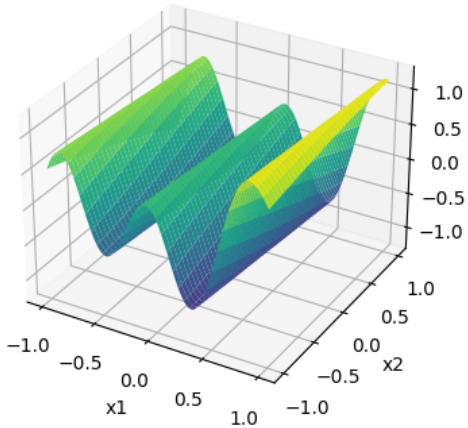}
        \caption{Neuron 4.}
        \label{fig:bump4}
    \end{subfigure}
    \hfill
    \begin{subfigure}[b]{0.15\textwidth}
        \centering
        \includegraphics[width=\textwidth]{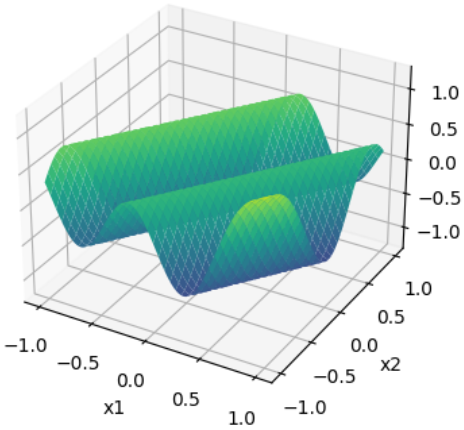}
        \caption{Neuron 5.}
        \label{fig:bump5}
    \end{subfigure}
    \hfill
    \begin{subfigure}[b]{0.15\textwidth}
        \centering
        \includegraphics[width=\textwidth]{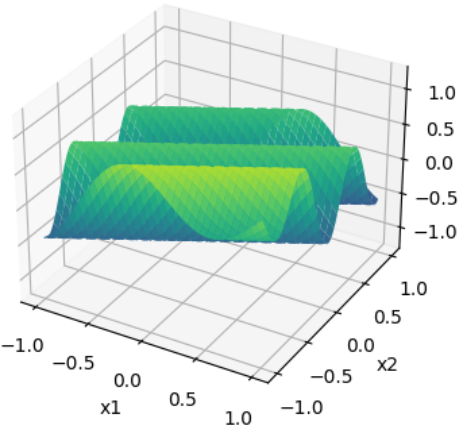}
        \caption{Neuron 6.}
        \label{fig:bump6}
    \end{subfigure}

    \caption{Bumps of the hidden neurons from Figure \ref{fig:aaf_ring}.}
    \label{fig:bumps_ring}
\end{figure}

A key concept for understanding the ENN is the notion of the bump, which refers to the response of a neuron across the input space. For a two-dimensional input, the bump represents the neuron’s output over all possible input combinations, effectively visualizing its contribution to the overall model. Figure \ref{fig:bumps_ring} illustrates the bumps of each hidden neuron from the problem $\text{P}_1$.
The bump provides valuable insight into the role of each neuron and its parameters in shaping the network’s expressiveness. Specifically, the nonlinear activation function determines the shape and complexity of the bump, while the linear weights control its orientation within the input space. This decomposition highlights why developing \gls{AAF} is so crucial: the expressiveness and flexibility of each neuron, and ultimately the network as a whole, are largely governed by the nonlinear stage of processing.

To generate the binary ring map in Figure \ref{fig:aaf_ring}(\subref{fig:map}), the \gls{ENN} produces \glspl{AAF} that oscillate to form two peaks. Since the decision map is rotationally symmetric, all \glspl{AAF} have equivalent shapes. The \gls{ENN} leverages this by orienting each bump in a different spatial direction, allowing each neuron to contribute uniquely. This diversity in bump orientation is particularly advantageous for modeling the ring structure, as more neurons provide more angular coverage, enabling the network to form a finer and more precise ring.

\begin{figure}[t]
    \centering
     \hspace{-2pt}
     $\text{Accuracy}=99.3\%$
     \hspace{2pt}
     $\text{Accuracy}=97.7\%$
     \hspace{2pt}
     $\text{Accuracy}=86.4\%$\\
    \begin{subfigure}[b]{0.15\textwidth}
        \centering
        \includegraphics[width=\textwidth]{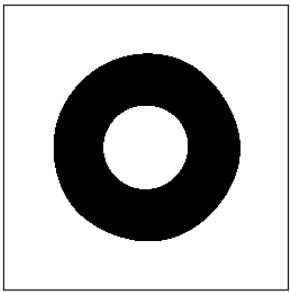}
        \caption{$M_1=8$.}
        \label{fig:width8}
    \end{subfigure}
    \hfill
    \begin{subfigure}[b]{0.15\textwidth}
        \centering
        \includegraphics[width=\textwidth]{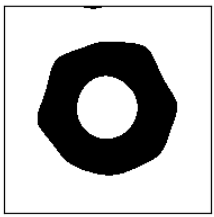}
        \caption{$M_1=4$.}
        \label{fig:width4}
    \end{subfigure}
    \hfill
    \begin{subfigure}[b]{0.15\textwidth}
        \centering
        \includegraphics[width=\textwidth]{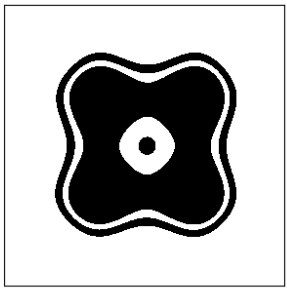}
        \caption{$M_1=2$.}
        \label{fig:width2}
    \end{subfigure}

    \caption{Decision map of the \gls{ENN} for different values of $M_1$.}
    \label{fig:width_ring}
\end{figure}

An important question is how the number of hidden neurons affects the model. This behavior is clearly interpretable and illustrated in Figure \ref{fig:width_ring}. As expected, increasing the number of neurons leads to diminishing returns, with the map’s accuracy eventually saturating. Decreasing the width to 4 neurons, the ENN can form responses in up to 4 distinct directions, resulting in a decision map that resembles an octagon.
When using exactly two neurons, the ENN constructs the \glspl{AAF} with two peaks and arranges them orthogonally in space. This results in a square-shaped ring, essentially the best approximation of a circular ring that can be achieved using only two directional responses.

In all previous experiments, the number of DCT coefficients was fixed to $Q=6$. As illustrated in Figure \ref{fig:aaf_ring}, the resulting AAF are smooth, indicating that high-frequency coefficients contribute minimally to the overall shape of the activation. This frequency perspective provided by the DCT makes the neuron’s expressiveness interpretable in terms of complexity. Notably, even when reducing the number of coefficients to $Q=3$, the ENN is still able to generate the ring-shaped decision map with a competitive accuracy of $97.0\%$, highlighting the efficiency and interpretability of the representation.


\section{Regression on Images}

Having discussed the performance of the shallow ENN, we now turn to a more complex task that requires deeper and wider architectures. Specifically, we consider an ENN for INR of an image. This task involves approximating the underlying continuous signal of the image by learning a mapping from pixel coordinates to corresponding pixel values. 

\subsection{Training Setup and Benchmarks}

The size of the dataset $D$ corresponds to the number of pixels in the image, which is $256\times256=65536$ pixels.
The ENN consists of 4 hidden layers, each containing 240 neurons, and uses $Q=6$ coefficients per AAF with $N=512$. This results in $180.247$ trainable parameters.

The benchmark models have the same number of layers and are configured to incur in the same parameter count:
\begin{itemize}
    \item \textbf{ReLU:} The model has $M_\ell=256$ neurons per layer.
    \item \textbf{Fourier:} To preserve the structural similarity between the ENN and Fourier, the number of Fourier coefficients is kept at $Q=6$. Due to the additional number of parameters, the number of neurons per layer is set to $M_\ell=235$.
    \item \textbf{KAN:} The model has $M_\ell=82$ neurons, each spline is cubic ($k=3$) with $G=5$ grid intervals.
    \item \textbf{SIREN:} This model has fixed sine activation functions. We include it due to its state-of-the-art performance in INR \cite{sitzmann2020siren}. SIREN contains $M_\ell=256$ neurons in al layers and the frequency scaling hyperparameter is fixed to $\omega = 30$.
\end{itemize}

This time, models are trained with full gradient descent (GD), this is, all training samples are fed simultaneously to the network. This is because the number of pixels in the image are relatively small and it guarantees a better estimate of the gradient. All models are trained for 300 epochs, this is, the number of times the whole dataset is fed to the network. As for the learning rate, to ensure a training as comparable as possible, all the models use a learning rate of $10^{-3}$ to update their linear weights, while for the models that have AAF, the learning rate of their activations is fixed to $10^{-2}$. The ADAM optimizer is employed for all models. This choice is motivated by its proven effectiveness across all benchmarks, even though it implies that we are not strictly applying the previously discussed optimization rules to the ENN. Nevertheless, the ENN demonstrates sufficient versatility to converge successfully even with this standard optimizer. It is important to note that while this configuration has shown to work reasonably well across all models, fully optimizing the hyperparameter settings for each benchmark model falls outside the scope of this work.

\subsection{Results}

\begin{figure}[t]
    \centering
    \begin{subfigure}[b]{0.24\columnwidth}
        \centering
        \includegraphics[width=\textwidth]{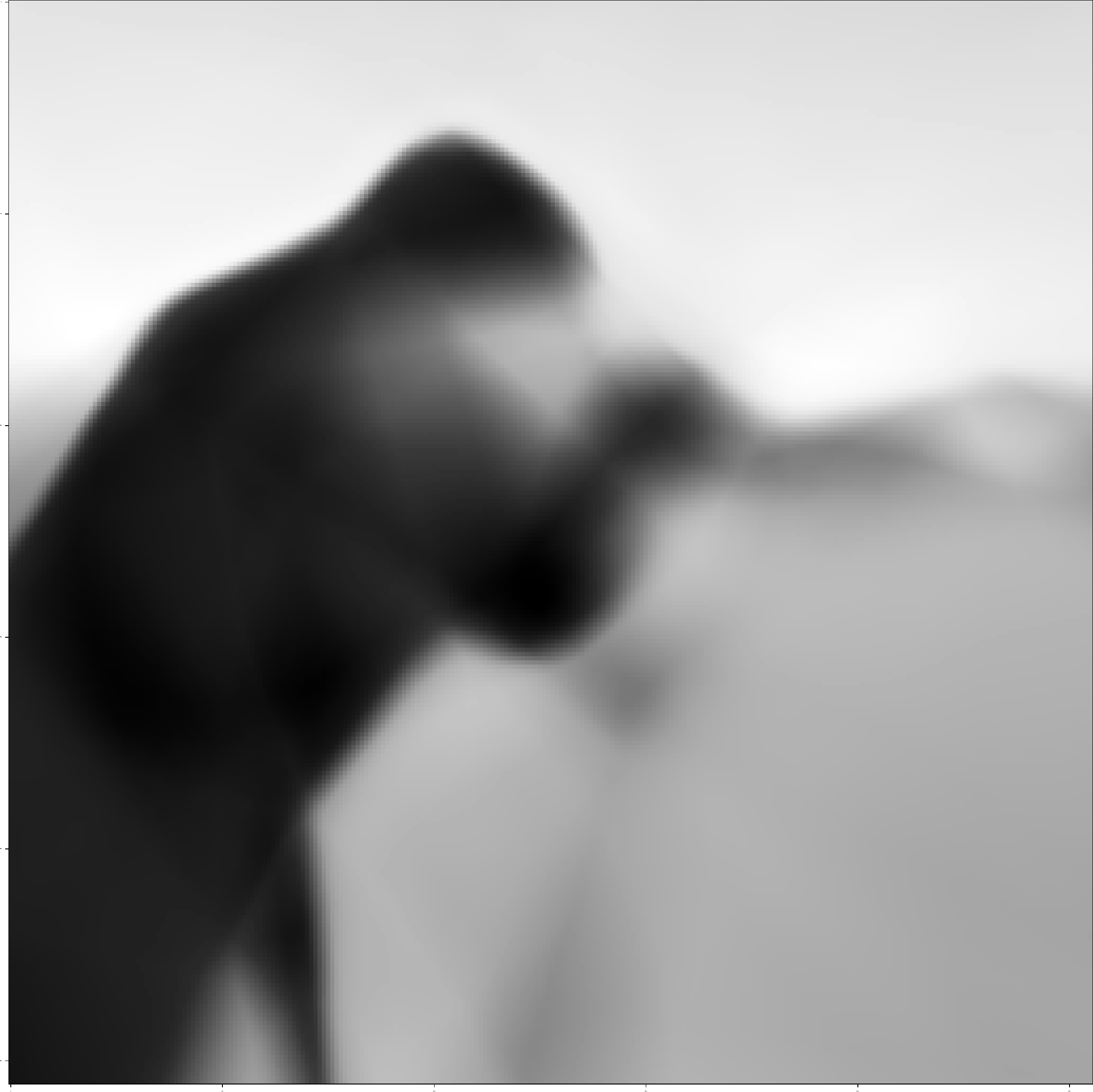}
        \caption{ReLU.}
        \label{fig:mlp_basic_cameraman_prediction}
    \end{subfigure}
    \hspace{20pt}
    \begin{subfigure}[b]{0.24\columnwidth}
        \centering
        \includegraphics[width=\textwidth]{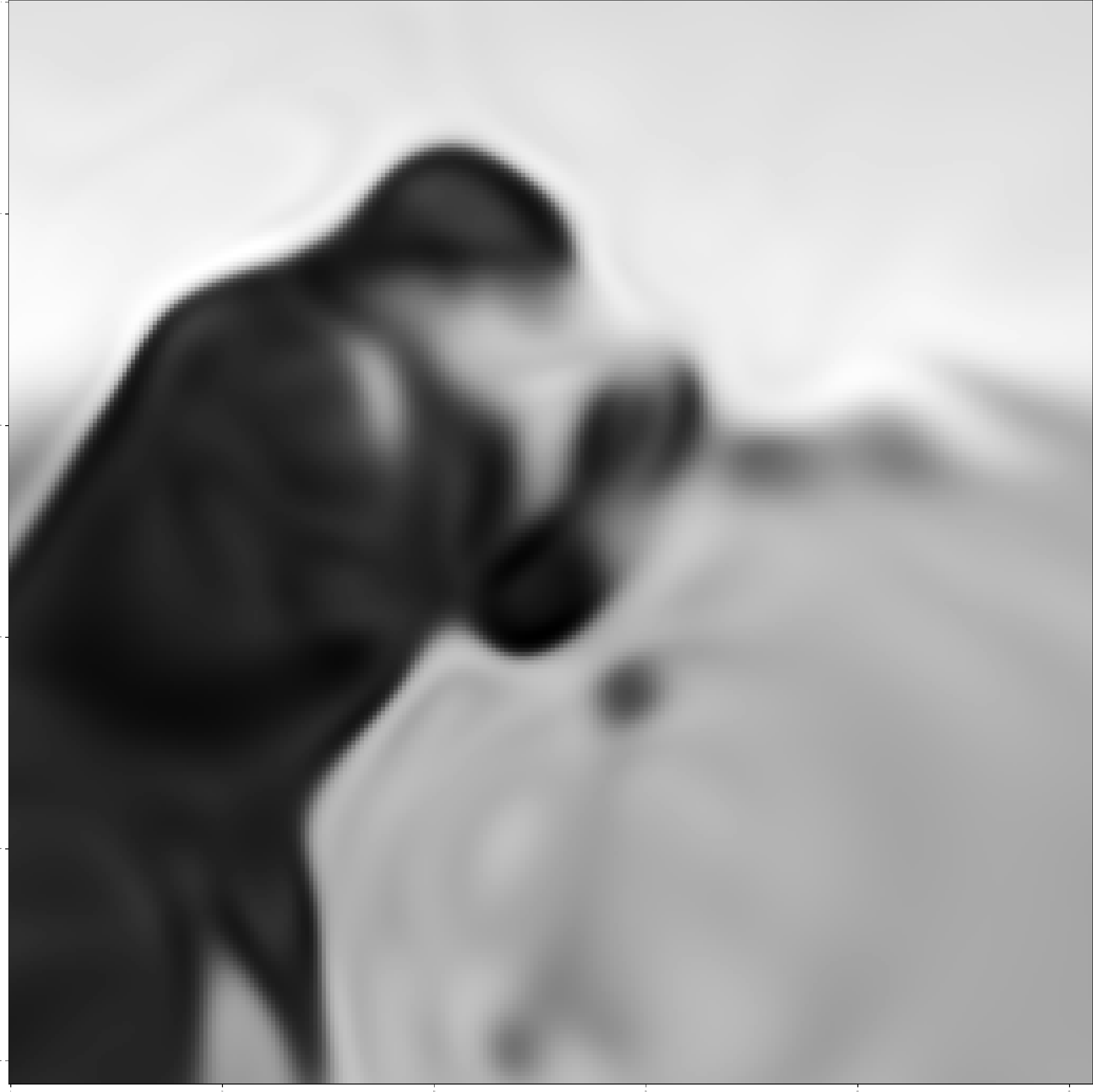}
        \caption{Fourier.}
        \label{fig:fourier_basic_cameraman_prediction}
    \end{subfigure}
    \hspace{20pt}
    \vspace{10pt}
    \begin{subfigure}[b]{0.24\columnwidth}
        \centering
        \includegraphics[width=\textwidth]{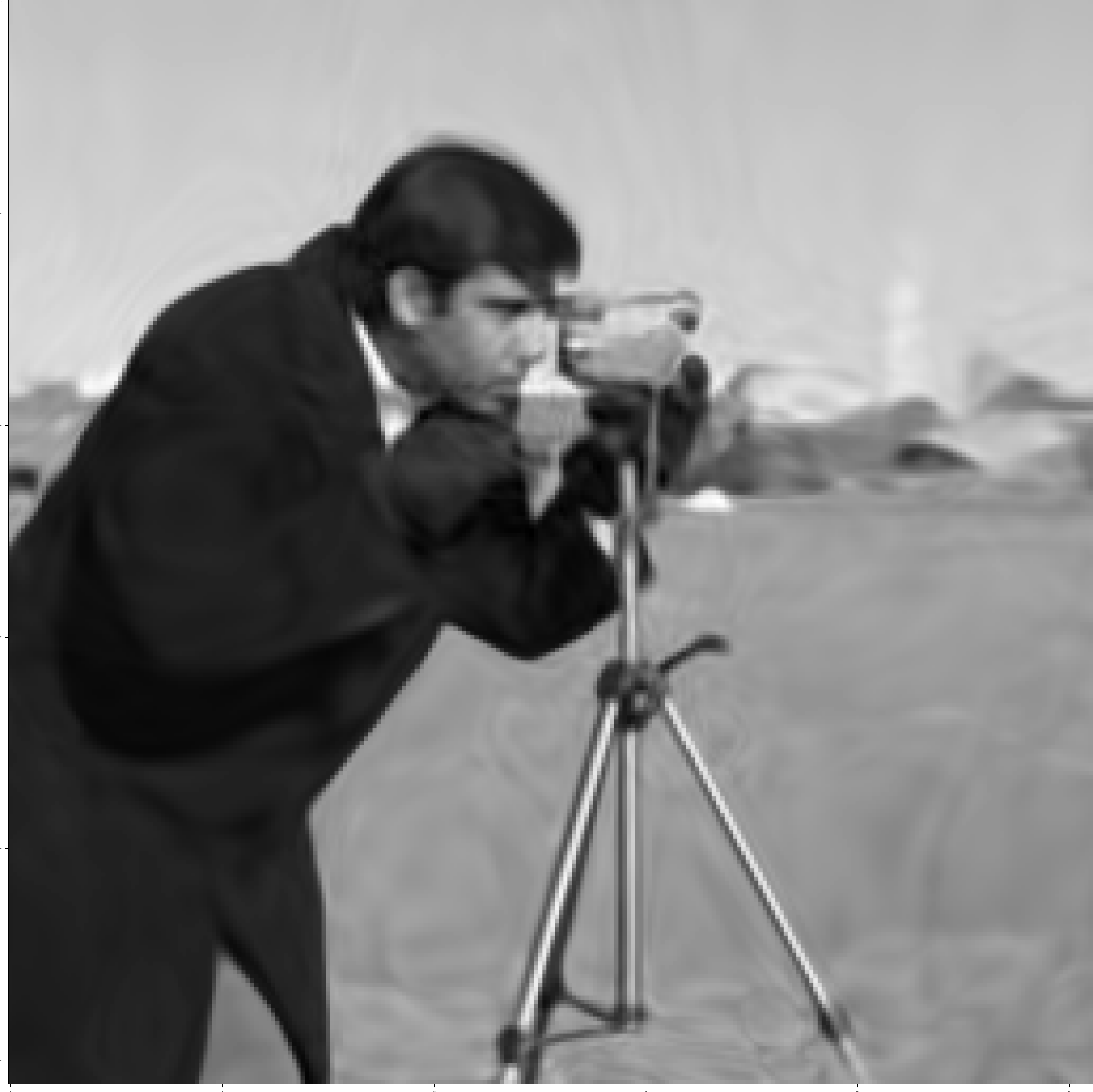}
        \caption{KAN.}
        \label{fig:kan_basic_cameraman_prediction}
    \end{subfigure}
    \hspace{20pt}
    \begin{subfigure}[b]{0.24\columnwidth}
        \centering
        \includegraphics[width=\textwidth]{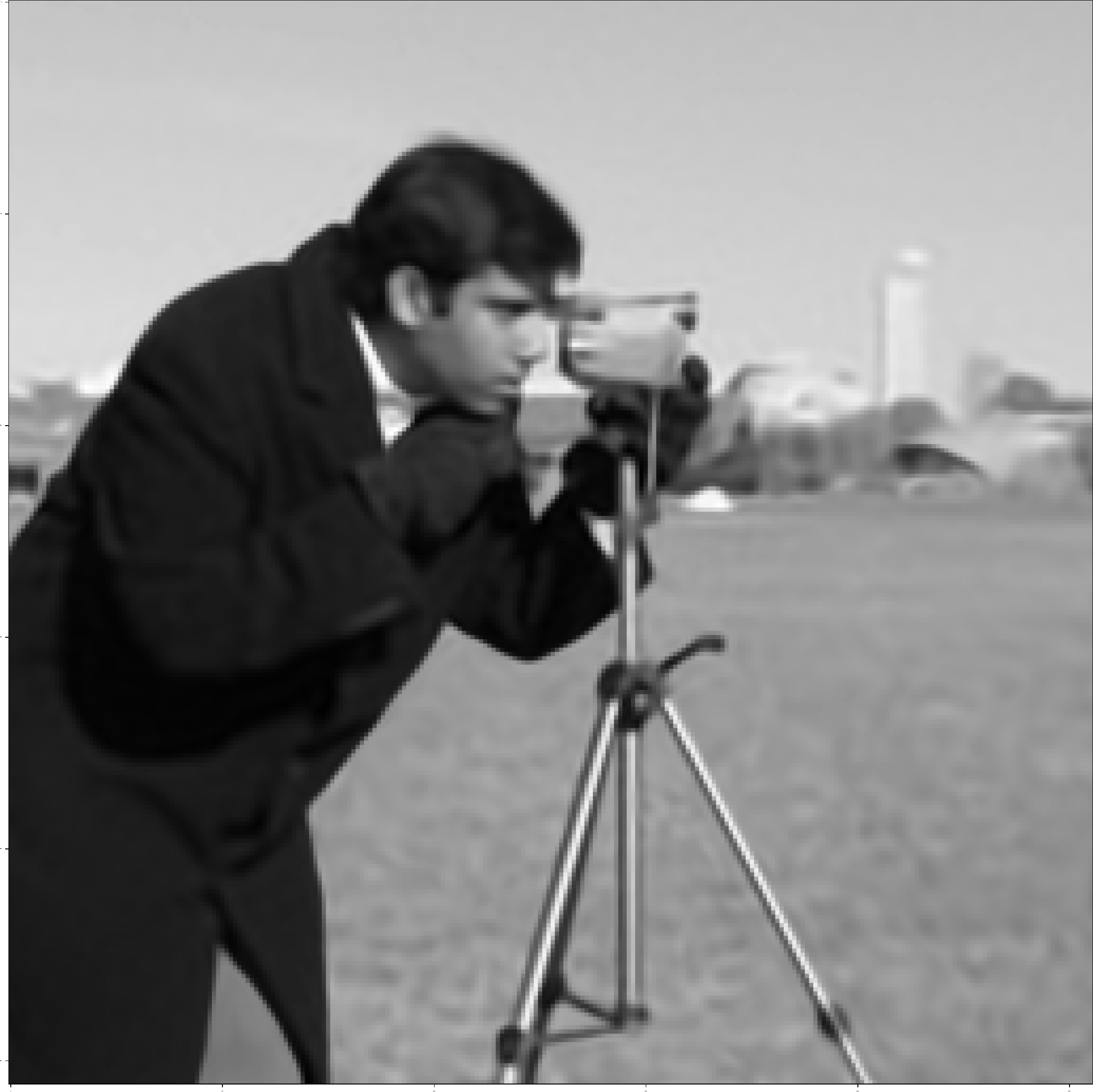}
        \caption{SIREN.}
        \label{fig:siren_basic_cameraman_prediction}
    \end{subfigure}
    \hspace{20pt}
    \begin{subfigure}[b]{0.24\columnwidth}
        \centering
        \includegraphics[width=\textwidth]{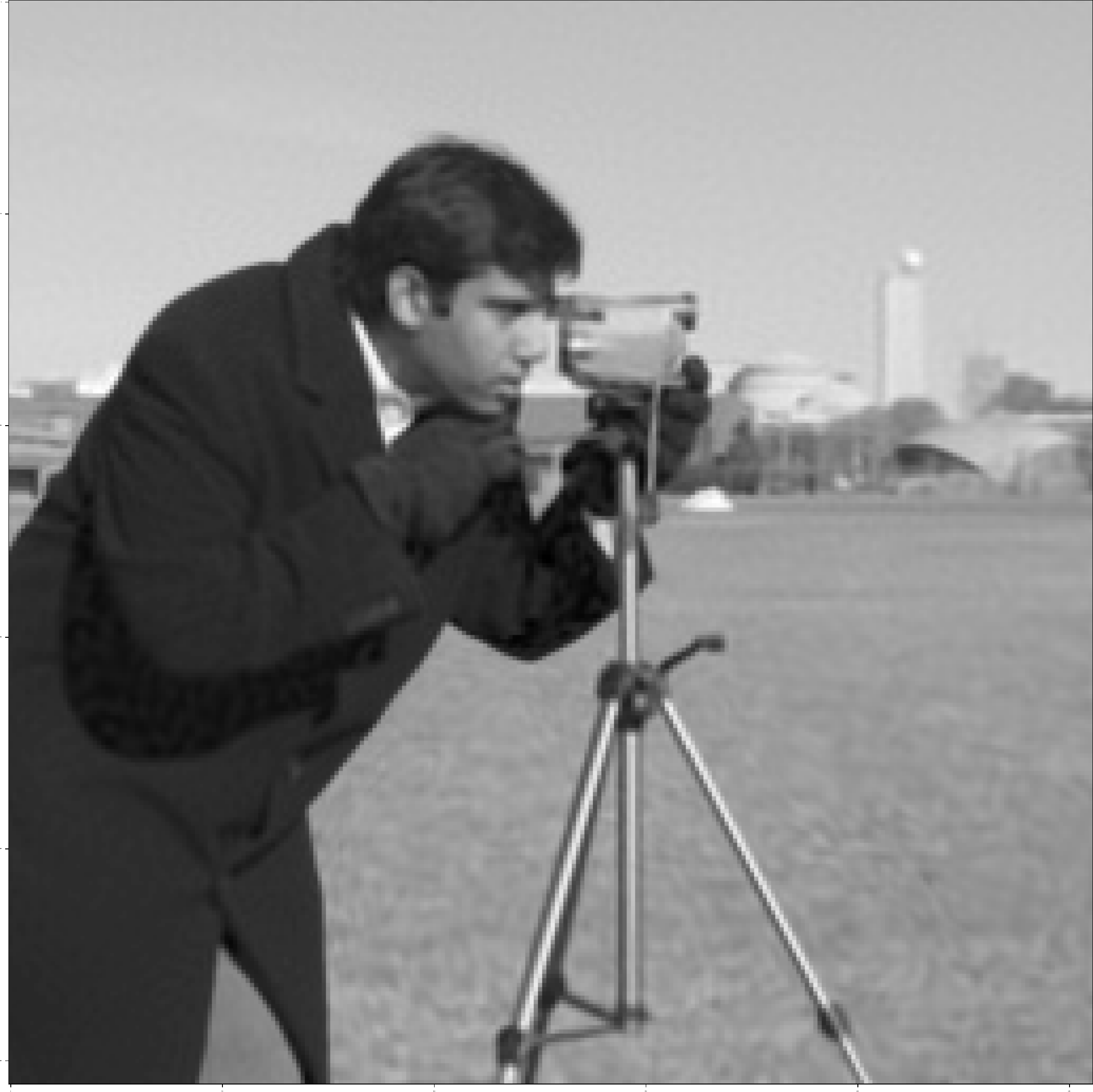}
        \caption{ENN.}
        \label{fig:enn_basic_cameraman_prediction}
    \end{subfigure}
    \hspace{20pt}
    \begin{subfigure}[b]{0.24\columnwidth}
        \centering
        \includegraphics[width=\textwidth]{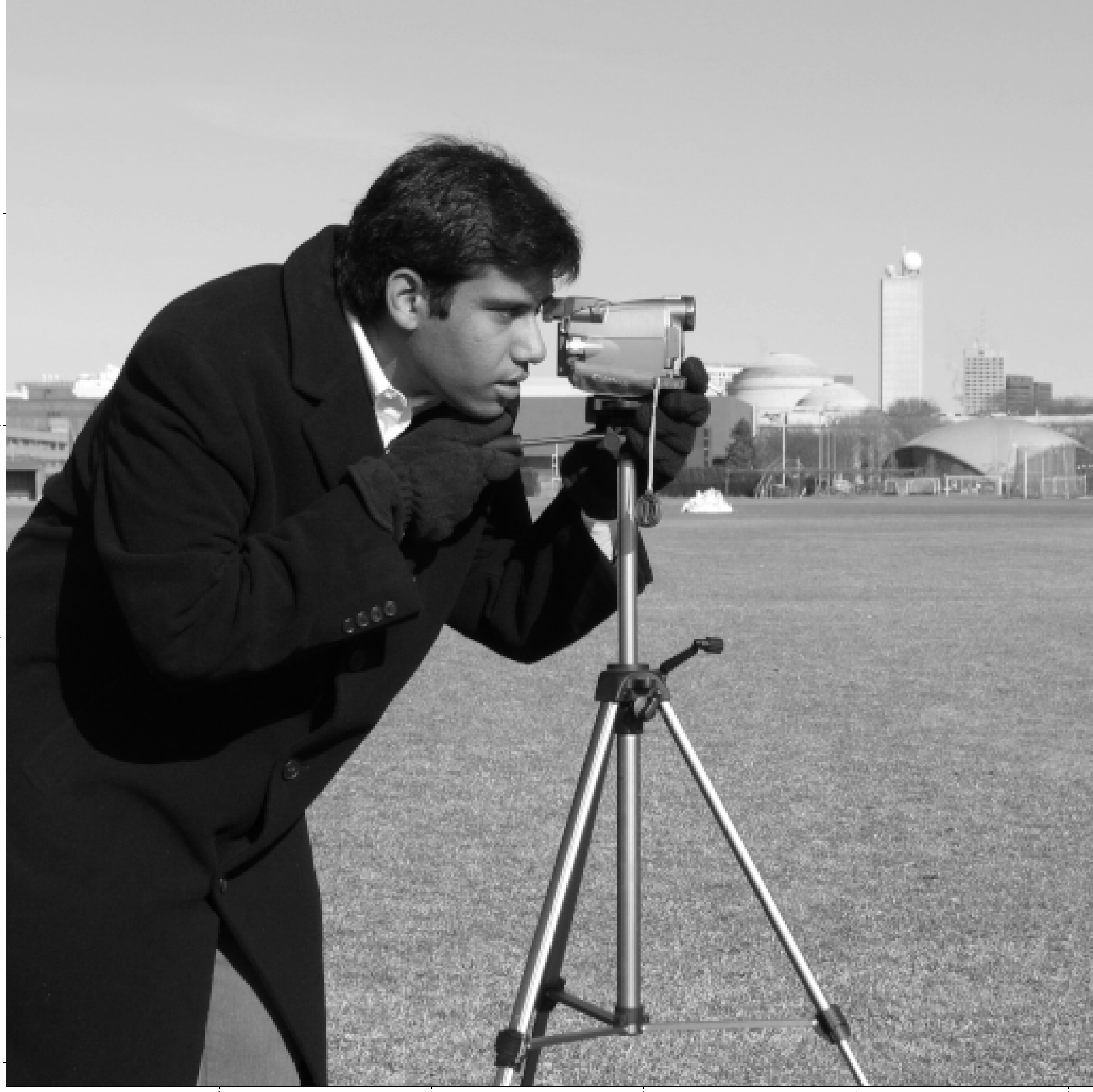}
        \caption{Ground truth.}
        \label{fig:basic_cameraman}
    \end{subfigure}
    \caption{Image predictions and ground truth on task 1.}
\label{fig:basic_cameraman_prediction}
\vspace{-1.5pt}
\end{figure}

\begin{figure}[t]
    \centering
    \begin{subfigure}[b]{0.24\columnwidth}
        \centering
        \includegraphics[width=\textwidth]{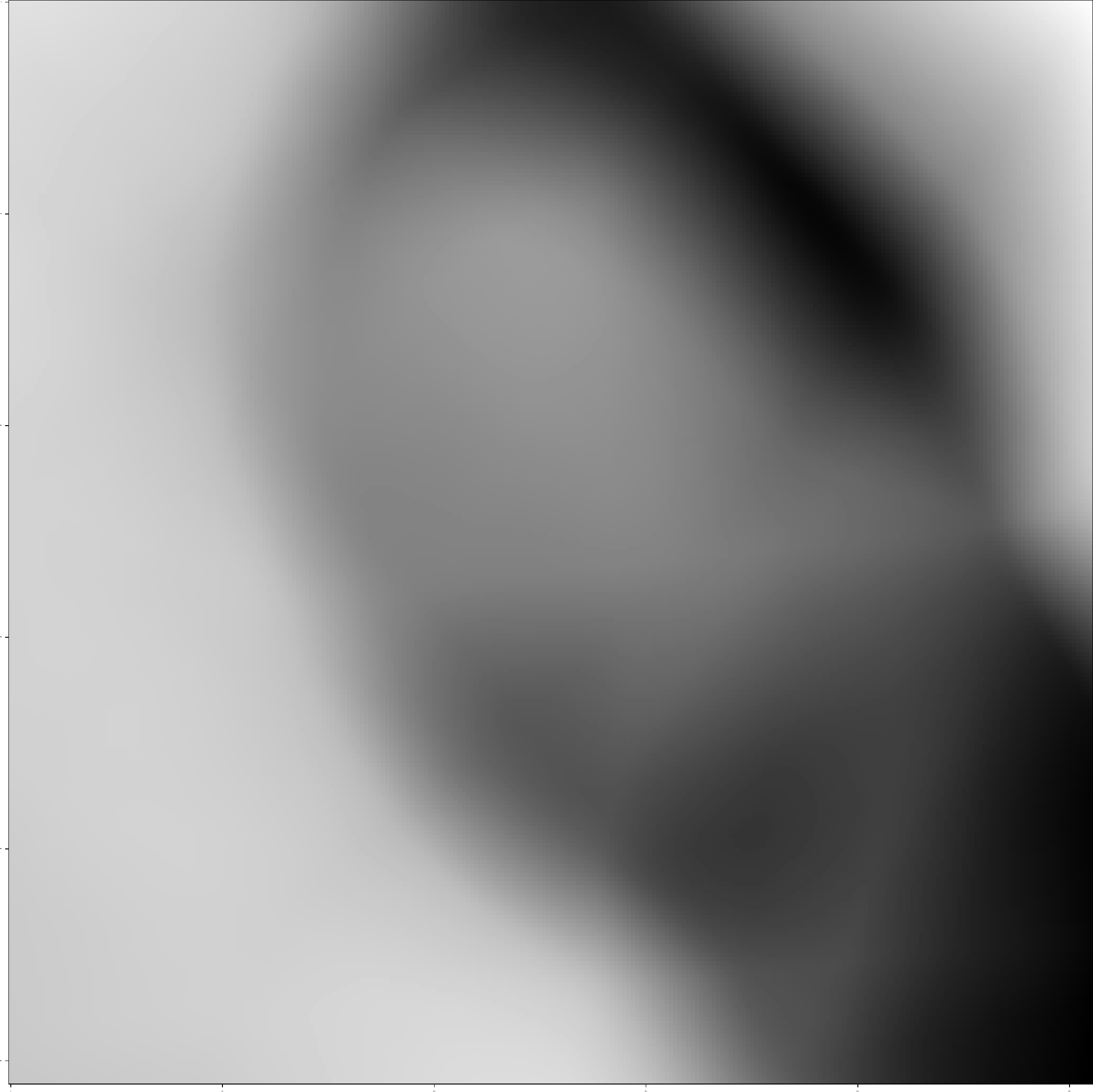}
        \caption{ReLU.}
        \label{fig:mlp_dali}
    \end{subfigure}
    \hspace{20pt}
    \begin{subfigure}[b]{0.24\columnwidth}
        \centering
        \includegraphics[width=\textwidth]{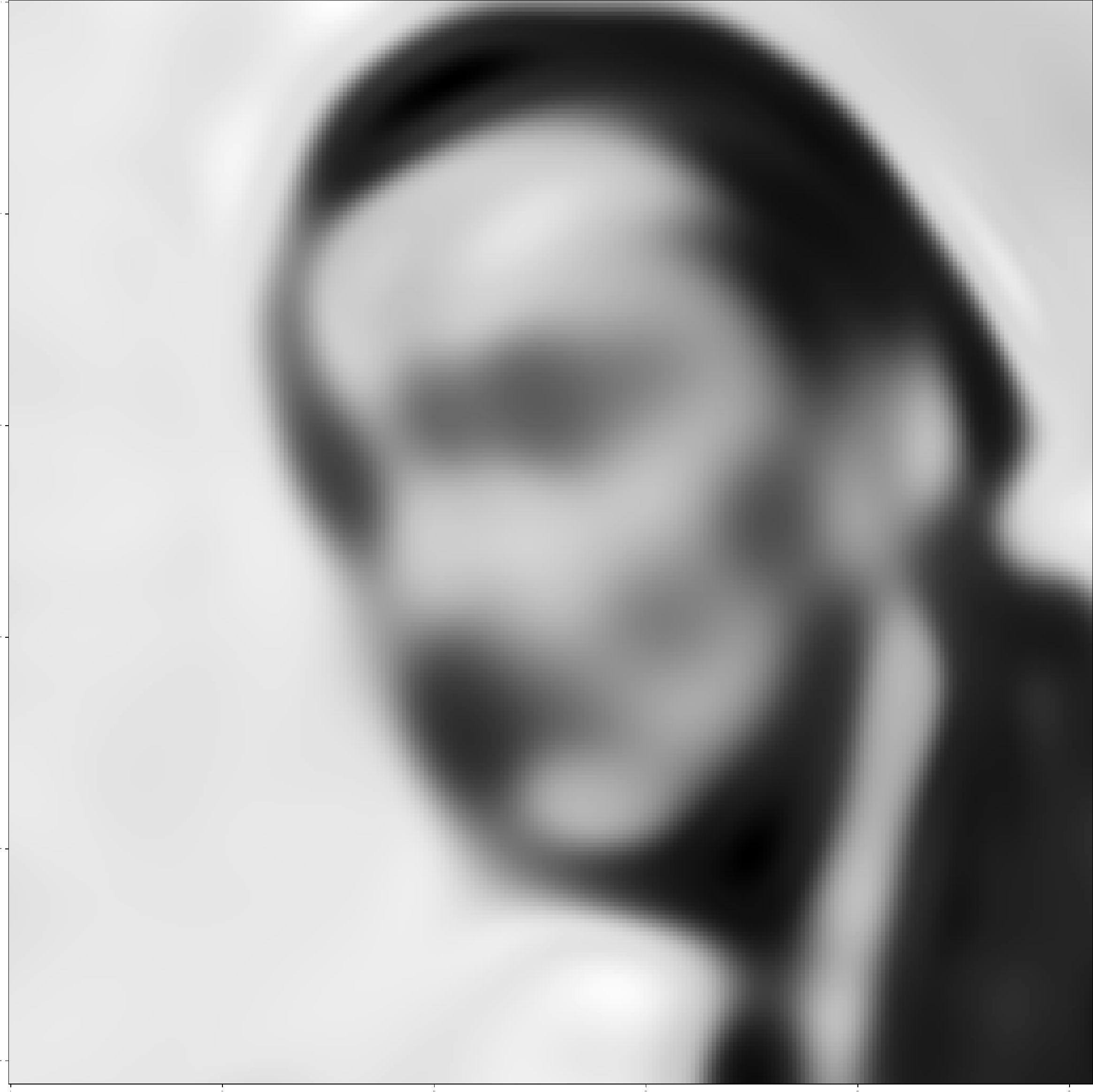}
        \caption{Fourier.}
        \label{fig:fourier_dali}
    \end{subfigure}
    \hspace{20pt}
    \vspace{10pt}
    \begin{subfigure}[b]{0.24\columnwidth}
        \centering
        \includegraphics[width=\textwidth]{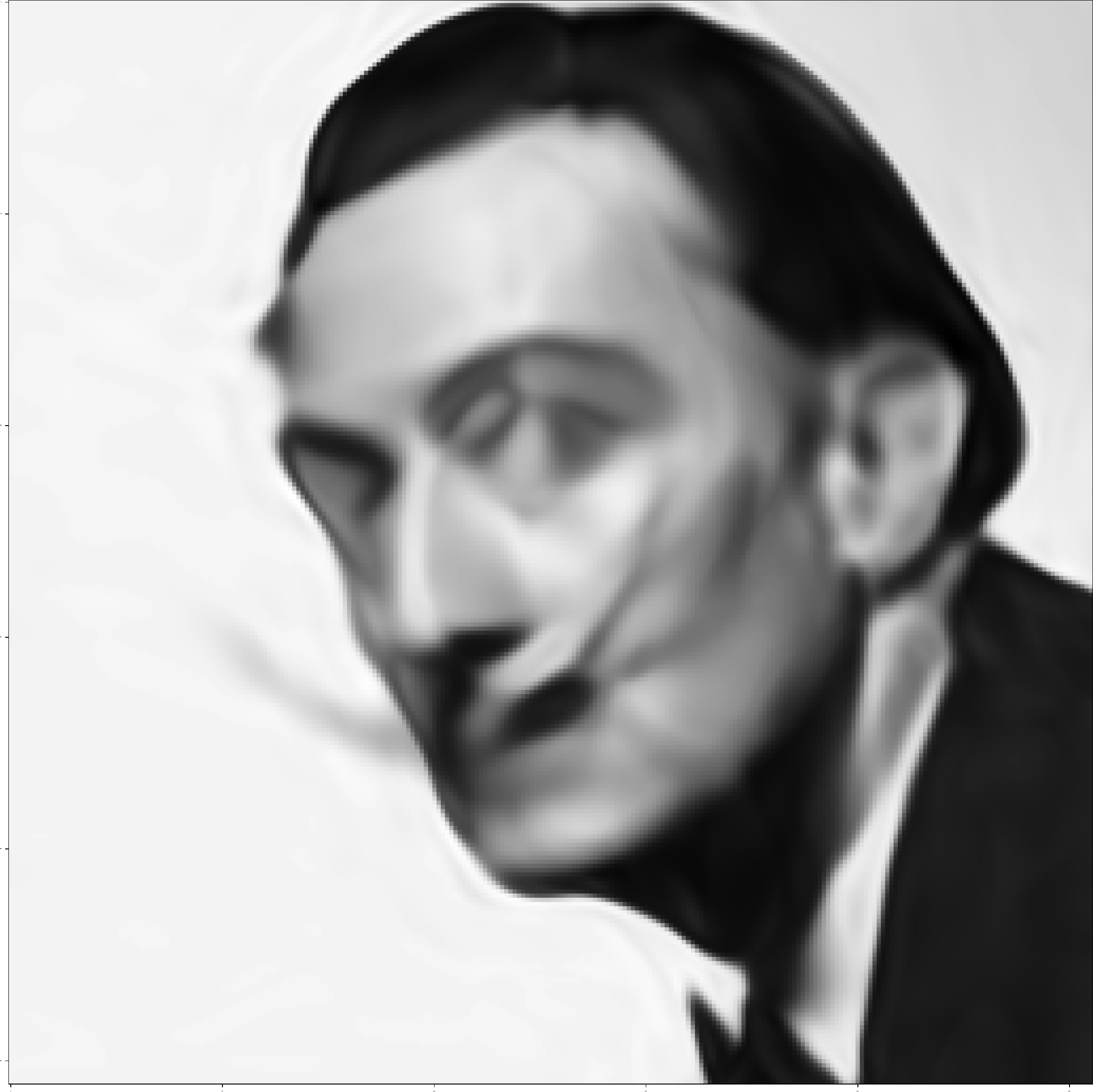}
        \caption{KAN.}
        \label{fig:kan_dali}
    \end{subfigure}
    \hspace{20pt}
    \begin{subfigure}[b]{0.24\columnwidth}
        \centering
        \includegraphics[width=\textwidth]{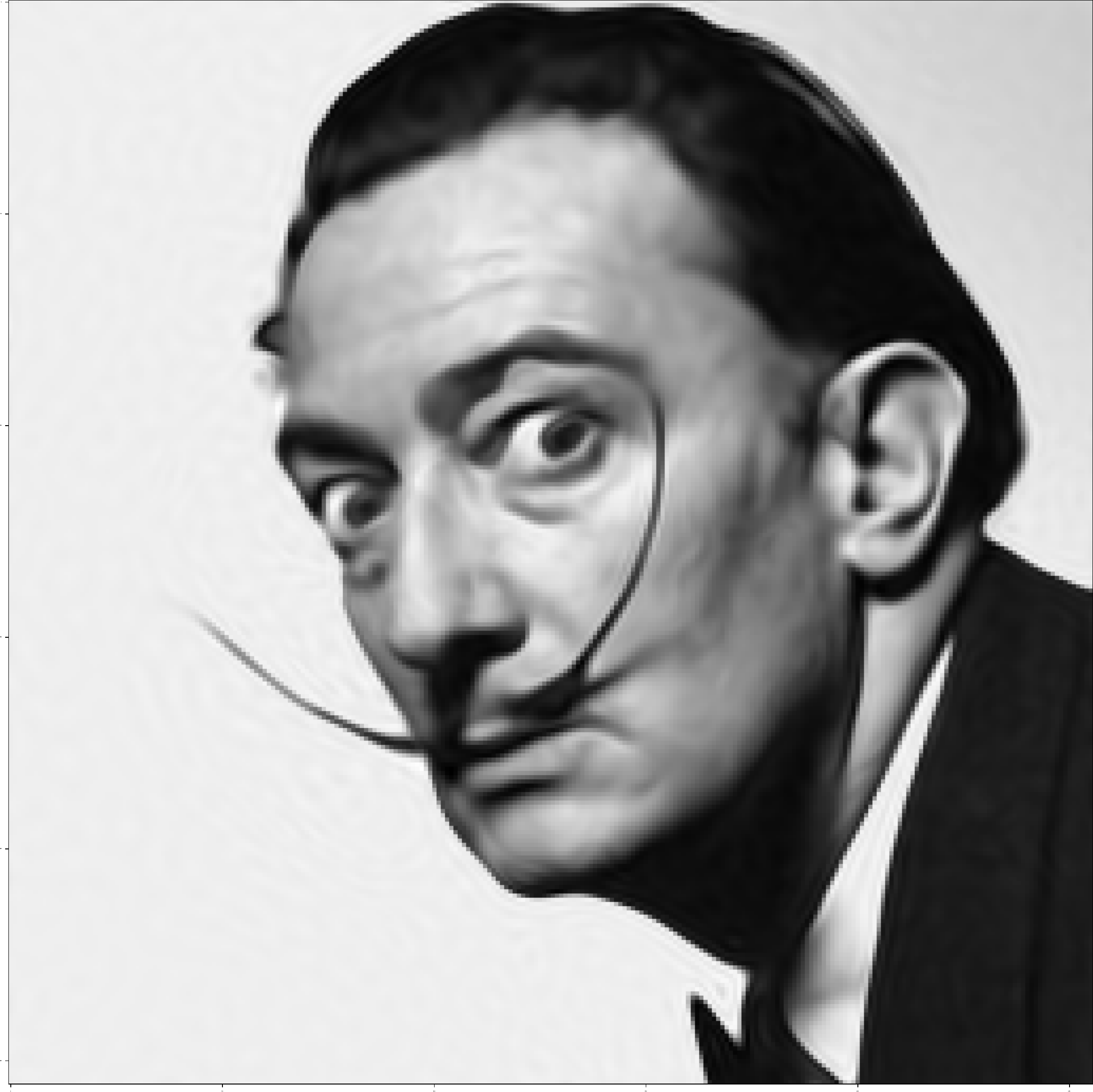}
        \caption{SIREN.}
        \label{fig:siren_dali}
    \end{subfigure}
    \hspace{20pt}
    \begin{subfigure}[b]{0.24\columnwidth}
        \centering
        \includegraphics[width=\textwidth]{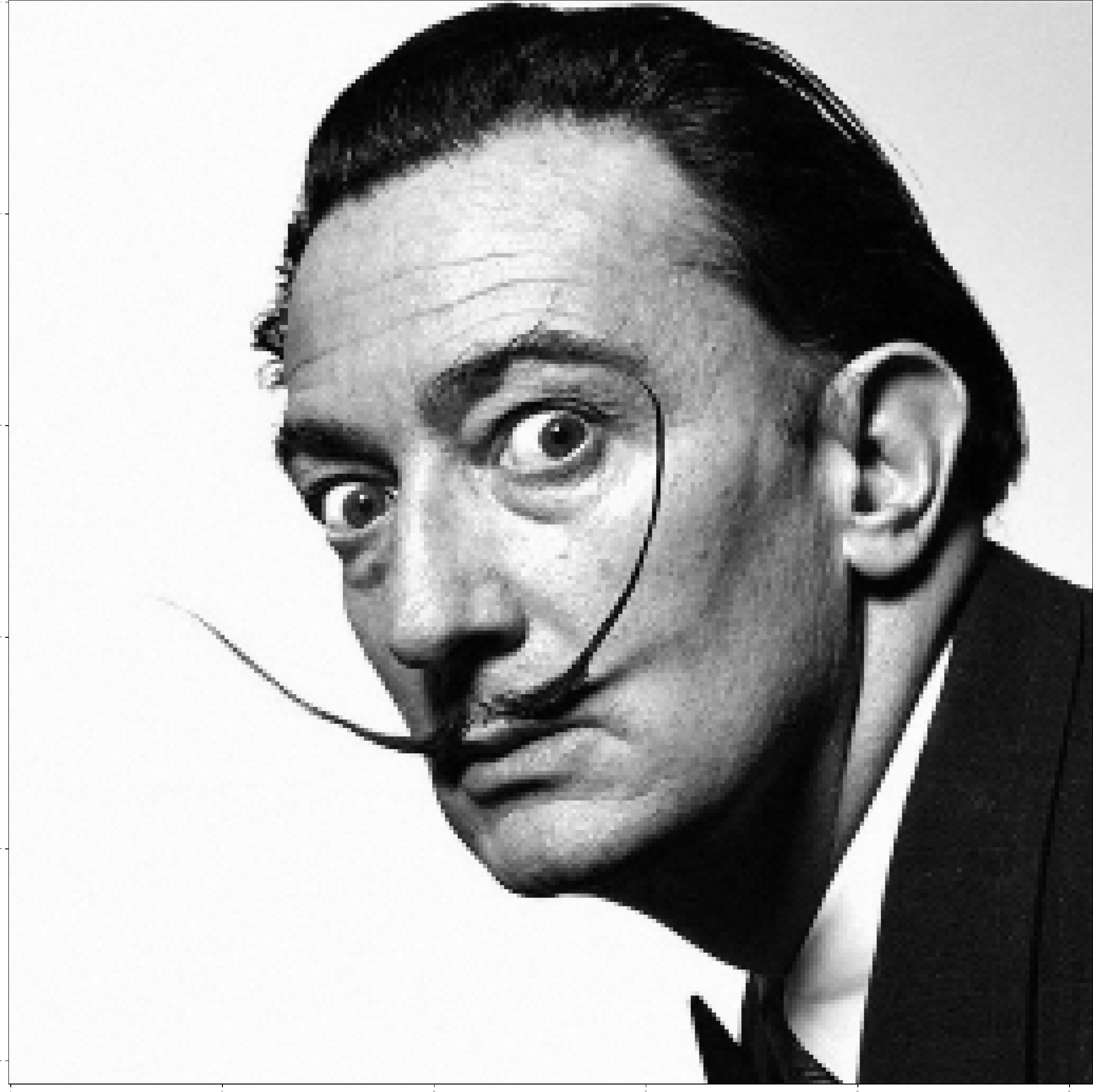}
        \caption{ENN.}
        \label{fig:enn_dali}
    \end{subfigure}
    \hspace{20pt}
    \begin{subfigure}[b]{0.24\columnwidth}
        \centering
        \includegraphics[width=\textwidth]{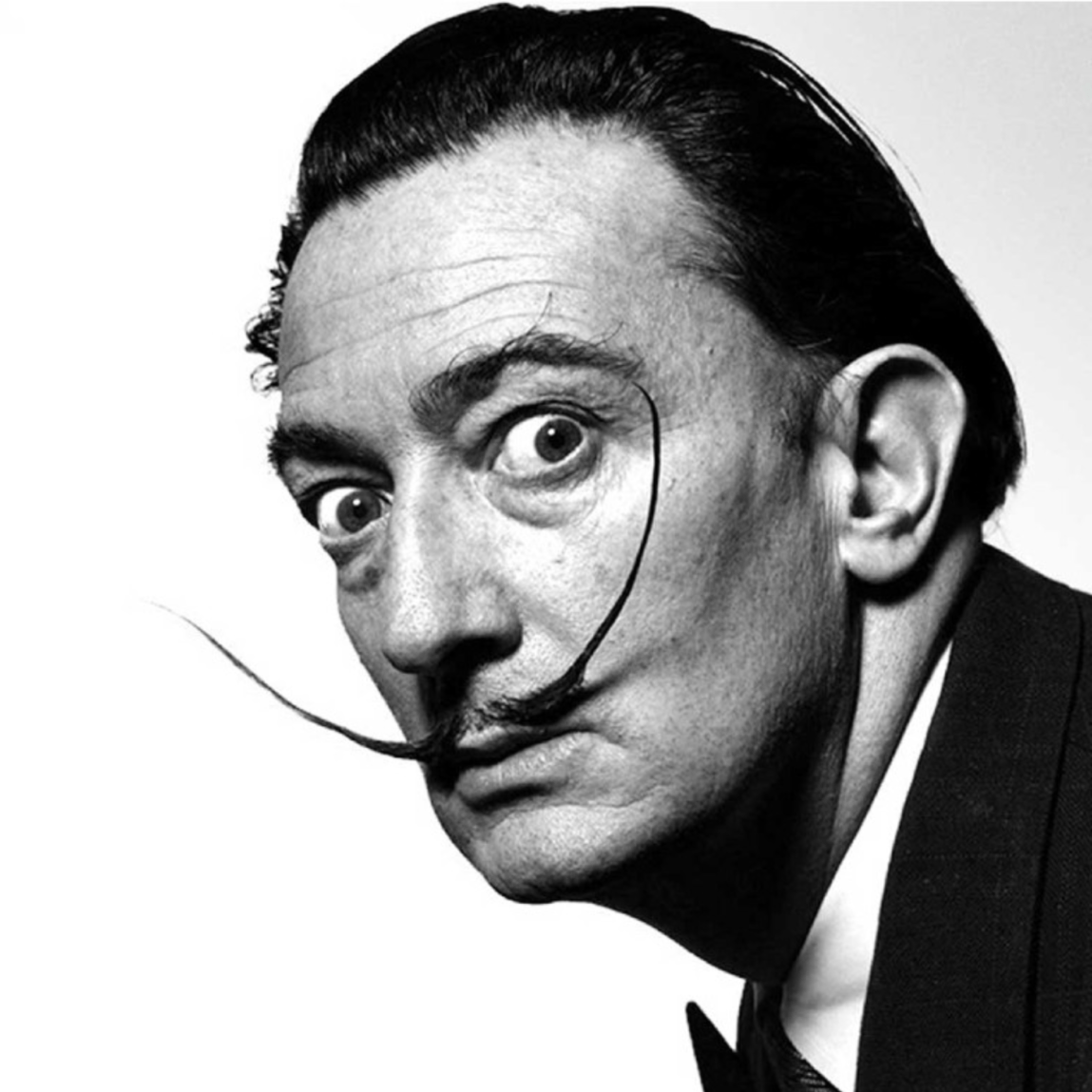}
        \caption{Ground truth.}
        \label{fig:dali}
    \end{subfigure}
    \caption{Image predictions and ground truth on task 2.}
\label{fig:dali_predictions}
\end{figure}

\begin{figure}[t]
    \centering
    \begin{subfigure}[b]{0.24\columnwidth}
        \centering
        \includegraphics[width=\textwidth]{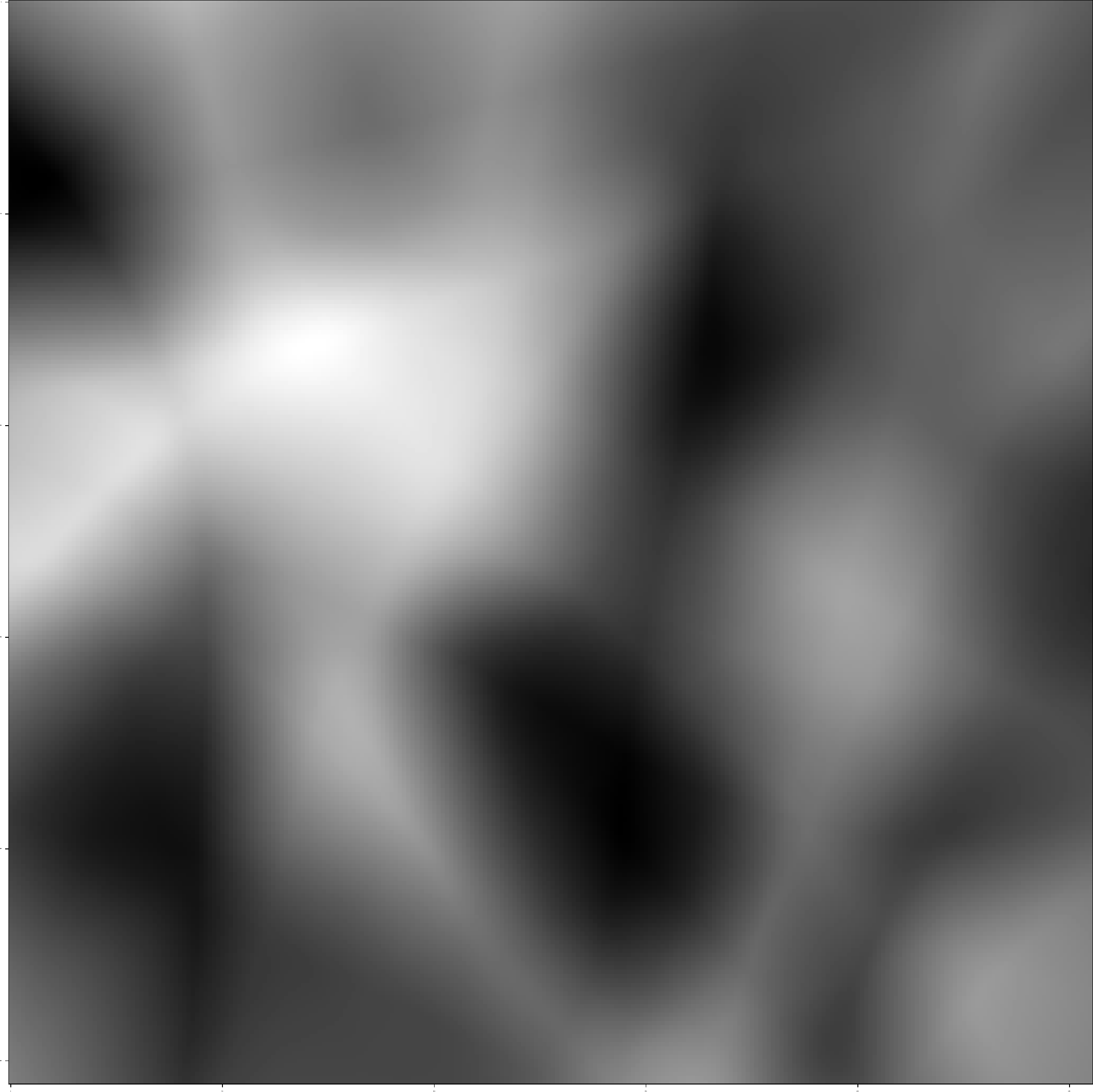}
        \caption{ReLU.}
        \label{fig:mlp_barbara}
    \end{subfigure}
    \hspace{20pt}
    \begin{subfigure}[b]{0.24\columnwidth}
        \centering
        \includegraphics[width=\textwidth]{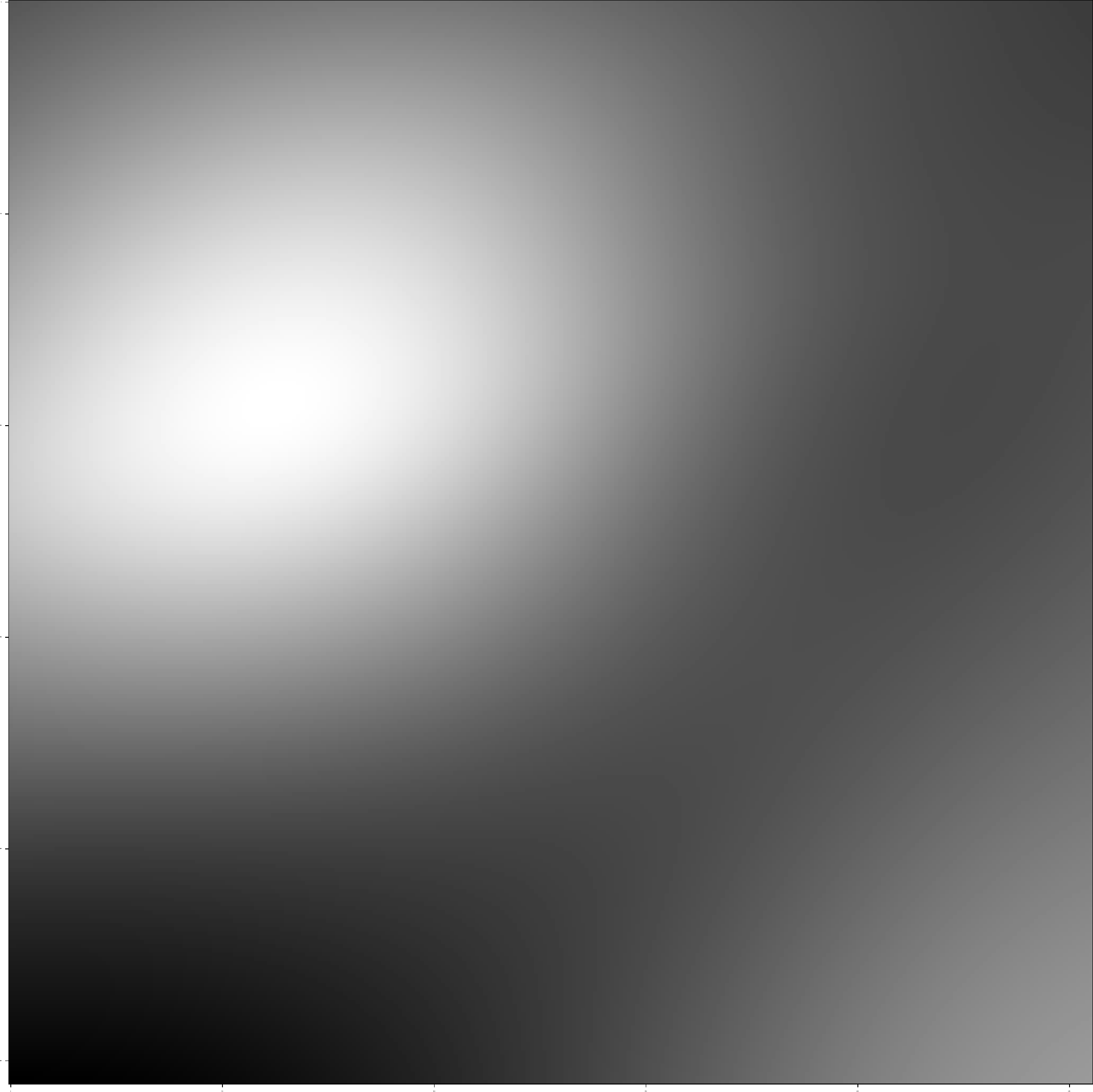}
        \caption{Fourier.}
        \label{fig:fourier_barbara}
    \end{subfigure}
    \hspace{20pt}
    \vspace{10pt}
    \begin{subfigure}[b]{0.24\columnwidth}
        \centering
        \includegraphics[width=\textwidth]{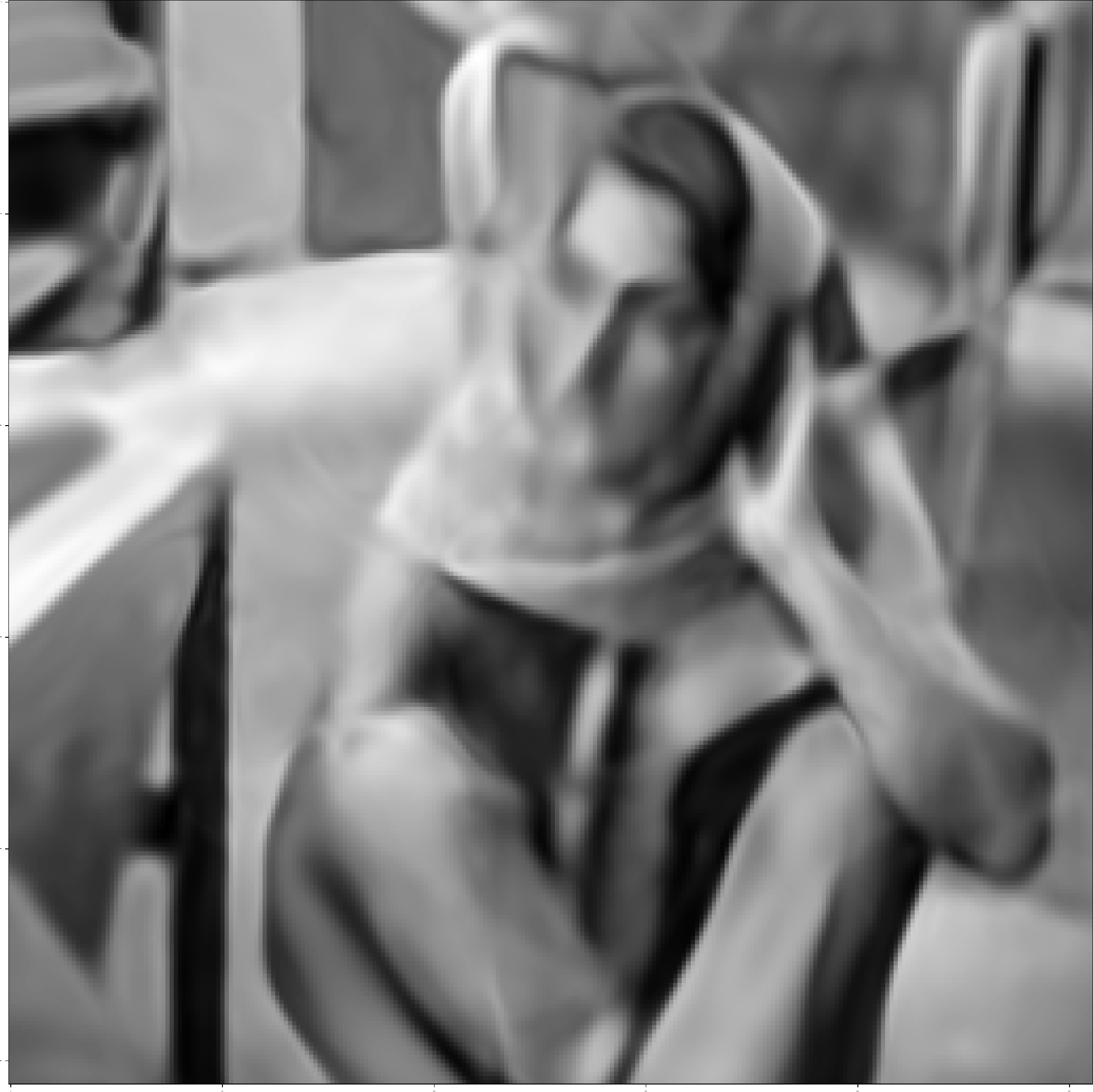}
        \caption{KAN.}
        \label{fig:kan_barbara}
    \end{subfigure}
    \hspace{20pt}
    \begin{subfigure}[b]{0.24\columnwidth}
        \centering
        \includegraphics[width=\textwidth]{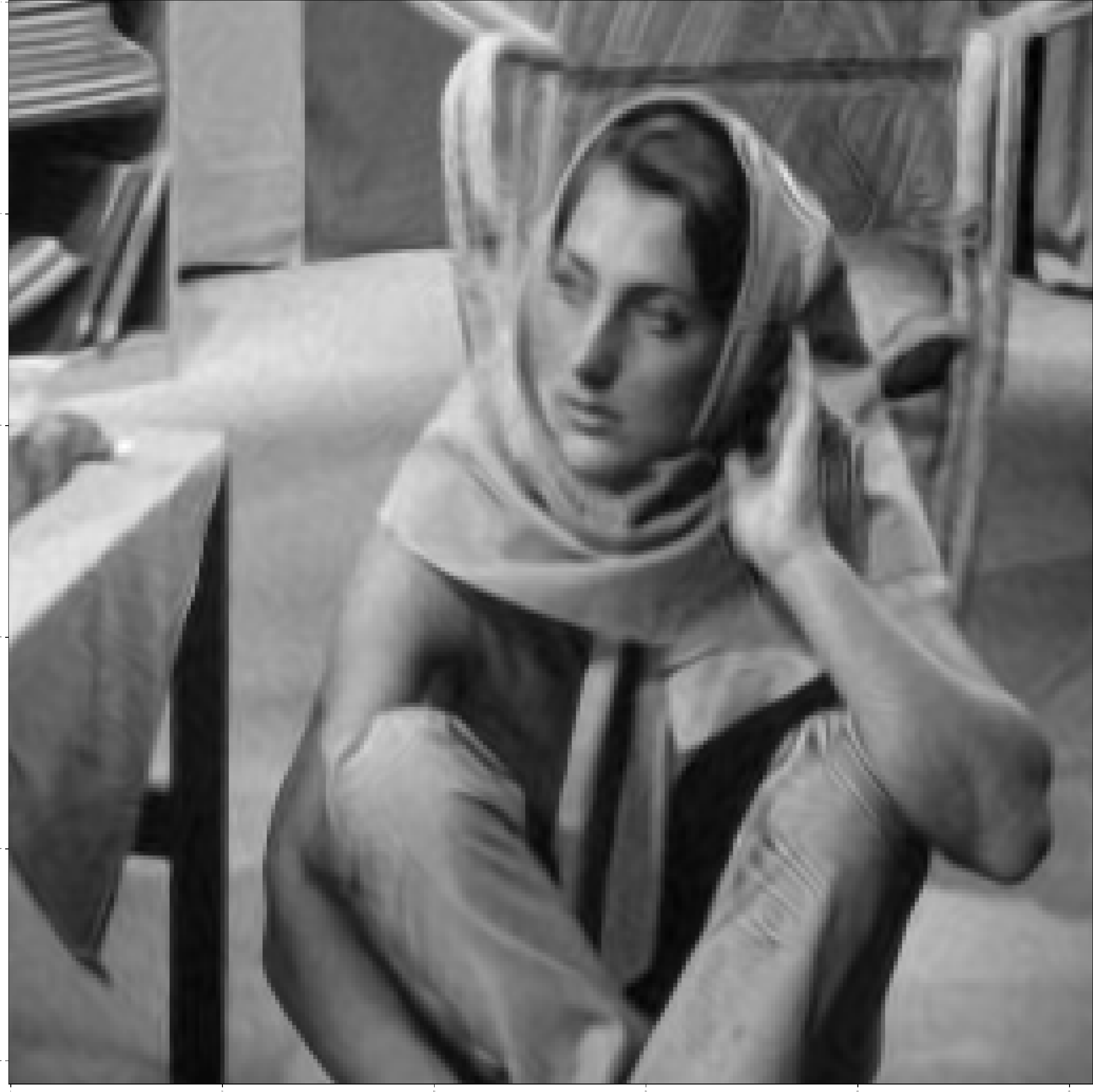}
        \caption{SIREN.}
        \label{fig:siren_barbara}
    \end{subfigure}
    \hspace{20pt}
    \begin{subfigure}[b]{0.24\columnwidth}
        \centering
        \includegraphics[width=\textwidth]{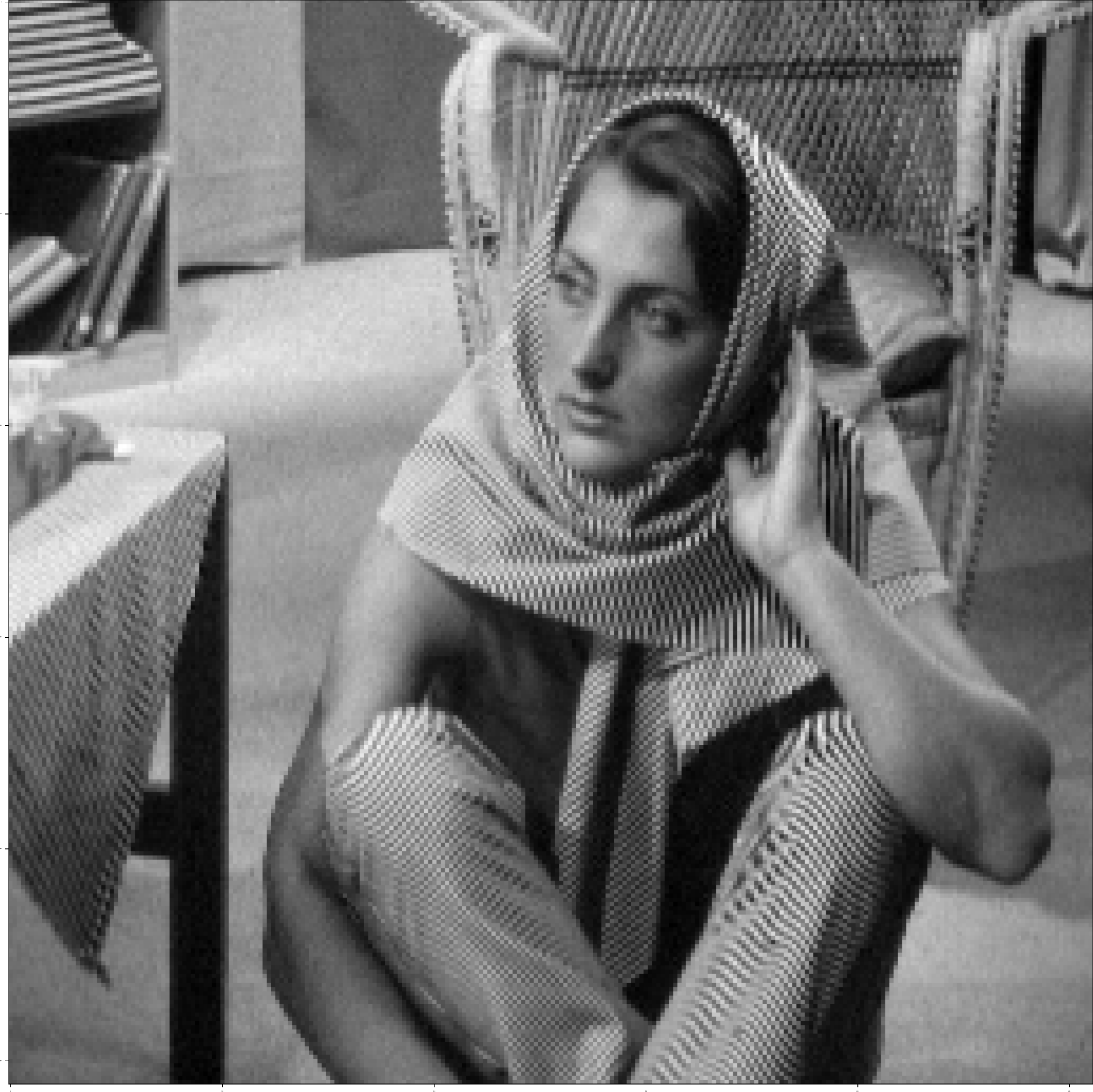}
        \caption{ENN.}
        \label{fig:enn_barbara}
    \end{subfigure}
    \hspace{20pt}
    \begin{subfigure}[b]{0.24\columnwidth}
        \centering
        \includegraphics[width=\textwidth]{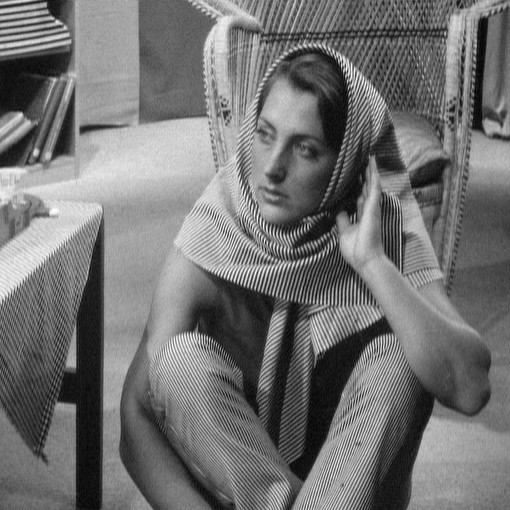}
        \caption{Ground truth.}
        \label{fig:barbara}
    \end{subfigure}
    \caption{Image predictions and ground truth on task 3.}
\label{fig:barbara_predictions}
\vspace{-1.73pt}
\end{figure}

\begin{figure}[t]
    \centering
    \begin{subfigure}[b]{0.24\columnwidth}
        \centering
        \includegraphics[width=\textwidth]{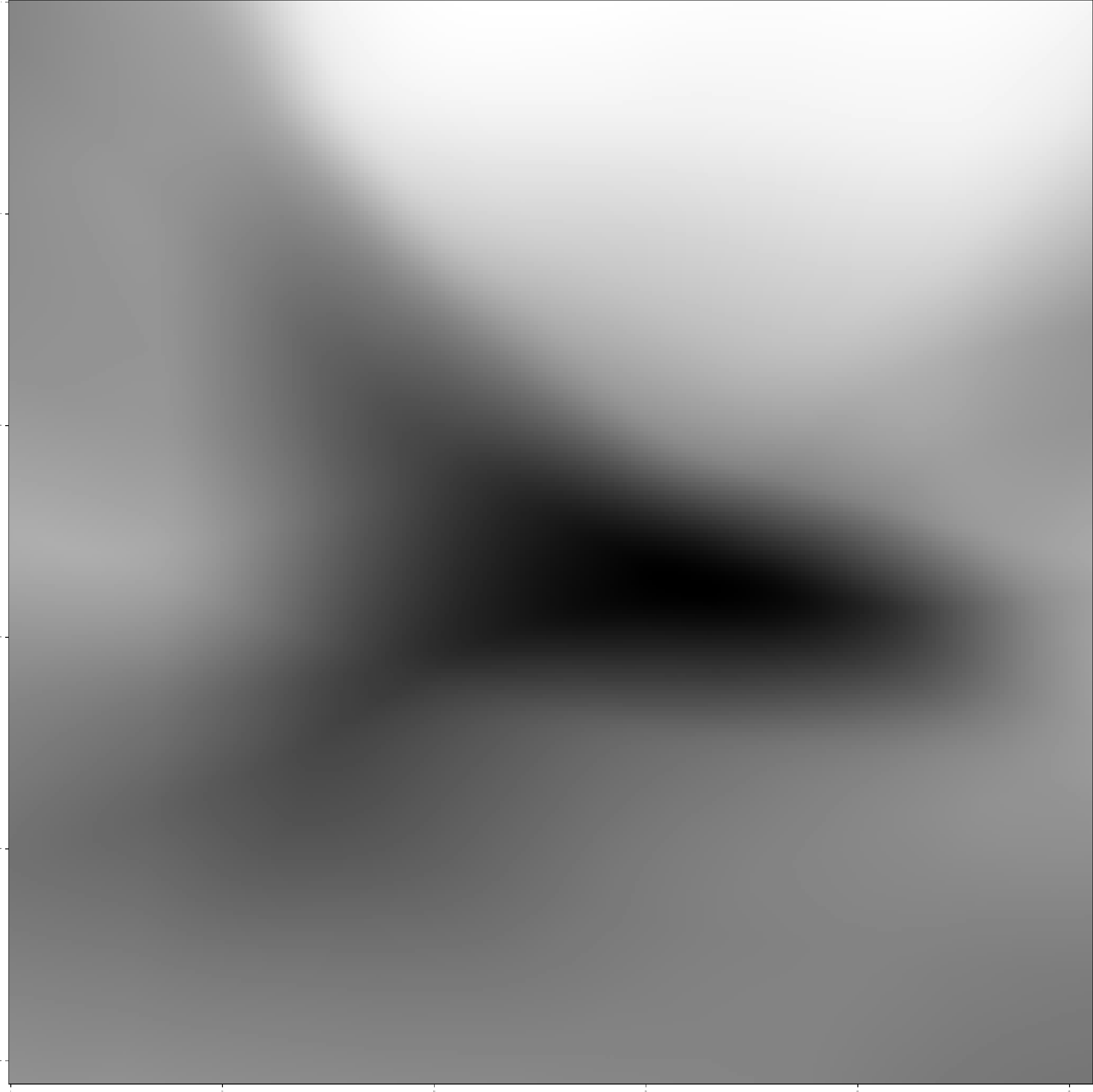}
        \caption{ReLU.}
        \label{fig:mlp_motion}
    \end{subfigure}
    \hspace{20pt}
    \begin{subfigure}[b]{0.24\columnwidth}
        \centering
        \includegraphics[width=\textwidth]{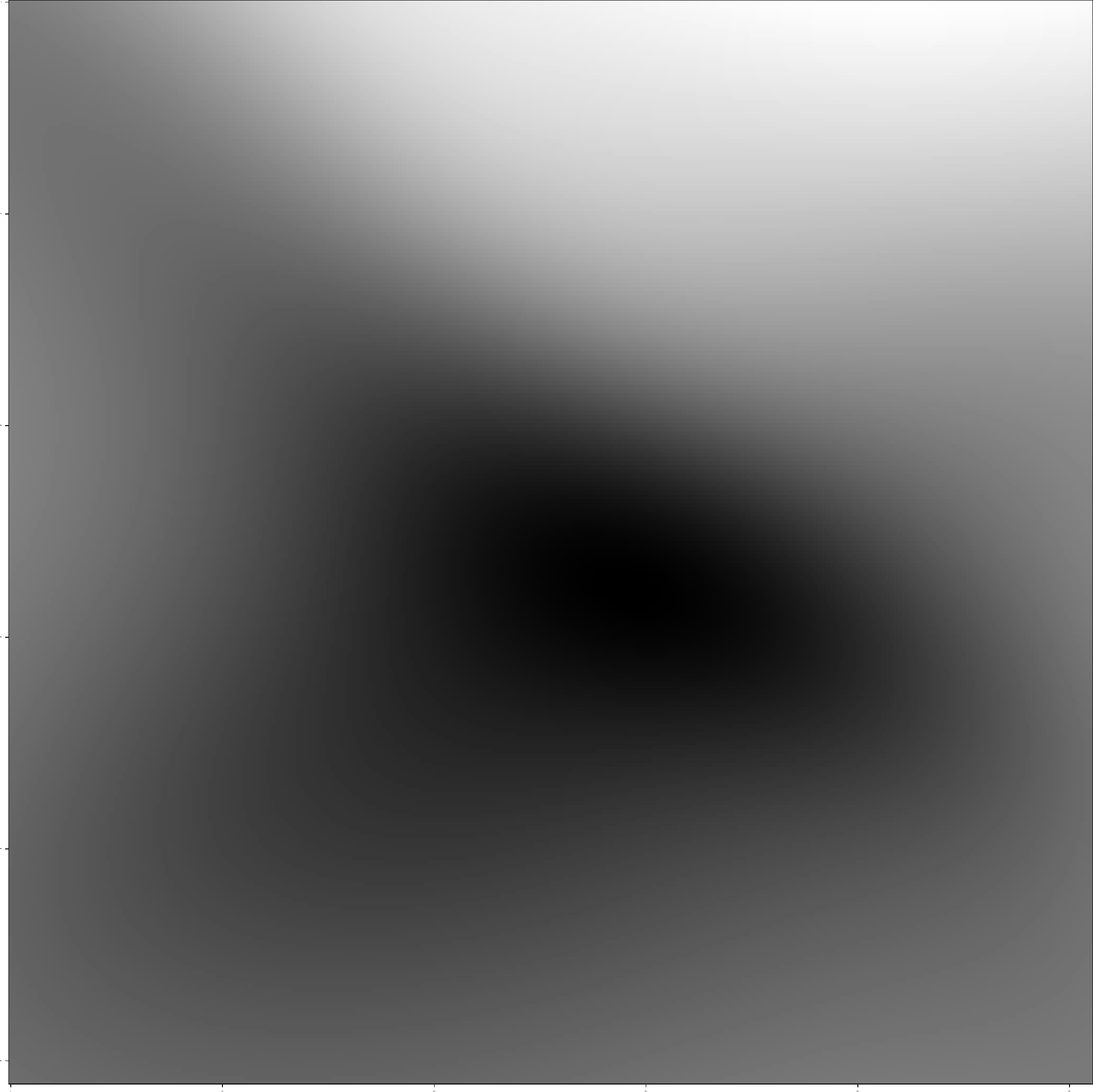}
        \caption{Fourier.}
        \label{fig:fourier_motion}
    \end{subfigure}
    \hspace{20pt}
    \vspace{10pt}
    \begin{subfigure}[b]{0.24\columnwidth}
        \centering
        \includegraphics[width=\textwidth]{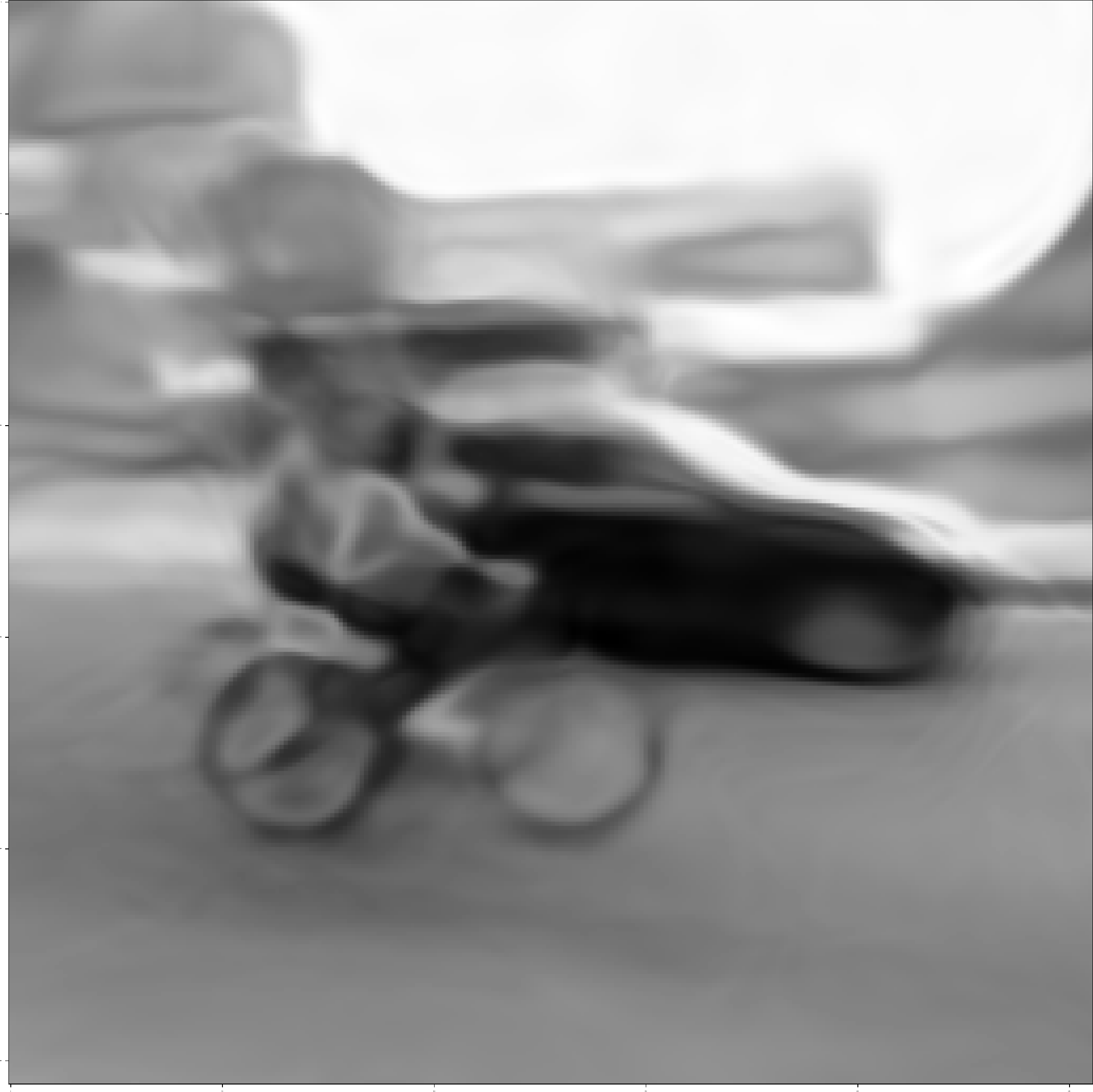}
        \caption{KAN.}
        \label{fig:kan_motion}
    \end{subfigure}
    \hspace{20pt}
    \begin{subfigure}[b]{0.24\columnwidth}
        \centering
        \includegraphics[width=\textwidth]{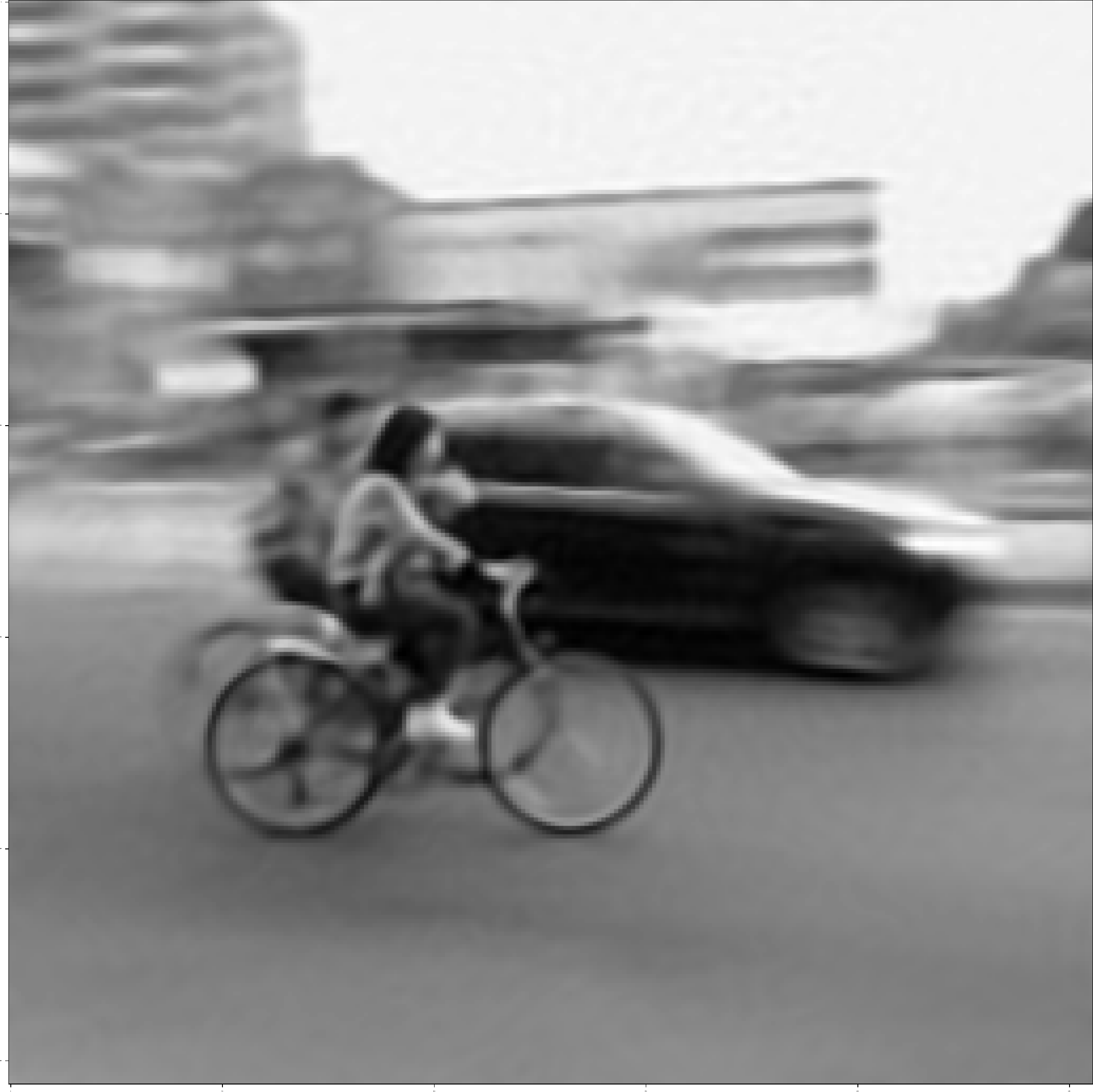}
        \caption{SIREN.}
        \label{fig:siren_motion}
    \end{subfigure}
    \hspace{20pt}
    \begin{subfigure}[b]{0.24\columnwidth}
        \centering
        \includegraphics[width=\textwidth]{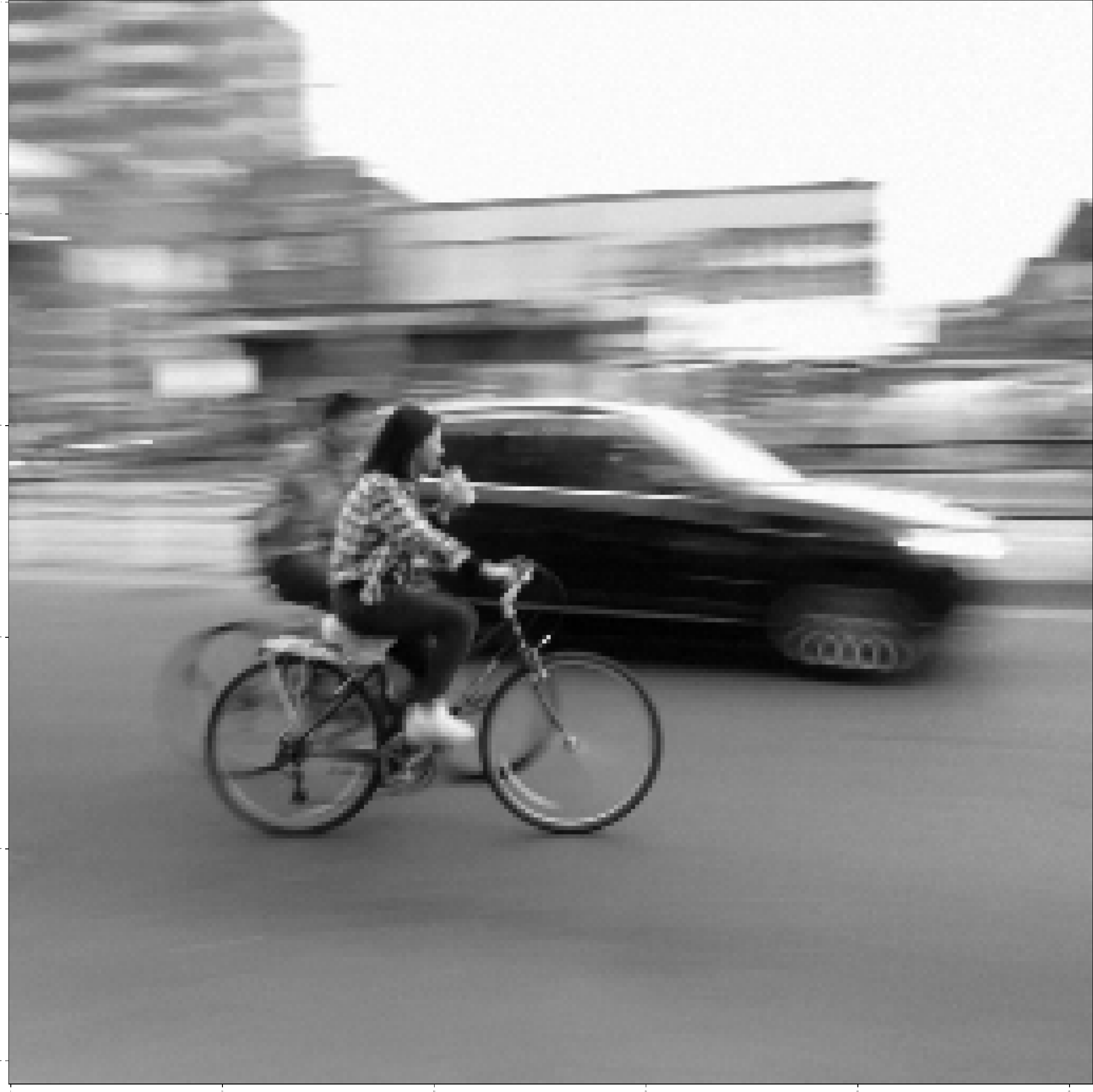}
        \caption{ENN.}
        \label{fig:enn_motion}
    \end{subfigure}
    \hspace{20pt}
    \begin{subfigure}[b]{0.24\columnwidth}
        \centering
        \includegraphics[width=\textwidth]{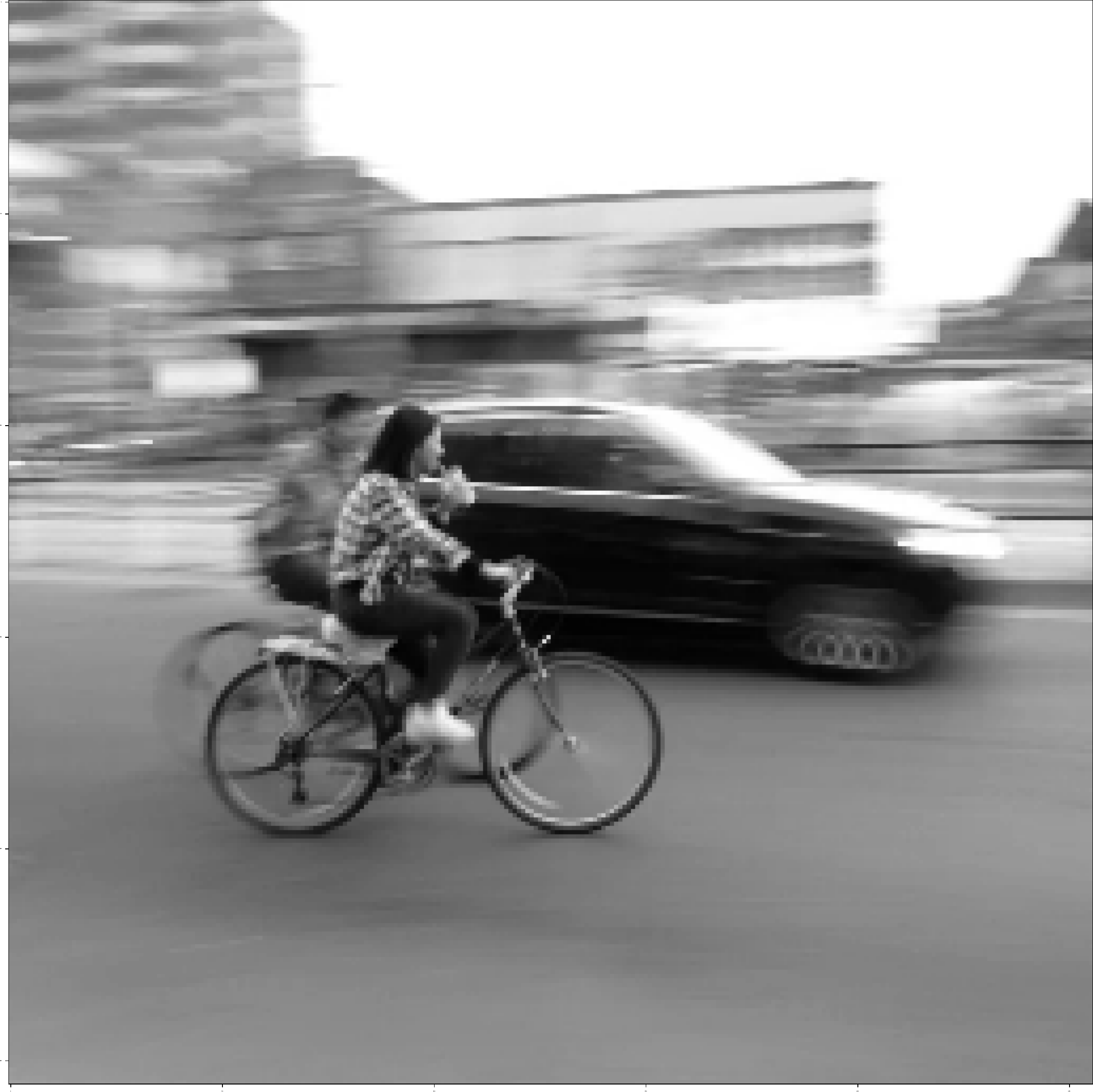}
        \caption{Ground truth.}
        \label{fig:motion}
    \end{subfigure}
    \caption{Image predictions and ground truth on task 4.}
\label{fig:motion_predictions}
\end{figure}

To systematically evaluate model performance, we consider four image reconstruction tasks. These images vary in their content and structural complexity, providing a diverse benchmark for comparison. Table \ref{tab:results_renn} summarizes the performance across all models and tasks, and Figures~\ref{fig:basic_cameraman_prediction}-\ref{fig:motion_predictions}illustrate the corresponding image reconstructions. 

\begin{table*}[t]
\centering
\normalsize
\caption{MSE on various INR tasks.}
\begin{tabular}{lccccc}
\toprule
\textbf{Task} & \textbf{ReLU} & \textbf{Fourier} & \textbf{KAN} & \textbf{SIREN} & \textbf{ENN} \\
\midrule
Figure \ref{fig:basic_cameraman_prediction} & $7.3 \cdot 10^{-2}$  & $3.4 \cdot 10^{-2}$  & $2.4 \cdot 10^{-2}$  & $3.6 \cdot 10^{-3}$  & $\mathbf{8.4 \cdot 10^{-4}}$ \\
Figure \ref{fig:dali_predictions} & $2.1 \cdot 10^{-1}$  & $7.3 \cdot 10^{-2}$  & $3.9 \cdot 10^{-2}$  & $5.2 \cdot 10^{-3}$  & $\mathbf{7.6 \cdot 10^{-4}}$ \\
Figure \ref{fig:barbara_predictions} & $8.9 \cdot 10^{-2}$  & $1.1 \cdot 10^{-1}$  & $3.3 \cdot 10^{-2}$  & $1.0 \cdot 10^{-2}$  & $\mathbf{2.7 \cdot 10^{-4}}$ \\
Figure \ref{fig:motion_predictions} & $5.2 \cdot 10^{-2}$  & $7.3 \cdot 10^{-2}$  & $1.3 \cdot 10^{-2}$  & $4.0 \cdot 10^{-3}$  & $\mathbf{1.7 \cdot 10^{-4}}$ \\
\bottomrule
\end{tabular}
\label{tab:results_renn}
\end{table*}

A summary of the results and limitations of each model follows:
\begin{itemize}
    \item \textbf{ReLU}: As documented in the literature, ReLU activations exhibit a low-frequency bias \cite{rahaman2019spectralbias}. This behavior leads to a consistent failure to capture sharp transitions across all tasks.
    
    \item \textbf{Fourier}: The Fourier model shows some capability in representing higher-frequency components, particularly visible in Figures~\ref{fig:basic_cameraman_prediction} and~\ref{fig:dali_predictions}. However, its overall performance is hindered by slow convergence and poor parameter efficiency, as it requires a large number of coefficients to achieve comparable expressiveness to other models. This limitation becomes critical in Figure~\ref{fig:barbara_predictions}, where the model fails to reconstruct even coarse structures after 300 training epochs, resulting in an almost uniform output.
    
    \item \textbf{KAN}: This one demonstrates higher expressiveness compared to the previous models. It captures some sharp transitions and can learn coarse patterns in signals rich in frequency content, as seen in Figure~\ref{fig:barbara_predictions}. However, the overall reconstructions are still excessively smoothed and lacking fine detail, suggesting that the model acts more as a denoiser than a high-resolution reconstructor.

    \item \textbf{SIREN}: The SIREN model, designed specifically for INR, delivers the best performance among the benchmarks. Its sinusoidal activations enable it to model sharp transitions while retaining a degree of smoothness for generalization. Nonetheless, its slower convergence limits its capacity to fully recover fine-grained structures. For example, in Figure \ref{fig:barbara_predictions}, the stripes on the subject’s clothing and the chair’s intricate details are only partially reconstructed.
    
    \item \textbf{ENN}: Our model achieves the most accurate reconstructions across all tasks. As seen in Figures~\ref{fig:dali_predictions} and~\ref{fig:barbara_predictions}, it consistently outperforms the baselines, achieving MSE values on the order of $10^{-4}$ even where other models remain above $10^{-2}$. While it captures detailed features with high precision (especially in complex backgrounds) it may slightly overfit certain regions, introducing minor artifacts (e.g., in the subject’s pants). This trade-off illustrates the model’s high-frequency sensitivity, which, while advantageous for detail recovery, may lead to unwanted oscillations in some areas.
\end{itemize}

\section{Pruning the \gls{ENN}}

Pruning is the process of removing unnecessary or redundant components, such as neurons, connections, or entire layers, from a neural network. This technique is motivated by several important goals. First, pruning reduces the model size and memory footprint, making it more suitable for deployment in environments with constrained resources. Second, it improves inference speed and computational efficiency, allowing the model to make faster predictions with fewer operations. Third, by simplifying the model, pruning can help mitigate overfitting and improve generalization. Pruning also enhances the interpretability of the network by reducing architectural complexity, making it easier to analyze and understand the role of individual components.
These benefits are especially important in settings with tight resource or latency constraints.

There are several common approaches to prune neural networks \cite{shen2022prune}. The \textit{train-prune-finetune} strategy first trains the full network, then prunes redundant components, and finally retrains the pruned model to recover performance. This approach is widely used and often effective but comes at the cost of extra training time due to the additional fine-tuning stage. A more sophisticated strategy, pruning-aware training, integrates pruning directly into the training process, allowing the model to gradually adjust to a smaller architecture while learning. Although this approach is more complex to implement, it strikes a better balance between efficiency and performance.

However, most of the approaches tackle more complex architectures, such as convolutional neural networks (CNN), whose techniques do not extrapolate to standard MLPs \cite{prune_cnn}. Furthermore, most of the techniques follow heuristics that do not resort on fundamental results \cite{guo2016dynamic}. 


\subsection{Redundancies in the \gls{ENN}}


The expressiveness of the \gls{ENN} reveals interpretable patterns and highlights redundancies that can be effectively targeted through pruning.

\textbf{Negligible DCT coefficients:} A key advantage of the \gls{DCT}-based model is the orthogonality of its coefficients. This property ensures that each coefficient contributes independently to the shape of the \gls{AF}, meaning that pruning one has no direct impact on the others. 
Additionally, the \gls{DCT} is highly effective at compression: it concentrates most of the signal’s energy in the first few coefficients, allowing a compact representation of the \gls{AF}. As a result, many high-order coefficients tend to be negligible and can be safely removed without degrading performance. 
Specifically, each \gls{DCT} coefficient can be individually pruned if it satisfies the condition $|F_{\ell m q}|^2 \leq \rho$, where $\rho$ is an energy threshold that can be adjusted based on the acceptable level of performance degradation.
These characteristics the \gls{DCT} particularly well-suited for pruning strategies, whether applied during training, after convergence, or in iterative schemes, offering flexibility and robustness in model simplification.

\begin{figure}[t]
    \centering
    \begin{subfigure}[b]{0.2\textwidth}
        \centering
        \includegraphics[width=\linewidth]{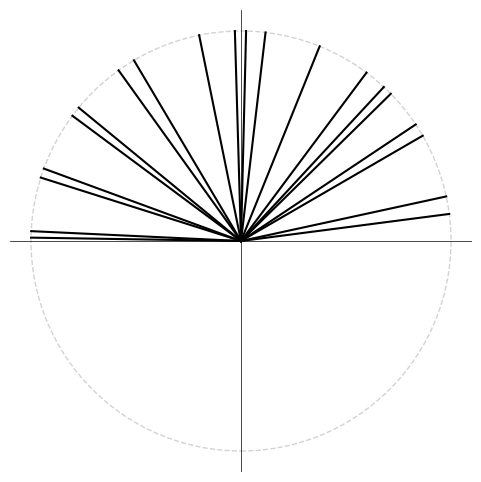}
        \caption{Orientation of $M_1=20$ bumps for the ring problem.}
        \label{fig:bump_angles_shallow}
    \end{subfigure}
    \hfill
    \begin{subfigure}[b]{0.24\textwidth}
        \centering
        \includegraphics[width=\linewidth]{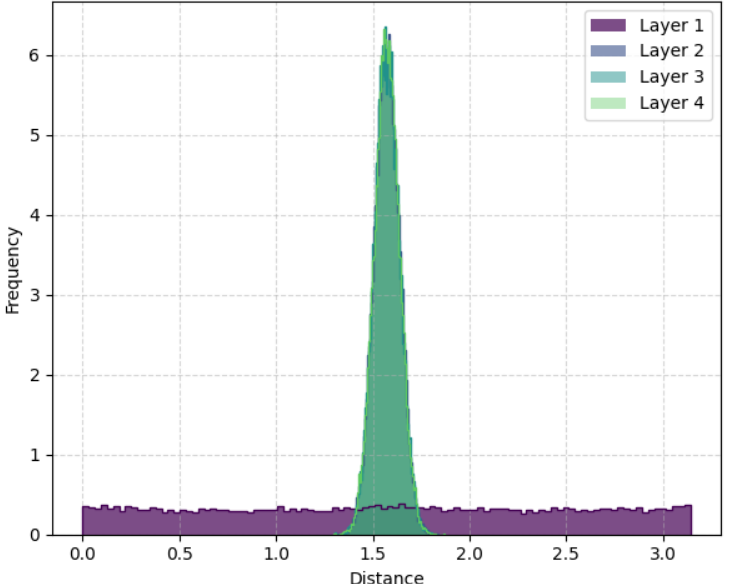}
        \caption{Distribution of angles between bumps in a deep ENN.}
        \label{fig:weight_distance}
    \end{subfigure}
    \caption{Orientation of bumps for shallow (left) and deep (right) networks.}
    \label{fig:distance_distributions}
\end{figure}

\textbf{Redundant bumps:} Another common source of redundancy arises when multiple neurons produce nearly identical bumps, sharing both shape and orientation. While such behavior may occur in various neural architectures, we observe that the \gls{ENN} model often exhibits this pattern. The similarity between \gls{AAF} response can be quantitatively assessed using the cosine distance between the \gls{DCT} coefficients: 
\begin{equation}
\text{dist}\left(\mathbf{F}_{\ell m}, \mathbf{F}_{\ell m'}\right) =
1 - \frac{
{\mathbf{F}_{\ell m}}^T\mathbf{F}_{\ell m'}
}{
\left\|\mathbf{F}_{\ell m} \right\|
\left\|\mathbf{F}_{\ell m'} \right\|,
}
\label{eq:dist_DCT}
\end{equation}
where $\mathbf{F}_{\ell m}\in\mathbb{R}^Q$ is a vector with the coefficients of the $m$th neuron at layer $\ell$ and $\left\|\mathbf{F}_{\ell m} \right\|$ is the Euclidean norm. When coefficients are similar, the projection is 1 and the distance is zero.
When coefficients are close, 
the orientation between bumps can be computed as the angle between the two linear sets of synaptic weights:
\begin{equation}
\text{angle}\left(\mathbf{w}_{\ell m}, \mathbf{w}_{\ell m'}\right) =
\arccos\left(1-\text{dist}\left(\mathbf{w}_{\ell m}, \mathbf{w}_{\ell m'}\right)\right),
\label{eq:dist_weights}
\end{equation}
where $\mathbf{w}_{\ell m}\in\mathbb{R}^{M_\ell}$ is the $m$th column of $\mathbf{W}_{\ell}$.
Two bumps are redundant if \eqref{eq:dist_DCT} and \eqref{eq:dist_weights} are 
simultaneously below some thresholds.
Figure \ref{fig:distance_distributions}(\subref{fig:bump_angles_shallow}) shows the orientation of bumps for the binary ring problem with $M_1=20$ neurons. While all AAF are identical, bumps are oriented in pairs. Thus, at least half of the redundant neurons can be pruned without compromising performance.

Unfortunately, this method for identifying redundant bumps does not scale well to deep networks. In large models, each bump spans a high-dimensional space, causing them to become nearly orthogonal to one another. This effect is illustrated in Figure \ref{fig:distance_distributions}(\subref{fig:weight_distance}), where we train an ENN with $M_0 = 2$ input dimensions and four hidden layers, each with $M_\ell = 256$ neurons. The figure shows the distribution on the cosine similarity \eqref{eq:dist_weights} between each pair of linear weights in each layer. In the first layer, the angles between bumps are uniformly distributed, indicating that the ENN explores the input space in as many directions as possible. However, in deeper layers, the cosine similarity between 256-dimensional weight vectors concentrates around $\pi/2$. This indicates that most pairs of bumps are nearly orthogonal and never point in a similar direction. This curse of dimensionality makes it ineffective to prune similar bumps in deeper layers, and such pruning is only feasible in shallow networks.

\begin{figure}[t]
    \centering
    \begin{subfigure}[b]{0.24\textwidth}
        \centering
        \includegraphics[width=\linewidth]{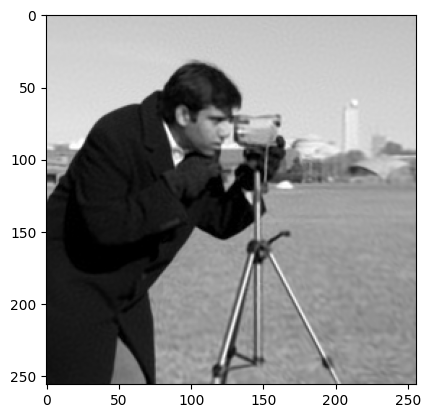}
        \caption{Original model.}
        \label{fig:original_cameraman}
    \end{subfigure}
    \hfill
    \begin{subfigure}[b]{0.24\textwidth}
        \centering
        \includegraphics[width=\linewidth]{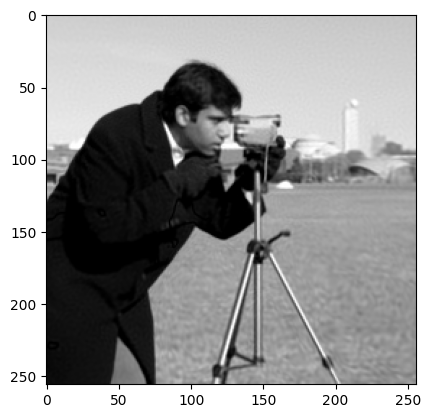}
        \caption{Model after 20\% pruning.}
        \label{fig:20pc_cameraman}
    \end{subfigure}
    \caption{Comparison of image reconstructions before and after coefficient pruning.}
    \label{fig:cameraman_comparison}
\end{figure}

\begin{table}[th]
\centering
\begin{tabular}{cccc}
\toprule
\textbf{Threshold $\rho$} & \textbf{Pruned (\%)} & \textbf{MSE} & \textbf{MSE Factor} \\
\midrule
$4.2 \cdot 10^{-2}$   & 20\%  & $8.0 \cdot 10^{-4}$   & $\times 1$   \\
$6.5 \cdot 10^{-2}$   & 30\%  & $8.5 \cdot 10^{-4}$   & $\times 1.125$ \\
$8.3 \cdot 10^{-2}$   & 40\%  & $8.5 \cdot 10^{-4}$   & $\times 1.125$  \\
$1.05 \cdot 10^{-1}$  & 50\%  & $2.0 \cdot 10^{-3}$   & $\times 2.5$ \\
$1.35 \cdot 10^{-1}$  & 60\%  & $2.2 \cdot 10^{-3}$ & $\times 2.75$ \\
$1.68 \cdot 10^{-1}$  & 70\%  & $3.2 \cdot 10^{-3}$ & $\times4$ \\
\bottomrule
\end{tabular}
\caption{Impact of pruning DCT coefficient for different thresholds.}
\label{tab:coefficient_pruning_results}
\end{table}

Even though detecting redundant bumps is tricky due to the curse of dimensionality, it is still possible to remove negligible DCT coefficients from activations whose value is below a threshold $\rho$. This approach is highly scalable since it is independent of both the width and depth of the network. Moreover, it is also not dependent on the hyperparameter $Q$, representing the number of DCT coefficients used per activation, since each coefficient is evaluated independently. The choice of $\rho$ does not simply depend on the coefficient distribution, but rather has a signal processing orientation, as in the DCT the coefficients that do not contribute much to reconstructing the signal are the ones whose magnitude is close to zero. 

\subsection{Results}
For this experiment, different $\rho$ have been tested, so as to prune different percentages of the total DCT coefficients. As summarized in Table \ref{tab:coefficient_pruning_results}, even when pruning up to 40\% of the activation coefficients there is no performance loss. The first significant performance drop appears when pruning half of the DCT coefficients, specifically in this case the threshold is $1.05 \cdot 10^{-1}$ while the MSE increases by a factor of $\times2.5$. Nevertheless, the MSE remains acceptable for most of the considered thresholds, indicating that the described pruning strategy is robust even to a significant pruning percentage. Moreover, this indicates that the ENN has the capability of achieving low error even after pruning a substantial portion of its activation coefficients, suggesting that the overhead introduced by its adaptive activations is negligible while preserving higher expressiveness than a non-adaptive model. 
One interesting advantage of this pruning technique is that it requires little to almost no fine-tuning, which makes it computationally light. This is because the pruned coefficients corresponded to frequencies that gave a minimal contribution to the overall function, hence their elimination does not significantly affect the model’s performance.

\begin{figure}[t]
    \centering
    \includegraphics[width=1\linewidth]{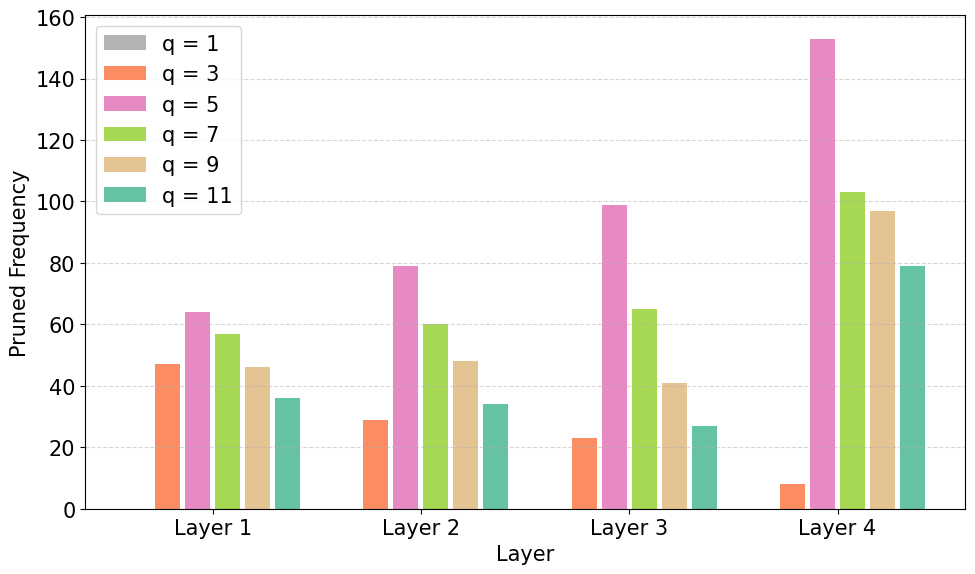}
    \caption{Coefficients pruned per layer, with an overall reduction of 20\%.}
    \label{fig:coeff_pruning_distro}
\end{figure}

To have a better insight into what type of coefficients are being pruned, Figure \ref{fig:coeff_pruning_distro} shows the distribution of the removed coefficients for $20\%$ pruning and for each of the four hidden layers. Clearly, all the layers follow approximately the same distribution: the most pruned coefficients are the ones of medium frequencies, namely $q=5$ and $q=7$, while the lower coefficient $q=1$ is never pruned, probably indicating the importance of low-frequency component to fit the image without extra noise. 
Results show that this distribution is also maintained across different pruning percentages. 

The tendency also shows that pruning happens more in deeper layers. This is consistent with general results on pruning, since early layers learn low-level features that are often essential, while features in deeper layers are more redundant. Also, deeper layers have more room to compensate for pruned parameters using remaining neurons and the propagation of errors has less impact in the rest of the model. The ENN seems to capture this structure as well in the DCT coefficients.

\section{Conclusion}
This work extends and consolidates the \gls{ENN} framework, demonstrating its effectiveness and versatility across diverse learning tasks. Through extensive evaluations on binary classification and INR tasks, the ENN consistently outperforms state-of-the-art architectures while maintaining a remarkably compact structure. Its DCT-based activation parameterization not only enhances expressiveness but also provides an interpretable and well-structured representation that simplifies both configuration and training.

Moreover, the ENN exhibits an inherent advantage in terms of efficiency and scalability. The number of trainable parameters grows linearly with the input dimension, and its decorrelated DCT coefficients allow for straightforward pruning. Empirical results show that up to 40\% of the coefficients can be removed without noticeable degradation in performance, owing to the orthogonality and bounded nature of the DCT basis. This property enables efficient compression with minimal or no fine-tuning, reinforcing the model’s practical applicability.

Overall, these findings highlight the strength of integrating signal processing principles into neural network design. The ENN offers a compelling balance between accuracy, compactness, interpretability and computational efficiency, establishing a solid foundation for future developments in structured and efficient neural architectures.

This work advances the ENN framework by demonstrating how its DCT-based activation parametrization not only enhances expressiveness but also enables meaningful interpretability and efficient pruning. Our experiments show that a significant portion of DCT coefficients (up to 40\%) can be pruned without degrading performance, highlighting the redundancy present in the activation space. Thanks to the orthogonality of DCT coefficients, this pruning requires little to no fine-tuning, making it both practical and computationally efficient. These results underscore the strength of incorporating signal processing principles into neural network design, offering a compelling balance between performance, compactness and explainability.


\bibliographystyle{IEEEbib}
\bibliography{refs}

\end{document}